\documentclass{article}

\usepackage{PRIMEarxiv}
\usepackage[utf8]{inputenc} 
\usepackage[T1]{fontenc}    
\usepackage{url}           
\usepackage{booktabs}       
\usepackage{amsfonts}       
\usepackage{nicefrac}       
\usepackage{microtype}      
\usepackage{lipsum}         
\usepackage{graphicx}      
\usepackage{setspace}      
\usepackage{chngcntr}      
\usepackage{multirow}       
\usepackage{caption}
\usepackage{tabularx} 
\usepackage{float}
\usepackage{hyperref}  

\counterwithout{figure}{section}   
 
\graphicspath{{image/}}             

\newcounter{SIsec}
\renewcommand{\theSIsec}{S\arabic{SIsec}}

\title{Reshaping MOFs text mining with a dynamic multi-agents framework of large language model}

\author{
  \normalfont
  Zuhong Lin\textsuperscript{1,2,3,$\dag$}, 
  Daoyuan Ren\textsuperscript{4,$\dag$}, 
  Kai Ran\textsuperscript{1,2}, 
  Jing Sun\textsuperscript{5}, 
  Songlin Yu\textsuperscript{6}, 
  Xuefeng Bai\textsuperscript{7}, 
  \\[3pt]
  Xiaotiang Huang\textsuperscript{1,2},
  Haiyang He\textsuperscript{8}, 
  \normalfont
  Pengxu Pan\textsuperscript{1}, 
  Xiaohang Zhang\textsuperscript{1}, 
  Ying Fang\textsuperscript{1,9}, 
  Tianying Wang\textsuperscript{1,2}, 
  Minli Wu\textsuperscript{10}, \\[3pt]
  \normalfont
  Zhanglin Li\textsuperscript{10}, 
  Xiaochuan Zhang\textsuperscript{11}, 
  Haipu Li\textsuperscript{1,2,$\ast$}, 
  Jingjing Yao\textsuperscript{1,2,$\ast$}\\[6pt]
  \normalfont
  \textsuperscript{1}Center for Environment and Water Resources, 
  College of Chemistry and Chemical Engineering, \\ Central South University, Changsha, 410083, PR China\\[3pt]
  \normalfont
  \textsuperscript{2}Key Laboratory of Hunan Province for Water Environment and Agriculture Product Safety, \\ Changsha, 410083, PR China\\[3pt]
  \normalfont
  \textsuperscript{3}School of Resources and Environment, Hunan University of Technology and Business, \\ Changsha 410205, P.R.~China\\[3pt]
  \normalfont
  \textsuperscript{4}School of Artificial Intelligence, \\ University of Chinese Academy of Sciences, Beijing, 100049, PR China\\[3pt]
  \normalfont
  \textsuperscript{5}Faculty of Data Science, \\ City University of Macau, 999078, Macao SAR, PR China\\[3pt]
  \normalfont
  \textsuperscript{6}State Key Laboratory of High Performance Ceramics and Superfine Microstructure, \\ Shanghai Institute of Ceramics, \\
  Chinese Academy of Sciences, Shanghai 200050, China\\[3pt]
  \normalfont
  \textsuperscript{7}Beijing Key Laboratory for Green Catalysis and Separation, Department of Chemical Engineering, \\ College of Materials Science and Engineering, \\
  Beijing University of Technology, Beijing 100124, China\\[3pt]
  \normalfont
  \textsuperscript{8}State Key Joint Laboratory of Environment Simulation and Pollution Control, School of Environment, \\
  Tsinghua University, Beijing, 100084, PR China\\[3pt]
  \normalfont
  \textsuperscript{9}School of Chemical Engineering and Materials Science, \\ Yueyang University, Yueyang 414000, PR China\\[3pt]
  \normalfont
  \textsuperscript{10}School of Computer Science and Engineering, \\ Central South University, Changsha, 410083, PR China\\[3pt]
  \normalfont
  \textsuperscript{11}School of Software Engineering, \\ Sun Yat-sen University, Zhuhai 519000, PR China\\[3pt]
  \normalfont
}

\begin{document}
\addtocontents{toc}{\protect\setcounter{tocdepth}{-1}}
\maketitle
\begingroup
  \footnotetext[2]{Zuhong Lin (lzhzzzzwill@hutb.edu.cn) and Daoyuan Ren (rendaoyuan23@mails.ucas.ac.cn) contribute equally in this work.}
  \footnotetext[1]{Haipu Li (lihaipu@csu.edu.cn) and Jingjing Yao (yaojj0412@163.com) are corresponding authors.}
\endgroup

\begin{abstract}
Accurately identifying synthesis conditions for metal–organic frameworks (MOFs) remains a critical bottleneck in materials research, 
as translating literature-derived knowledge into actionable insights is hindered by the unstructured and heterogeneous nature of scientific texts. 
Here we present MOFh6, a large language model (LLM)-based multi-agent system designed to extract, structure, and apply synthesis knowledge from diverse input formats, including raw literature and crystal codes. 
Built on gpt-4o-mini and fine-tuned with up to few-shot expert-annotated, MOFh6 achieves 99\% accuracy in synthesis data parsing and resolves 94.1\% of complex co-reference abbreviations. 
It processes a single full-text document in 9.6 seconds and localizes structured synthesis descriptions within 36 seconds, with the cost per 100 papers reduced to \$4.24, a 76\% saving over existing systems. 
By addressing long-standing limitations in cross-paragraph semantic fusion and terminology standardization, MOFh6 reshapes the LLM-based paradigm for MOFs synthesis research, transforming static retrieval into an integrated and dynamic knowledge acquisition process. 
This shift bridges the gap between scientific literature and actionable synthesis design, providing a scalable framework for accelerating materials discovery.
\end{abstract}


\onehalfspacing
\section{Introduction}
Metal–organic frameworks (MOFs) are crystalline porous solids assembled from inorganic secondary building units interconnected by organic linkers into predictable, modular architectures\cite{RN1, RN2, RN3}.
Although MOFs offer great compositional and topological diversity due to their structural tunability, conventional trial-and-error synthesis remains inefficient for exploring such an expansive chemical space\cite{RN4, RN5, RN6, RN70, RN71}.
By 2022, the Cambridge Crystallographic Data Center (CCDC) had documented more than 110 thousand MOF structures, reflecting exponential growth and revealing the limitations of empirical discovery methods\cite{RN7}, 
underscoring the imperative shift toward data-driven intelligent material discovery paradigms.

Scientific literature underpins MOF research but is largely unstructured\cite{RN9,RN10,RN11,RN12,RN72,RN73,RN74}. 
Diverse linguistic styles yield inconsistent descriptions of synthesis conditions. 
Cross-sentence references and ambiguous abbreviations break semantic continuity and hide key parameters. This ambiguity inflates information entropy during text-to-data conversion.
Park et al. \cite{RN13} extracted synthesis parameters from the literature through manual annotation combined 
with bidirectional long short-term memory networks and constructed a structured JSON database. 
Glasby et al. \cite{RN14} developed the DigiMOF database based on the ChemDataExtractor tool \cite{RN8}, 
systematically integrating structural characteristics and synthesis process parameters for 15501 MOFs, 
including hydrothermal temperature and solvothermal time.
Current natural language processing (NLP) tools still struggle with semantic complexity and require extensive coding. High-precision, flexible extraction of synthesis details remains an open challenge.

Large language models (LLMs) driven significant advancements in NLP paradigms, particularly within material science research \cite{RN15, RN16, RN17}. 
Yaghi’s team pioneered efficient extraction of MOF synthesis parameters using prompt engineering with GPT-3.5-turbo across 228 research articles \cite{RN18}. 
Shi et al. further enhanced extraction accuracy via few-shot learning, demonstrating optimal GPT-4-turbo performance with only four training samples \cite{RN19}. 
Recently, Kang et al. constructed a comprehensive database from over 40,000 literature sources by fine-tuning GPT-3.5-turbo and using prompt-driven GPT-4 \cite{RN20}. 
Moreover, the ChatMOF system introduced by Kang illustrates the extension of LLM capabilities beyond pure text mining, integrating MOFTransformer \cite{RN23} and the genetic algorithm pormake \cite{RN24} for natural language-driven inverse design and performance prediction of MOFs \cite{RN21, RN22}. 
However, existing LLM-based approaches still face notable limitations, particularly in accurate cross-paragraph semantic fusion and in the standardization of highly variable chemical terminologies, underscoring the urgent need for more integrated and robust solutions.

\begin{figure}[hbt]
  \centering
  \includegraphics[width=0.8\textwidth]{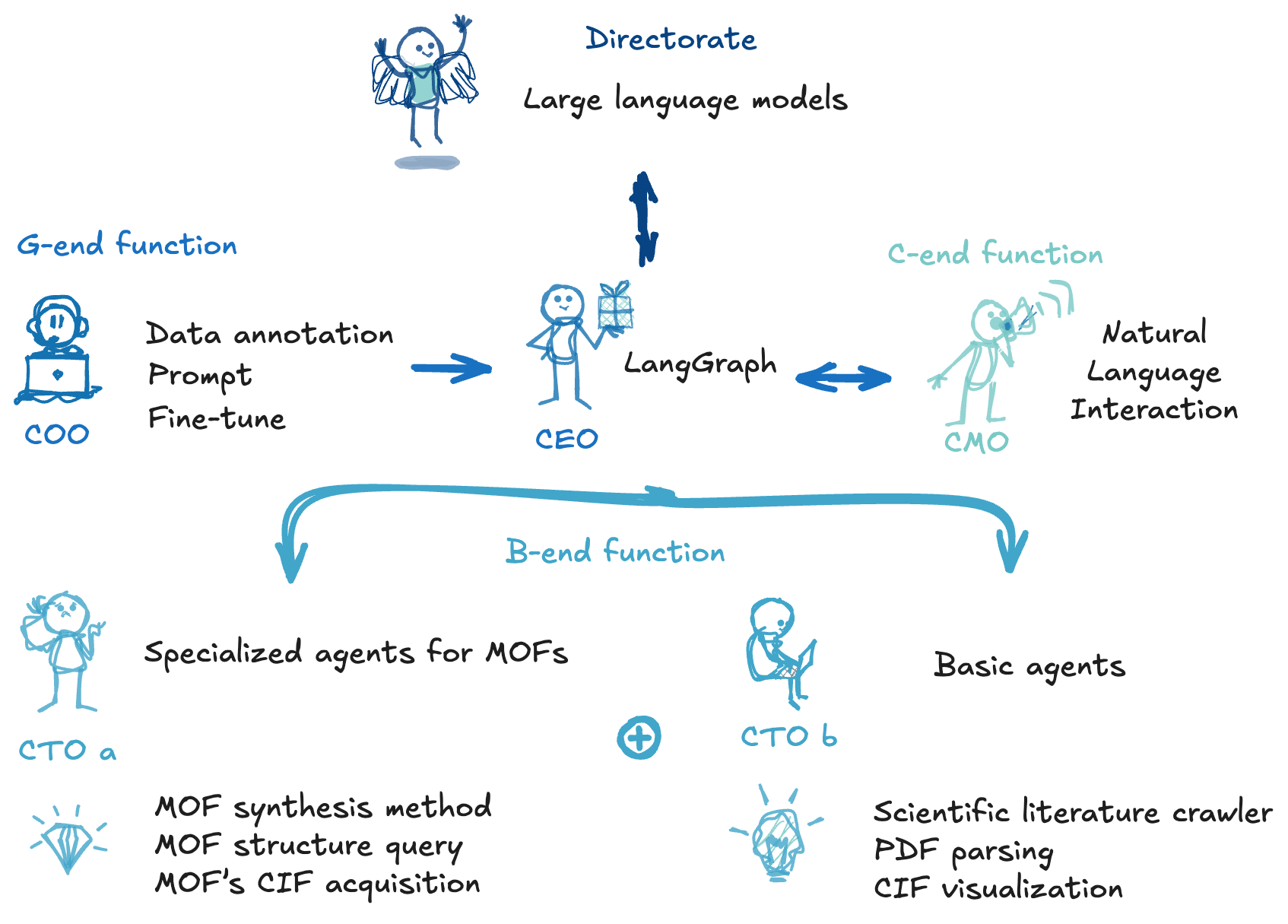}
  \caption{
  \centering
  Enterprise architecture of MOFh6.
  }
  \label{MOFh6frame}
\end{figure}

Building upon recent advances in LLM-based MOF information mining and interactive operations, 
we have developed MOFh6, an enterprise-level intelligent interaction system that provides end-to-end services, 
from text mining to application-oriented outputs. 
Guided by a hierarchical architecture concept, MOFh6 was designed around strategic decision-making, technical implementation, and task execution (Figure \ref{MOFh6frame}), 
A full-chain service system was constructed with four core roles operating collaboratively. 
The CEO is responsible for central coordination, leveraging the Langgraph framework;
the CMO oversees data management and user interaction;
the COO, comprising human experts, focuses on data annotation and workflow design;
the CTO, constituted by LLM agents, leads the integration and deployment of core technological capabilities.
Built on this organizational synergy, MOFh6 orchestrates multi-agent collaboration to support key functionalities across its core modules.
The B-end application has achieved full-text precise localization and structured mining of specified MOF synthesis conditions, 
overcoming the limitations of traditional NLP in terms of insufficient semantic reference parsing ability, 
difficulty in unifying terminology diversity, and lack of cross-paragraph semantic fusion mechanism. 
This module also supports users to implement literature crawling, MOF pore structure parameter query, 
and crystal structure visualization functions without the need for programming knowledge.
The C-end interaction module enables users to conveniently access various functions of B-end applications through natural language interfaces.
Ultimately, MOFh6 addresses the core challenge of translating literature-derived synthesis knowledge into actionable insights for materials design and evaluation.

\section{Results}

\subsection{Multi-tasking empowered MOFh6 pipeline}

\begin{figure}[hbt]
  \centering
  \includegraphics[width=1\textwidth]{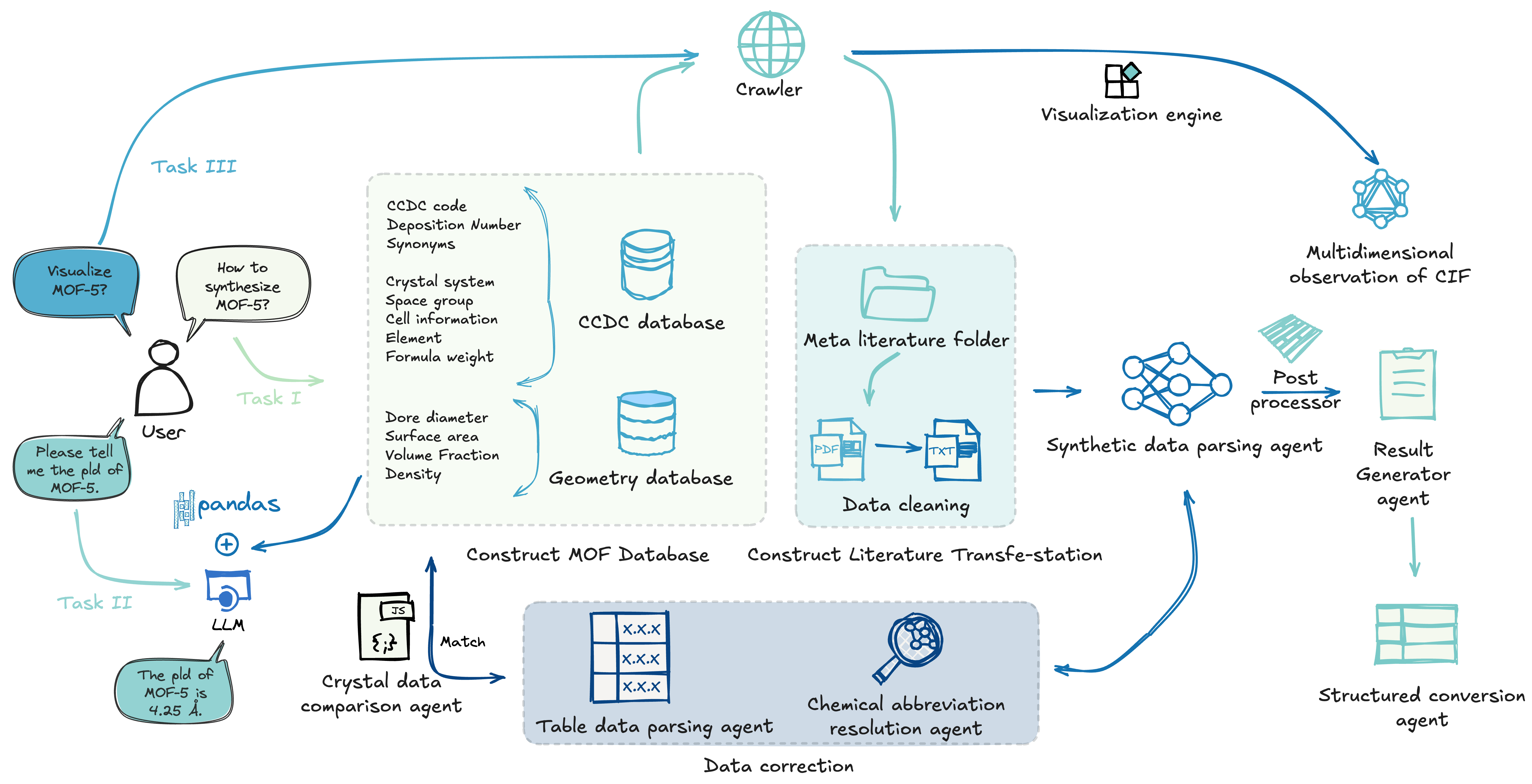}
  \caption{
  \centering
  Overview  pipeline of MOFh6. Task I operates through collaborative LLM agents; Task II integrates LLMs with a rule engine for constrained synthesis query parsing; and Task III leverages the Qt framework and the Hugging Face ecosystem to support structure visualization and CIF services.
  }
  \label{MOFh6pipline}
\end{figure}

MOFh6 reconstructs the MOF knowledge extraction paradigm by unifying LLM-based semantic parsing, rule-guided refinement, and interactive crystallographic services into a coherent multi-agent workflow (Figure \ref{MOFh6pipline}).
In the synthesis process analysis Task I, the system matches the publisher of the queried MOF through a DOI routing module, 
dispatches institutionally authorized customized crawler scripts, 
and achieves compliant acquisition of target literature and standardized conversion from PDF to TXT format. 
The synthesis information processing unit uses fine-tuned GPT-4o-mini to build a multi-level information extraction pipeline.
Firstly, MOFh6 reconstructing complete semantic contexts through the synthesis data parsing agent to obtain all descriptions about synthesis from the TXT text. 
Since most scientific papers provide crystallographic data for MOFs in tabular form when reporting them  \cite{RN25, RN26}, 
the table data parsing agent, crystal data comparison agent, 
and post-processor then refine and filter out the synthesis information corresponding to the queried MOF. 
For H\textsubscript{x}L\textsubscript{x}, L\textsubscript{x}H\textsubscript{x}, L\textsubscript{x} (x$\ge$0) type abbreviations referring to organic linkers, 
inspired by Shi et al. \cite{RN19}, 
the chemical abbreviation resolution agent uses a dual-verification mechanism 
of regular expressions \& LLM to resolve co-references from abbreviations to full names. 
The final standardized synthesis parameter table covers 14 key indicators \cite{RN18, RN19, RN27} including metal precursors, 
organic linkers, solvent systems, and synthesis environments (Table \ref{LLMIDex}), 
accommodating the processing needs of literature with different writing styles. 
MOFh6 simultaneously supports user-defined PDF uploads, 
executing an information extraction process with the same precision.

In the structure property analysis Task II, 
MOFh6  has deployed a metadata retrieval tool that supports users in retrieving information such as CCDC codes, 
chemical naming systems, and crystallographic parameters of specified MOFs through natural language queries(Figure \ref{MOFh6pipline}).
MOFh6's context-aware session management mechanism automatically resolves implicit references 
such as "this material" and establishes a paging index mechanism 
(e.g., "show 5 more results") to optimize the interactive experience of datasets. 
MOFh6's memory optimization strategy (which can cause it to forget when the context cache exceeds capacity) 
effectively avoids resource overload risks while ensuring smooth interaction. 
MOFh6 has a multi-attribute combination query engine that can handle complex retrieval conditions 
including extreme value screening and interval constraints, 
while integrating advanced analytical functions such as mean calculation 
and conditional statistics to form interactive data analysis capabilities (Text \ref{S4}).

The crystal service Task III constructs a full-cycle management system for crystallographic information files (CIFs). 
When users request crystal structure files through CCDC codes, 
they can choose whether to trigger the three-dimensional visualization engine, 
generating an interactive interface displaying ball-and-stick models and unit cell frameworks, 
supporting rotation, scaling, and other multi-dimensional observation modes (Figure \ref{viscif}).

\subsection{User oriented MOFh6 Q\&A Robot}

Existing work shows that interactive systems based on LLMs can effectively enhance the interpretability of material data. 
The dialogue-based retrieval platforms constructed by Zheng et al. \cite{RN18} and Kang et al. \cite{RN20} 
reduce the cognitive threshold for non-professional users by combining structured datasets with customized LLM services. 
Unlike dialogue systems that rely on static databases, 
MOFh6 establishes a real-time, closed-loop workflow integrating data acquisition, 
processing, and response generation. When a user initiates a query, 
MOFh6 dynamically mines original literature related to the target MOFs, 
rather than retrieving information from pre-constructed datasets. 
Following data acquisition, 
MOFh6 uses a task-decoupling strategy to categorize both prompt engineering-based and fine-tuned LLM agents into three coordinated stages,
information extraction, logical verification, and semantic structuring. 
Specifically, information extraction agents consist of synthesis data parsing agents, 
table data processors, and chemical abbreviation resolution agents. 
Logical verification is handled by crystal data comparison agents and result generation agents, 
while semantic structuring involves post-processors and structured conversion agents. 
This comprehensive, three-tier quality assurance mechanism effectively reduces hallucination risks and lowers the costs associated with LLM utilization
(Text \ref{S2}, Figure \ref{1-1sprompt_2}-Figure \ref{ftprompte_2}).
In doing so, MOFh6 reshapes the traditional role of LLMs from static responders to dynamically orchestrated agents for scientific knowledge extraction.

\begin{figure}[hbt]
  \centering
  \includegraphics[width=1\textwidth]{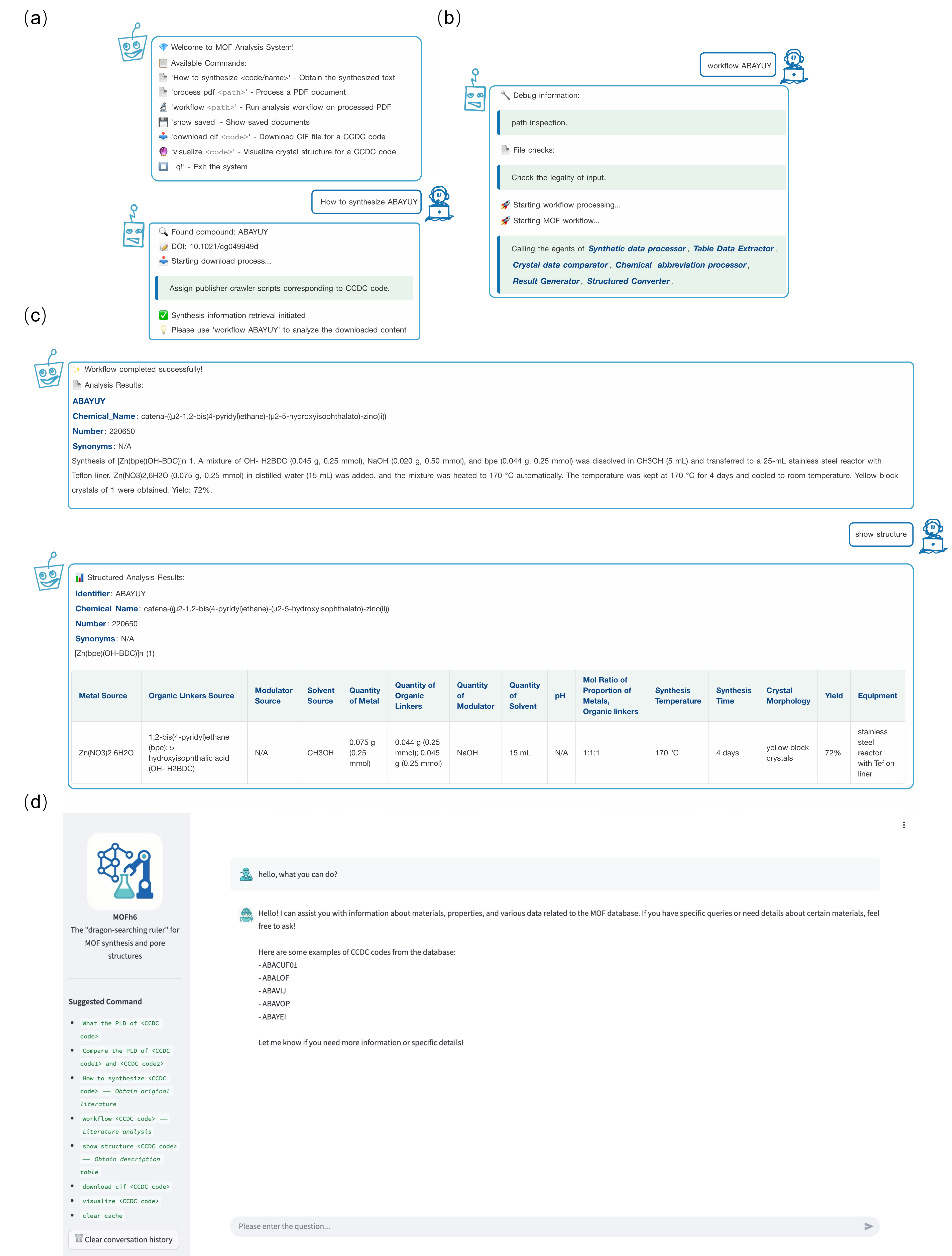}
  \caption{
  \centering
   Query \& Answer of MOFh6. User commands for initiating literature crawling (a), sequential agent-based mining of synthesis data (b), output of structured synthesis parameters (c).
  }
  \label{LLMM4}
\end{figure}

Building upon its modular agent architecture, MOFh6 further delivers interactive, low-latency query services tailored to diverse user inputs and research needs (Figure \ref{LLMM4}).
Upon user login, the interactive interface automatically loads guided query templates (synthesis condition analysis, 
PDF processing, CIF visualization, etc.). 
When users submit CCDC code queries, MOFh6 retrieves the corresponding literature in TXT format through compliant crawlers. 
It then activates the seven-stage agent processing pipeline described in Table \ref{LLMagent} to extract synthesis parameters. 
The final output consolidates key information, including the chemical name, 
CCDC number, commonly used abbreviations (such as ZIF-8), and detailed synthesis procedures. 
Additionally, MOFh6 supports one-click conversion of synthesis steps into structured Markdown tables (Figure \ref{1-2s_1}-Figure \ref{1-2s_4}).
Since our final input is in plain text TXT format, 
which greatly reduces the recognition difficulty for LLMs \cite{RN31, RN58}, 
the average time consumption for the entire process is 36 seconds, 
with an average cost of less than 0.05\$ per query (Text \ref{S5}).

MOFh6 adopts a dynamic context-aware architecture to implement intelligent query services, 
supporting users in querying MOF structural parameters in natural language (Figure \ref{er_1}-Figure \ref{er_9}). 
The system uses a JSON-Schema-driven semantic conversion engine that precisely maps natural language instructions 
to 18 structured query fields (Figure \ref{LLTS3}). The context management engine continuously tracks conversation states, 
records material screening sets, property constraint conditions, 
and historical operation trajectories, achieving semantic inheritance and logical coherence across multiple rounds of queries. 
MOFh6's multi-level query classification mechanism ensures the professionalism and efficiency of processing paths, 
with an average response time of only 3 seconds per query and an average API call cost controlled within 2.5×10\textsuperscript{-4}\$. 
MOFh6 also provides users with convenient CIF file acquisition channels, 
allowing users to flexibly choose direct file downloads or use the platform's 
visualization functions for immediate viewing and analysis of CIF files according to research needs, 
thereby providing strong support for subsequent in-depth research. 
Additionally, MOFh6 also meets the basic multilingual interaction capabilities of LLMs, 
supporting users in asking questions in different natural languages (Figure \ref{er_9}). 

\subsection{Data mining statistics from data to knowledge transformation}

Figure \ref{LLMM3} presents the results of structural and bibliographic mining enabled by MOFh6 across CCDC and literature datasets, highlighting key chemical preferences, crystallographic patterns, and data source distributions that shape modern MOFs synthesis.
Metal salts containing Cu, Zn, Cd, and Co elements occupy the top four usage frequencies in MOF synthesis, 
mainly attributed to their historical inertia, 
cost advantages, and stable +2 valence coordination characteristics \cite{RN28} (Figure \ref{LLMM3}a). 
The crystallographic statistics show that low-level crystal systems such as monoclinic and triclinic occupy dominant positions, 
with their structural flexibility mainly stemming from the conformational diversity of carboxylic acid linkers in most MOF syntheses. 
Notably, these low-symmetry crystal systems often correspond to the final stable state of MOF materials after activation or oxidative treatment \cite{RN29}. 
In contrast, the remaining medium and high-symmetry crystal systems account for only a smaller portion of the dataset, 
predominantly formed by rigid linkers and SBUs combinations (Figure \ref{LLMM3}b) \cite{RN30}. 
The space group distribution shows the same differential trend as the crystal systems (Figure \ref{LLMM3}c), 
with MOFs having low-symmetry space groups like \textit{P-1}, \textit{P21/c}, 
and \textit{C2/c} accounting for higher proportions than those with medium and high-symmetry space groups.

The results of literature crawling (Figure \ref{LLMM3}d) show that after 1995, 
most MOF structural studies were concentrated on three major publisher, the American Chemical Society (ACS), 
the Royal Society of Chemistry (RSC), and Elsevier. 
Among them, ACS journals such as \textit{Inorganic Chemistry} and \textit{Crystal Growth \& Design} became their main publication platforms; 
RSC's \textit{CrystEngComm} and \textit{Dalton Transactions} occupied the leading positions in their fields. 
Elsevier's \textit{Inorganica Chimica Acta}, \textit{Polyhedron}, 
and \textit{Inorganic Chemistry Communication} were more popular among researchers (Figure \ref{LLMS31}). 
This distribution characteristic is closely related to the traditional advantages 
and positioning of various journals in the field of coordination chemistry, 
reflecting the continuous preference of MOF research groups for professional chemistry journals.

The mining results of MOF structures (Figure \ref{LLMS32}) show that the dataset exhibits a significant concentration trend in pore characteristics, 
with LCD mainly distributed in the 2-4 Å range and PLD concentrated in the 0-2 Å range. 
In the specific surface area parameters, 
both VSA and GSA are primarily distributed in the 0-200 m\textsuperscript{2}/cm\textsuperscript{3} \& m\textsuperscript{2}/g interval. 
These statistical characteristics cover most known MOF structure types, 
providing benchmark references for structural adaptation in different application scenarios. 
Through the integration of retrieval and statistical dual modes, 
the MOFh6 system allows users to both precisely obtain structural parameters 
of individual MOFs and retrieve overall distribution characteristics of the dataset in real-time, 
achieving fast-response structural feature analysis services.

\subsection{Agents' performance of human-computer interaction verification}

\begin{figure}[hbt]
  \centering
   \includegraphics[width=1\textwidth]{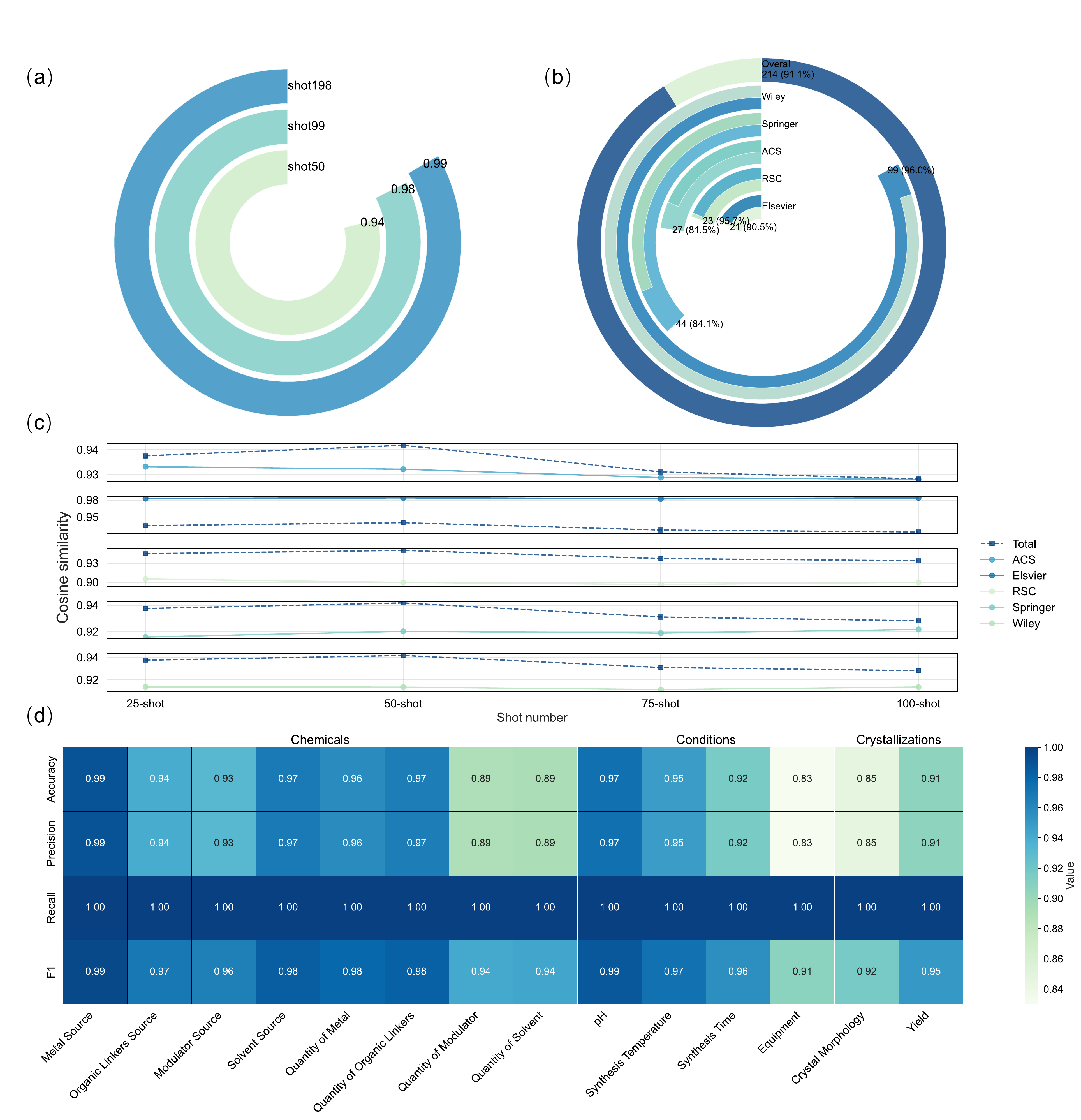}
   \caption{
   \centering
   Agents performance evaluation: 
   The performance of different sample fine-tuning models on their test set in the dataset (a),
   the ability of coreference resolution (b), 
   the ability of the model to extract specific MOFs synthesis paragraphs under different pool sizes (c), 
   and the ability of the model to extract structured data (d).
   }
   \label{LLMM5}
 \end{figure}

Numerous works have shown that LLM sample pool and model performance follow a positive correlation for performance optimization \cite{RN33, RN34, RN35}. 
Figure \ref{LLMM5}a shows the impact of increasing fine-tuning sample pool on model performance in the synthesis data parsing agent. 
When using a pool of 50 expert-annotated samples for fine-tuning, the model achieved a prediction accuracy of 0.94 on the test set. 
However, considering the significant differences in writing styles among global researchers, 
especially the bridging reference phenomena common in scientific papers, 
a smaller scale training sample pool is still challenging to precisely resolve cross-textual features. 
To address this, we further increased the fine-tuning sample pool to 99, at which point the model prediction accuracy significantly improved to 0.98, 
indicating that MOFh6's synthesis data parsing agent had developed cross-textual style adaptation capabilities. 
Finally, as the fine-tuning sample pool expanded to 198, the model accuracy further improved to 0.99, 
further confirming that MOFh6's synthesis data parsing agent could effectively handle the extraction 
and bridging reference resolution tasks for most synthesis paragraphs in scientific literature.

In MOF synthesis literature, the naming of organic linkers presents significant complexity, 
with researchers frequently using abbreviations like H\textsubscript{x}L\textsubscript{x}, L\textsubscript{x}H\textsubscript{x}, L\textsubscript{x} for co-reference expressions, 
posing additional challenges for automated parsing \cite{RN19}. 
To solve this problem, 
we designed a specialized chemical abbreviation resolution agent based on the GPT-4o model, 
with its performance shown in Figure \ref{LLMM5}b. In tests on 500 MOFs across journals from five major publishers, 
the agent accurately identified and resolved 214 instances of co-reference abbreviations, 
achieving an overall success rate of 94.1\%. 
However, differences in writing standards between publishers also caused fluctuations in co-reference resolution effectiveness, 
with the Wiley database performing best at a 96\% success rate, 
while the ACS database had a reduced success rate of 81.5\% due to the impact of textual expression peculiarities. 
Nevertheless, the resolution success rate for all databases exceeded 80\%, 
demonstrating stable parsing capabilities especially when processing texts involving complex chemical nomenclature, 
confirming the agent's practical value in actual scientific research applications.

Furthermore, researchers typically use bold symbols or numbers to bridge references to numerous MOF structures in the same literature \cite{RN36, RN37, RN38, RN39}. 
While these abstract expressions are concise, they may affect the quick understanding of newcomers to the field. 
Our synthesis data parsing agent can effectively resolve bridging reference phenomena, 
achieving complete semantic restoration. 
As shown in Figure \ref{LLMM5}c, when users submit queries with specific CCDC codes (such as ABAYUY, which includes three crystals in the literature \cite{RN40}), 
the system first automatically extracts multiple crystal synthesis paragraphs from the original literature, 
then generates complete semantic descriptions through bridging reference resolution. 
These structured data are subsequently processed further by the table data parsing agent and crystal data comparison agent, 
ultimately achieving precise localization of the synthesis description corresponding to the target crystal code (Figure \ref{llmcase}). 
System evaluations show that as the sample pool expanded from 25 to 100, MOFh6 agents' 
comprehensive accuracy in parsing synthesis information from literature across five major publishers fluctuated slightly between 0.94 and 0.93, 
maintaining overall performance stability within $\pm$1\%. 
Specifically, the Elsevier database exhibited extremely strong robustness, 
with cosine similarity between post-text parsing and manually annotated text remaining stable between 0.97 and 0.98, 
while the RSC database performed relatively weaker, maintaining cosine similarity between 0.89 and 0.90. 
The parsing accuracy for other publishers generally remained above 0.91. 
Overall, MOFh6 possesses cross-scale sample adaptation capabilities 
and can effectively achieve accurate restoration and localization of synthesis paragraphs corresponding to given CCDC codes.

To enhance the readability and practical utility of information on MOF synthesis conditions, 
we structured the chemical composition (Chemicals), reaction conditions (Conditions), 
and crystal characteristics (Crystallization) using the 3C format (Figure \ref{LLMM5}d).
Regarding specific chemical components, the identification of metal salts exhibited exceptional performance, 
largely due to the clear correspondence between chemical formulas and their full names, 
typically provided explicitly in literature sources. 
This clarity resulted in accuracy, precision, recall, and F1 scores all exceeding 0.99. 
In contrast, organic linkers presented higher semantic complexity, particularly between structurally similar isomers, 
such as "(R)-2,2'-dihydroxy-1,1'-binaphthyl-4,4',6,6'-tetrabenzoate" and "(R)-2,2'-diethoxy-1,1'-binaphthyl-4,4',6,6'-tetrabenzoate" \cite{RN41}. 
This complexity slightly impacted identification accuracy and precision, lowering these scores to approximately 0.94, 
although recall and F1 scores remained above 0.94. 
Additionally, additives used for pH adjustment (e.g., acids, bases, and triethylamine) \cite{RN18, RN42} and natural language descriptions of solvents are relatively standardized, 
resulting in stable identification performance with comprehensive scores exceeding 0.93.
While the overall identification performance for quantities of metal salts, 
organic linkers, and additives remained excellent, with scores consistently above 0.90, 
challenges arose when processing combinations involving multiple solvents or repeated mentions \cite{RN43}. 
Due to inherent limitations in arithmetic reasoning capabilities of LLMs, 
slight deviations occurred in the calculation of quantities, marginally reducing the comprehensive score to approximately 0.89.
In terms of Conditions, the structured conversion agent demonstrated stable and precise parsing performance for single-parameter conditions (e.g., pH, synthesis temperature, synthesis time), 
achieving comprehensive scores above 0.92. 
However, the accuracy was somewhat diminished when dealing with multi-stage equipment descriptions, 
such as "initial hydrothermal synthesis in a Teflon reactor followed by evaporation crystallization in a beaker" \cite{RN42}, 
resulting in a slightly reduced comprehensive performance score of approximately 0.83.
Regarding Crystallization, the structured conversion agent effectively extracted crystal morphology descriptions, 
achieving a comprehensive accuracy above 0.85. 
Its capability in recognizing yield, which is quantitative information, 
was particularly notable, reaching precision levels above 0.91. 
This performance underscores the agent’s strong ability to handle information containing both qualitative and quantitative elements.
Overall, the MOFh6 structured conversion agent demonstrates commendable stability and adaptability when processing complex and diverse chemical texts.

\section{Conclusion}

In this work, we developed MOFh6, an end-to-end, enterprise-level intelligent system that enables full-chain interaction for MOFs synthesis knowledge extraction and application. By orchestrating multiple agents in a collaborative framework, MOFh6 reshapes the conventional pipeline of MOF informatics, integrating dynamic literature acquisition, context-aware parsing, and structured knowledge generation into a unified workflow. This design significantly lowers the cognitive barrier for non-specialist users and reduces the data mining costs.
Experimental results show that a fine-tuning strategy based on 198 sets of expert annotations achieved a synthesis parameter extraction accuracy of 0.99, 
effectively overcoming the challenge of cross-literature bridging reference resolution. 
For the heterogeneous co-reference phenomenon of ligand abbreviations in chemical texts, 
the system demonstrated an average resolution success rate of 94.1\% in tests across literature from five major publishers. 
The system maintained a precision of 0.93±0.01 for synthesis descriptions of specified MOFs when the sample scale was expanded. 
Based on 3C structuring format, the system achieved accuracy, recall, 
and F1 scores above 0.8 for various structured descriptions of MOF synthesis, 
establishing a standardized data pipeline for high-throughput material development.
The successful validation of MOFh6 reshapes the feasibility of the LLM agents and rule engine-driven paradigm for acquiring synthesis methods of MOFs and broader material systems, with its modular architecture enabling an intelligent closed loop from database mining to synthesis protocol generation.
However, the current system is still limited by third-party tool compatibility in terms of dynamic data source integration. 
In the future, it will achieve seamless integration of CSD-Python-API and the LLM ecosystem through architectural upgrades, 
promoting the evolution of material development towards a fully digitalized stage of flexible human-machine interaction.

\section{Methods}

\subsection{Construction of crystallographic dataset}

The MOF crystal database constructed in this study ultimately formed a standardized dataset containing 43766 crystal structures. 
Through the CSD-Python-API interface of the Cambridge Crystallographic Data Centre (CCDC) \cite{RN45}, 
we achieved automated extraction and structured processing of crystal information, 
generating JSON format metadata including multiple dimensional features such as MOFs' CCDC codes, 
numbers, chemical names, abbreviations, DOI numbers and corresponding URLs of publications, 
as well as space group information, crystal information, unit cell lengths, angles, elemental composition, and molecular weight.

Simultaneously, through the open-source software Zeo++ \cite{RN46}, 
using N\textsubscript{2} with a kinetic radius of 1.82 Å as a molecular probe, 
we calculated core physical parameters of MOFs including pore limiting diameter, 
largest cavity diameter, crystal density, volumetric accessible surface area, 
gravimetric accessible surface area, and volume fraction, constructing a physical parameter database for MOFs.

\subsection{Literature data mining}

We strictly adhered to academic publishing ethical guidelines and digital copyright management standards in our literature acquisition work. 
Relying on institutional subscriptions to mainstream publishing platforms such as Elsevier, 
Wiley, American Chemical Society (ACS), Royal Society of Chemistry (RSC), and Springer, 
we systematically constructed a target literature dataset within 
the range of institutionally authorized IPs using an automated literature mining system. 
Compared to existing research limited by platform interfaces or unable to fully obtain supporting materials \cite{RN20, RN47}, 
MOFh6's crawler module achieves deep parsing of full-text formats, 
specifically addressing the disciplinary characteristic where synthesis descriptions are often distributed in supporting materials. 
By completely acquiring the main body of the literature and supplementary material content, 
it constructs a complete underlying data architecture for fine-grained semantic analysis by LLMs, 
effectively solving the parsing challenge of fragmented distribution of key information in scientific literature.

\subsection{Literature processing system framework}

The core of our text processing system is based on LangGraph\footnote{\href{https://github.com/langchain-ai/langgraph}
{https://github.com/langchain-ai/langgraph}}
constructing a three-layer processing architecture in the form of a directed state graph, 
achieving collaborative processing of multiple modules through declarative workflows (Table \ref{LLMagent}). 
The first layer is the data collection and preprocessing structure, 
where the synthesis data parsing agent parses input documents through fine-tuned LLMs, 
focusing on solving bridging reference problems in synthesis paragraphs 
(e.g., context-dependent expressions like "this method," "above conditions"), 
generating synthesis description texts with complete semantics. 
The table data parsing agent, running in parallel, 
adopts LLM prompt engineering to precisely identify crystallographic parameter tables 
in documents (including unit cell parameters, space groups, etc.) and convert them into standardized JSON format. 
The second layer is the core data processing structure, 
where the crystal data comparison agent establishes associative mapping between preprocessed data, 
matching crystallographic parameters in tables with specific crystals described in synthesis texts through feature matching algorithms. 
For chemical abbreviation patterns like H\textsubscript{x}L\textsubscript{x}, L\textsubscript{x}H\textsubscript{x}, 
L\textsubscript{x} (x$\ge$0) in the literature, 
the chemical abbreviation resolution agent adopts a dual verification mechanism combining regular expression rule libraries with constrained prompt engineering, 
achieving normalized naming of chemical entities while controlling LLM hallucination risks. 
The third layer is the structured output structure, 
where the result generation agent constructs a document-level co-reference resolution pipeline based on the BM25 algorithm, 
integrating previous processing results to generate standardized MOF synthesis description texts. 
The post-processor executes multi-file output management, generating independent sub-files for each input document. 
The structured conversion agent converts the final text into a Markdown table containing structured fields 
such as synthesis conditions and crystal parameters through predefined templates (Text~\ref{S1}).

The system achieves state transfer between nodes through directed edges, 
with each processing module supporting independent configuration and extension, 
demonstrating stable processing performance and effectively solving key problems in chemical literature processing 
such as reference resolution and cross-modal data association. 
This architecture provides a verifiable technical path for automated processing of textual data in the field of materials science.

\subsection{Query \& Answer system framework}

To help non-professionals quickly obtain structural and synthesis information about MOFs, 
we developed an intelligent question-answering system based on natural language interaction. 
This system adopts a three-layer architectural design, 
achieving efficient information retrieval through intelligent data parsing and visualization technology, 
with conversational query and interaction capabilities.

At the data architecture level, the system constructs a standardized preprocessing workflow, 
implementing efficient positioning of raw data through the Pandas data framework, and adopting a dual strategy combining LLMs with dynamic field mapping. 
This design effectively avoids the hallucination problems that might be produced by a single LLM, 
while supporting users in querying using any structural description terminology. For complex query scenarios, 
the system developed a context-aware query parsing mechanism, 
achieving semantic association understanding in conversational scenarios through multi-level mapping dictionaries and persistent context management. 
The core parsing module adopts a primary-backup dual-engine design, 
with the main parser completing semantic conversion from natural language to JSON structures through LLMs, 
and the backup parser ensuring the reliability of key parameter extraction through function calls, 
accurately identifying seven types of query modes including direct queries, range queries, and comparison queries.

The interaction system adopts a state-management command-line interface, 
integrating diversified functions such as PDF parsing, CIF downloading, and visualization. 
The PDF processing module implements parallel extraction of document structure and content based on the PyMuPDF engine, 
and can effectively process special symbols and layout anomalies in academic literature through a text cleaning process optimized by regular expressions. 
For literature materials uploaded by users, the system preserves paragraph semantic structures through standardized processing, 
laying the foundation for subsequent extraction of synthesis information.

For specific MOF synthesis information queries, the system deploys a crawler router. 
When users submit CCDC numbers, the DOI routing module automatically identifies the corresponding publisher's DOI and URL, 
triggering customized crawlers to obtain full text. 
The crystal structure visualization module adopts a dual-engine architecture of pymatgen and py3Dmol, 
achieving precise conversion from CIF files to interactive three-dimensional models. 
The visualization interface, developed based on the PyQt5 framework, 
clearly displays the channel topology and coordination modes of MOF materials through a combination of ball-and-stick models and unit cell parameters, 
supporting multi-dimensional observation including rotation and scaling.

\subsection{Prompt engineering and fine-tuning of model}

In LLM applications, prompt engineering has been proven to be a key technical strategy. 
By setting clear instructions, constructing contextual constraint conditions, 
and providing a small number of sample cases, 
it can effectively suppress model hallucination phenomena and significantly improve information extraction precision and 
data generation quality \cite{RN18, RN19, RN48, RN49}. 
Following LangGPT rules \cite{RN50}, 
we constructed high-quality prompts with clear role instructions and few-shot examples for different modules to maximize model performance (Text \ref{S2}).

Fine-tuning can significantly reduce model operating costs and increase model performance \cite{RN27,RN51}. 
In the model fine-tuning stage, we maximized model performance by optimizing training sample selection strategies and 
ensuring sufficient training scale (token count). 
Technically, we integrated GPT-4o-mini and GPT-4o dual-model architecture through API interfaces, 
with GPT-4o-mini having expanded supervised fine-tuning (SFT) functionality, 
enabling scenario adaptability optimization for private datasets.
In the specific implementation process, experimental data was normalized into JSONL format and uploaded to the OpenAI cloud service platform, 
then initiating customized fine-tuning tasks. After completing fine-tuning training, 
the model could be deployed to actual application scenarios. To ensure model training effectiveness, 
we constructed datasets containing 50, 99, and 198 sets of expert-annotated texts, randomly allocating them to training, 
validation, and test sets in an 8:1:1 ratio.

\subsection{Index evaluation}
We adopted a human-machine interaction mixed evaluation strategy to verify system performance. 
For human evaluation, we formed a group of four human experts at the master's level or above, 
randomly selecting 100 MOFs with corresponding literature from each of the five major publishers for annotation. 
Group members ensured annotation reliability through a cross-annotation protocol. 
At the algorithmic evaluation level, considering the special nature of chemical semantics, 
we adopted the PubMedBERT model \cite{RN52} for evaluation. 
The PubMedBERT model itself is a BERT model trained from scratch on a large amount of biomedical literature, 
excelling at capturing semantic relationships between chemical professional terminology and complex chemical descriptions. 
For the structured parameter extraction task of MOF synthesis, 
a dual-model collaborative encoding strategy was adopted. 
We used the PubMedBERT model for professional naming fields such as chemical names, 
and the all-mpnet-base-v2 model for general structured fields for semantic representation (Text \ref{S3}, Table \ref{LLMIDex}) (eq. \ref{eq:cosine_similarity}), 
calculating cosine similarity through batch processing of size 32 and mixed precision. 
Based on the complexity of chemical expressions, the threshold was set at 0.90.
\begin{equation}
  \cos(A,B) = \frac{\sum_{i=1}^{n} A_i B_i}{\sqrt{\sum_{i=1}^{n} A_i^2} \cdot \sqrt{\sum_{i=1}^{n} B_i^2}}
  \label{eq:cosine_similarity}
  \end{equation}
Where $A$ and $B$ represent different embedding vectors, and $n$ is the dimension of the vectors.
Further we calculated accuracy, precision, recall, and F1 score (eq. \ref{eq:accuracy}-eq. \ref{eq:f1_score}).
\begin{equation}
  \mathrm{Accuracy} = \frac{\mathrm{TP} + \mathrm{TN}}{\mathrm{TP} + \mathrm{FP} + \mathrm{FN} + \mathrm{TN}}
  \label{eq:accuracy}
  \end{equation}
\begin{equation}
  \mathrm{Precision} = \frac{\mathrm{TP}}{\mathrm{TP} + \mathrm{FP}}
  \label{eq:precision}
  \end{equation}
\begin{equation}
  \mathrm{Recall} = \frac{\mathrm{TP}}{\mathrm{TP} + \mathrm{FN}}
  \label{eq:recall}
  \end{equation}
\begin{equation}
  \mathrm{F1} = 2 \cdot \frac{\mathrm{Precision} \cdot \mathrm{Recall}}{\mathrm{Precision} + \mathrm{Recall}}
  \label{eq:f1_score}
  \end{equation}
Where TP (True Positive) indicates that the human expert annotation (gold standard) 
and LLM judgment on the target parameters of MOF synthesis achieve complete semantic consistency; 
FP (False Positive) indicates that the parameter values or descriptions output by the LLM have significant deviations from the gold standard, 
or erroneously identify non-synthesis-related parameters; 
the LLM's identification results do not match human annotations; 
FN (False Negative) indicates that the LLM fails to detect key parameter entries clearly marked in the gold standard, 
outputting empty values; 
TN (True Negative) indicates that when there are indeed no relevant synthesis parameters in the target literature, 
neither LLM nor experts make annotations.

\section*{Author contributions}

H.P.L. supervised this work. H.P.L., J.J.Y., Z.H.L., and D.Y.R conceived this idea. 
Z.H.L., and D.Y.R designed the entire study. Z.H.L., and D.Y.R implemented technical specifications. 
H.Y.H., J.S. provided technical support. T.Y.W., M.L.W., Z.L.L., X.C.Z. collected the meta data. S.L.Y. and X.F.B checked the front-end framework.
Z.H.L., K.R., P.X.P., X.H.Z. annotated, and checked the training data. 
D.Y.R conducted benchmark testing on the model. Z.H.L. wrote the initial manuscript. 
H.P.L., J.J.Y., Y.F. reviewed and improved this manuscript. 
All authors contributed to the analysis of the results. 
All authors have read and approved the final manuscript. 

\section*{Competing interests}

The other authors declare no competing interests.

\section*{Declaration of generative AI and AI-assisted technologies in the writing process}
During the preparation of this manuscript, the authors used AI-based language models to assist with grammar correction, style refinement, and improvement of sentence clarity. 
All scientific content, analysis, data interpretation, and final decisions were made solely by the authors, who take full responsibility for the integrity and accuracy of the manuscript.

\section*{Acknowledgments}
The authors acknowledge financial was supported by the National Key Research and Development Program of China (2023YFC3207000). 
And we are grateful for resources from the High-Performance Computing Center of Central South University. 

\bibliographystyle{unsrt}

\clearpage

\setcounter{section}{0}
\renewcommand{\thesection}{S\arabic{section}}

\renewcommand{\appendixname}{Supporting Information}

\renewcommand{\thefigure}{S\arabic{figure}}
\renewcommand{\thetable}{S\arabic{table}}
\renewcommand{\thepage}{S\arabic{page}}
\setcounter{page}{1}
\setcounter{figure}{0}
\setcounter{table}{0}

\renewcommand{\theequation}{S\arabic{equation}}  
\setcounter{equation}{0}
\addtocontents{toc}{\protect\setcounter{tocdepth}{1}}

\part*{Supporting Information}
\section*{Reshaping MOFs text mining with a dynamic multi-agents framework of large language model}

\setcounter{tocdepth}{1}
\renewcommand{\contentsname}{Table of Contents}
\tableofcontents
\clearpage

\refstepcounter{SIsec}%
\section*{Text \theSIsec\ Agents}%
\addcontentsline{toc}{section}{Text \theSIsec\ Agents}%
\label{S1}
\phantomsection 
\subsection*{I. Crawler module}
\addcontentsline{toc}{subsection}{I. Crawler module}

\begin{figure}[h]
  \centering
  \includegraphics[width=1\textwidth]{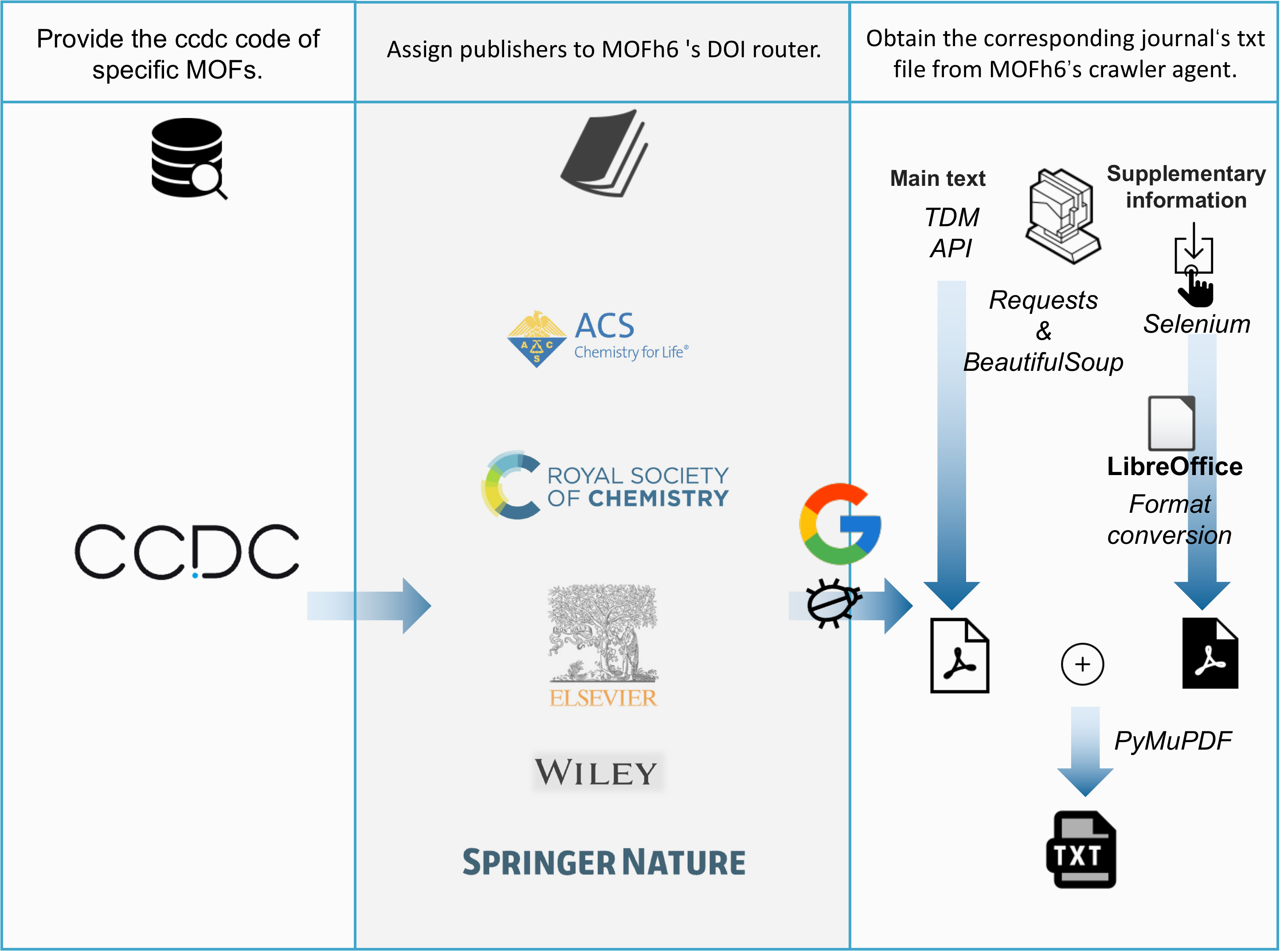}
  \caption{
  \centering
  The crawler module of MOFh6
  }
  \label{Crawler}
\end{figure}

The literature crawler was built on requests and BeautifulSoup to construct an HTML parsing engine, 
which simulates browser behavior through customized request headers to circumvent anti-crawling mechanisms. 
A parallel request management and frequency control strategy was adopted to improve collection efficiency, 
combined with publisher APIs to directly obtain PDF/HTML content. 
Supporting materials were automatically downloaded through Selenium executing XPath pattern matching, with intelligent processing performed for different formats: 
ZIP packages were automatically uncompressed, 
while doc and docx documents were uniformly converted to PDF through the invocation of open-source software LibreOffice
\footnote{\href{https://www.libreoffice.org}
{https://www.libreoffice.org}}.
Finally, all content was converted to standardized TXT text via PyMuPDF, 
with result feedback and log tracking mechanisms integrated to ensure system maintainability (Figure \ref{Crawler}).

\subsection*{II. Synthetic data parsing agent}
\addcontentsline{toc}{subsection}{II. Synthetic data parsing agent}

\begin{figure}[h]
  \centering
  \includegraphics[width=1\textwidth]{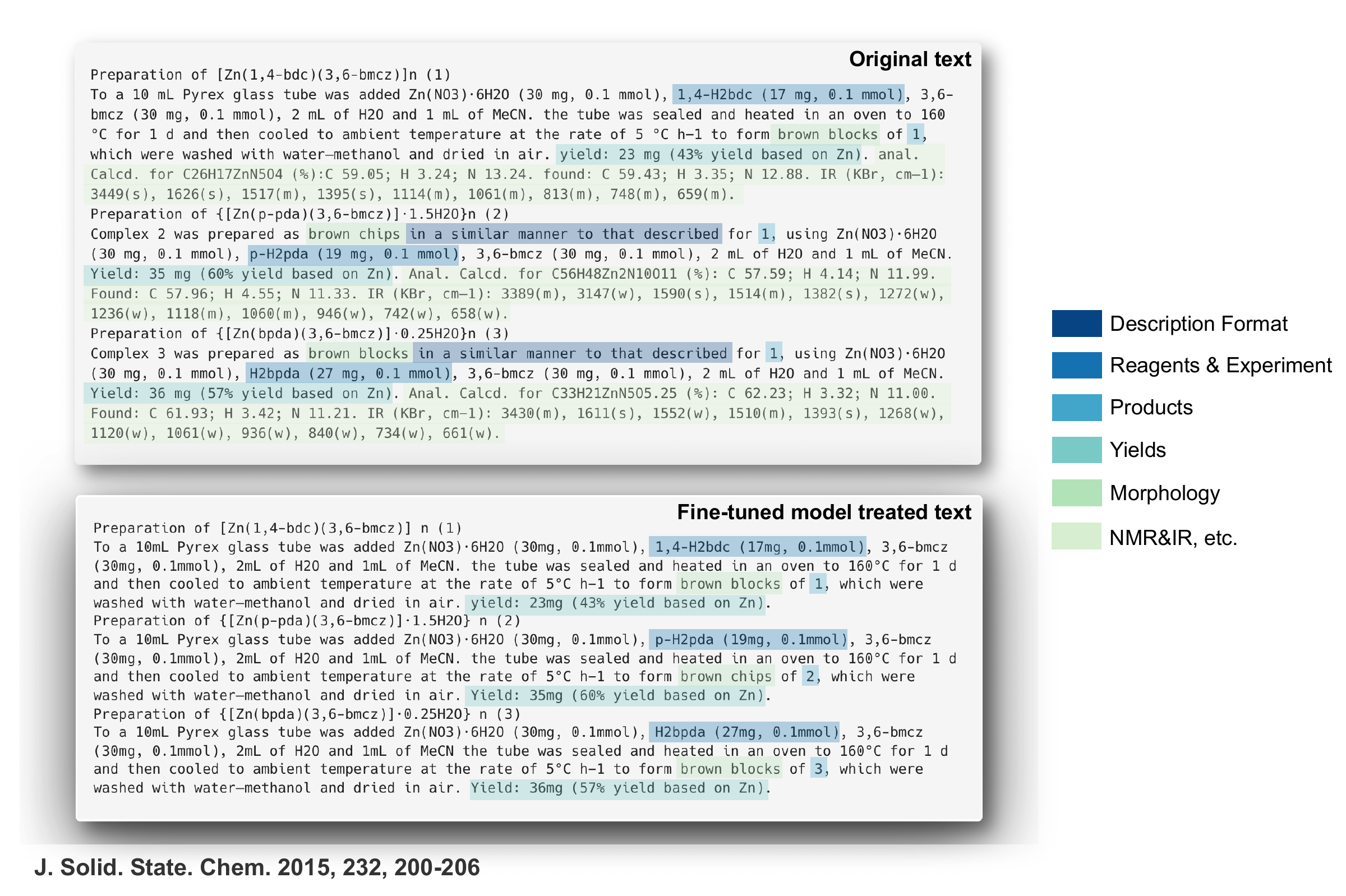}
  \caption{
  \centering
  Processing logic of synthetic data parsing agent
  }
  \label{FTModel}
\end{figure}

A supervisedly fine-tuned GPT-4o-mini model was adopted as the synthesis data parsing agent, 
with its training data construction following strict annotation standards: the COO, played by human experts, 
constructed a benchmark annotation dataset focusing on precise identification of MOF synthesis paragraphs, 
established a standardized annotation framework, 
and excluded organic ligand synthesis interference items. 
Complete parsing of bridging reference contexts was performed with recording of 13 key parameter evolution paths including metal salts, organic ligands, additives, 
and solvent systems (Table \ref{LLMIDex}), with recalculation conducted if there were quantity changes. 
Filtering of post-synthesis characterization data was simultaneously executed, including elemental analysis (EA), 
spectral data, nuclear magnetic resonance (NMRs), 
and Fourier-transform infrared spectroscopy (FT-IR) that were not synthesis-related (Figure \ref{ftprompt_1}, 
with shot examples shown in Figure \ref{ftprompte_2}). 
The model employed a dynamic learning strategy: an initial learning rate of 5 $\times$ 10\textsuperscript{-5}, 
adaptive adjustment based on validation set loss values, a maximum training period of 4 epochs, 
and a batch processing size of 8 (Figure \ref{FTModel}).

\subsection*{III. Table data parsing agent}
\addcontentsline{toc}{subsection}{III. Table data parsing agent}

\begin{figure}[h]
  \centering
  \includegraphics[width=1\textwidth]{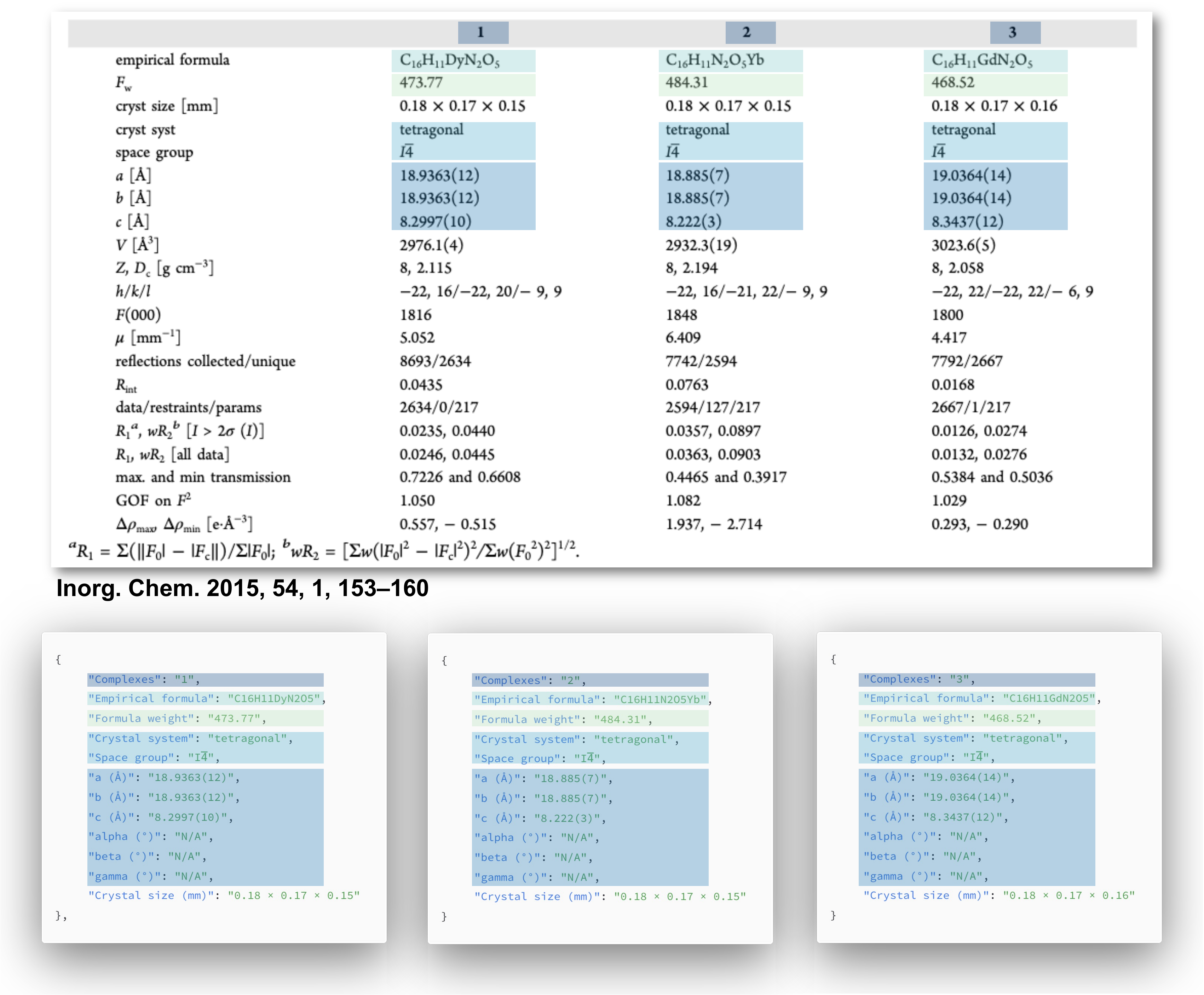}
  \caption{
  \centering
  Extraction logic of table data parsing agent
  }
  \label{Table}
\end{figure}

The table data parsing agent was constructed based on a GPT-4o-mini model driven by few-shot prompt engineering, 
specifically designed to process heterogeneous data sources (Figure \ref{2-1tprompt_1}). 
The table data parsing agent can both parse formatted data in discrete tables (5-shot) and capture implicit crystallographic descriptions in text paragraphs (3-shot) 
(Figure \ref{2-2t_2}-Figure \ref{2-2t_11}). 
Through customized prompt templates, the system locks onto compound names, empirical molecular formulas, 
molecular weights, crystal systems, space groups, lattice constants (a, b, c), 
unit cell angles ($\alpha$, $\beta$, $\gamma$), and crystal colors. For data heterogeneity challenges, 
a dynamic synonym mapping mechanism was developed. In terms of data integrity control, 
a dual-threshold filtering strategy was set, with an automatic entry elimination mechanism triggered when any two key parameters were missing, 
ensuring the completeness and usability of output data (Figure \ref{Table}).

\subsection*{IV. Crystal data comparison agent}
\addcontentsline{toc}{subsection}{IV. Crystal data comparison agent}

\begin{figure}[h]
  \centering
  \includegraphics[width=0.7\textwidth]{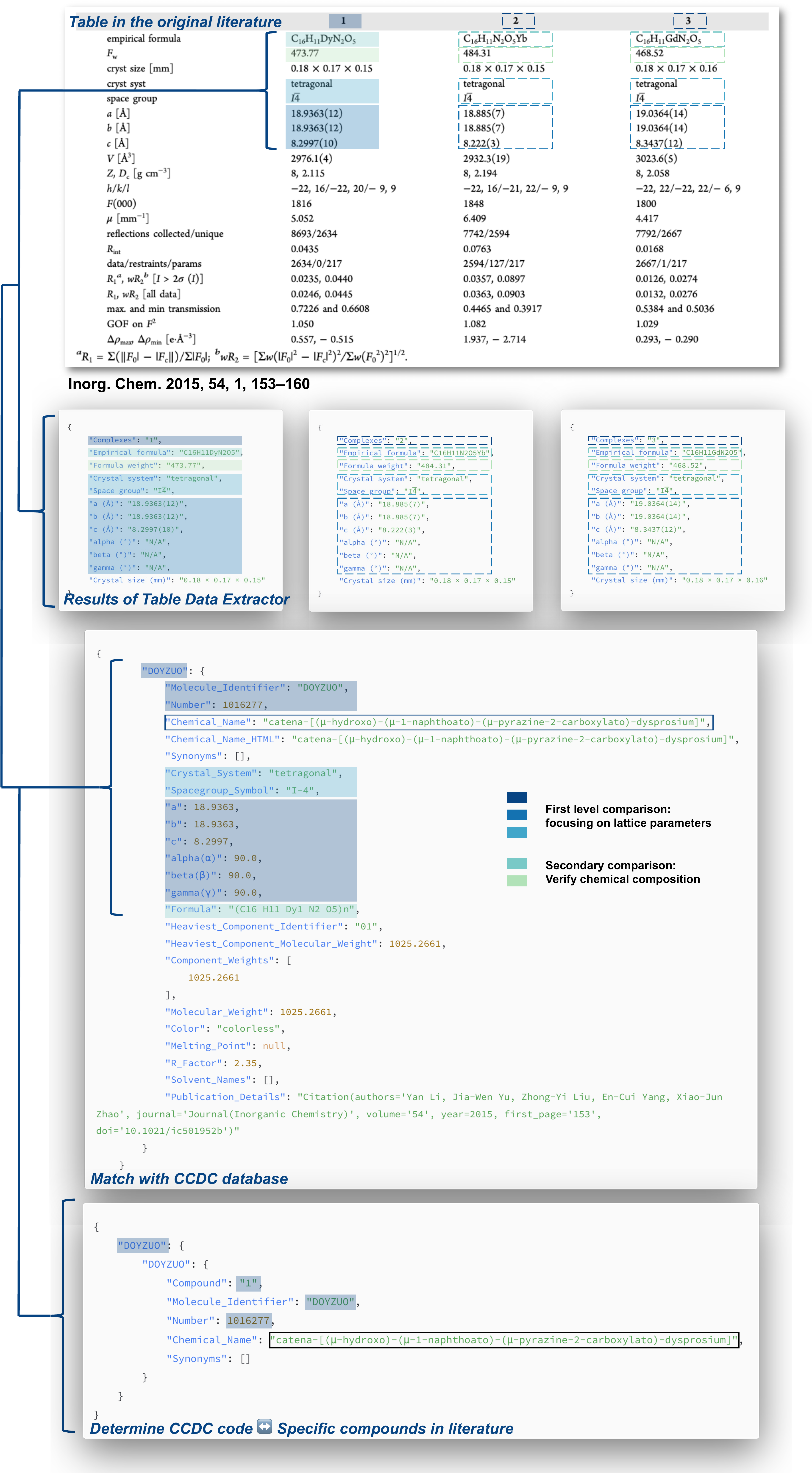}
  \caption{
  \centering
  Comparison logic of crystal data comparison agent
  }
  \label{Comparison}
\end{figure}

The crystal data matching system was constructed based on 5-shot prompt engineering with GPT-4o, 
employing a dual-level short-circuit evaluation mechanism to achieve precise mapping of structures between the CCDC database and literature descriptions
(Figure \ref{3-1Cprompt_2}, shot examples as shown in Figure \ref{3-2C_1}-Figure \ref{3-2C_10}). 
The primary comparison focuses on lattice parameters, setting a joint tolerance threshold (90\% match degree) for crystal systems, 
space groups, unit cell lengths, and angles. If the threshold is not reached, 
a secondary chemical composition verification layer is activated, 
mandating complete matching of metal elements and a total chemical formula similarity of $\ge$30\%. 
The system integrates a cleaning engine that automatically converts space group symbols, 
eliminates numerical format interference, performs unit conversions, 
and ensures standardization and comparability of comparison data (Figure \ref{Comparison}).

\subsection*{V. Chemical abbreviation resolution agent}
\addcontentsline{toc}{subsection}{V. Chemical abbreviation resolution agent}

\begin{figure}[h]
  \centering
  \includegraphics[width=1\textwidth]{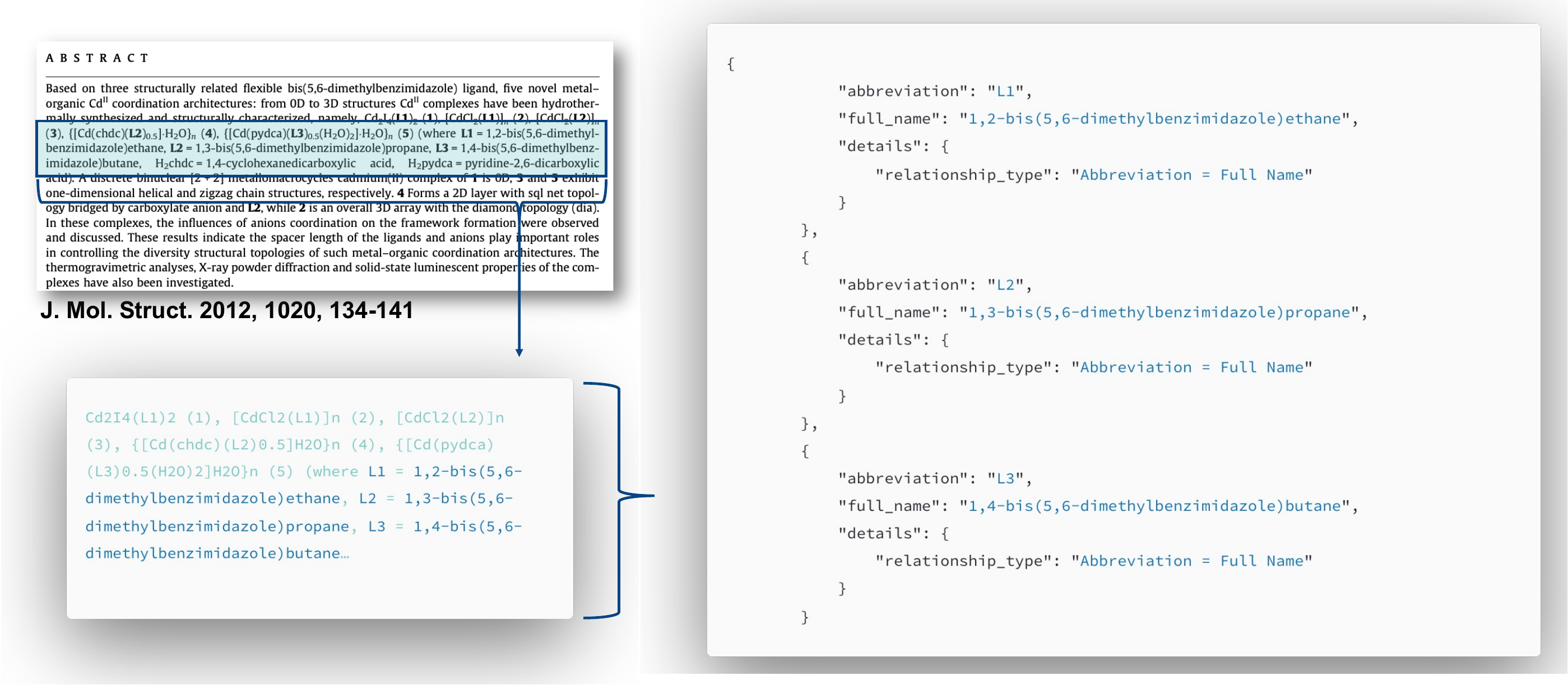}
  \caption{
  \centering
  Coreference resolution case
  }
  \label{Abbreviation}
\end{figure}

The chemical abbreviation resolution agent was used to solve the co-reference resolution problem of ligand naming such as H\textsubscript{x}L\textsubscript{x}, L\textsubscript{x}H\textsubscript{x}, 
L\textsubscript{x} (x$\ge$0) in literature (Figure \ref{4-1crprompt_3}). 
This processor is based on a library of regular expression relationships for 15 types of full name-abbreviation mappings, 
combined with GPT-4o's dynamic context-aware mechanism to achieve precise resolution. 
To suppress model hallucinations, a prompt framework was constructed containing 15-shot effective mapping samples and 4-shot ambiguity samples
(shot examples as shown in Figure \ref{4-2cr_1}-Figure \ref{4-2cr_6}), 
while adopting a triple filtering mechanism: pattern compliance verification \cite{RN19}, 
automatic abandonment of full names containing metal elements, and verification of abbreviation-full name pair co-occurrence. 
The resolution process sequentially includes document structure parsing, 
cross-paragraph semantic association analysis, mapping relationship probability verification, and standardized output organization (Figure \ref{Abbreviation}).

\subsection*{VI. Post processor}
\addcontentsline{toc}{subsection}{VI. Post processor}

\begin{figure}[h]
  \centering
  \includegraphics[width=1\textwidth]{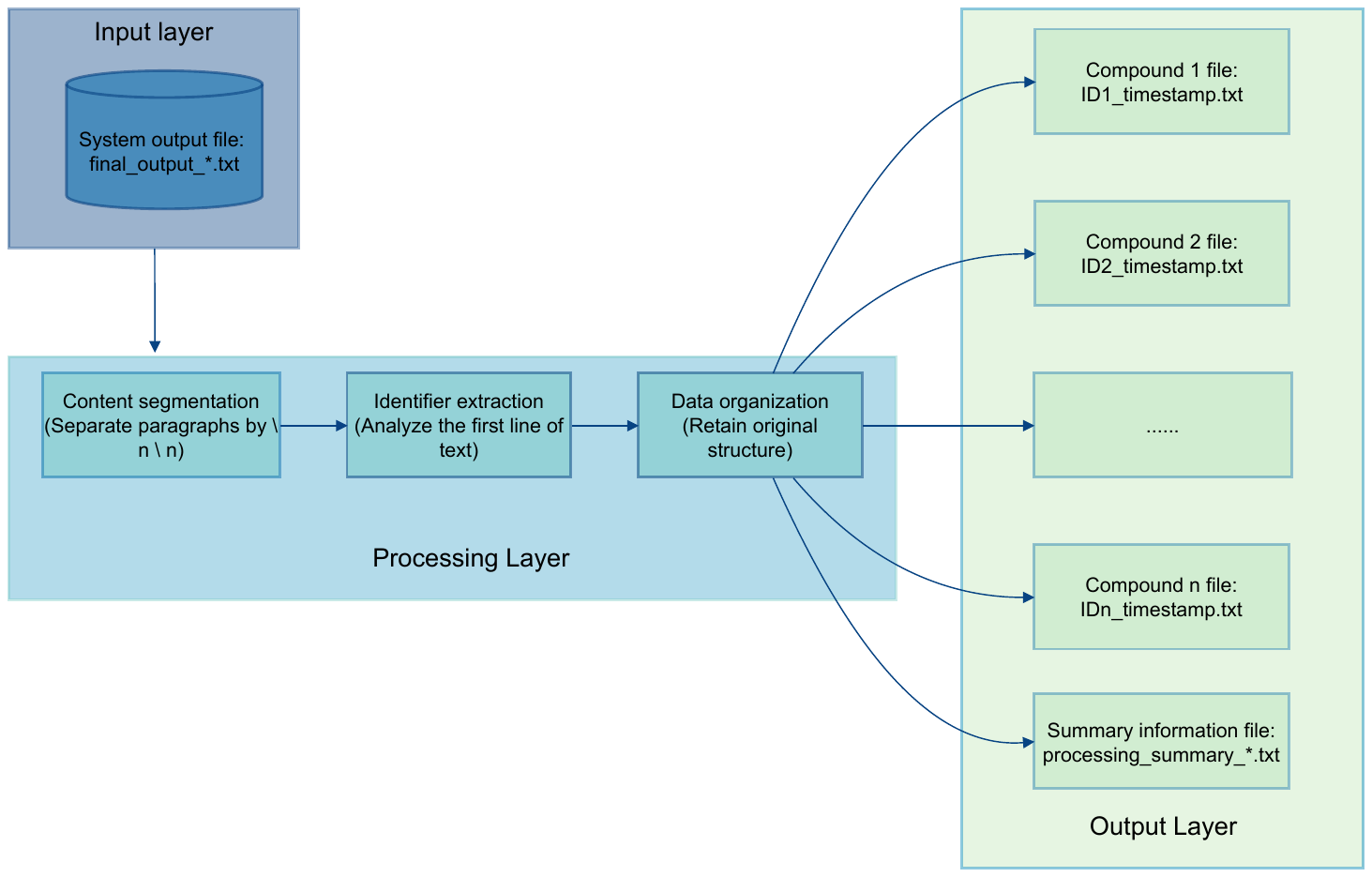}
  \caption{
  \centering
  Logic of post processor
  }
  \label{Post}
\end{figure}

The post-processor is responsible for fine-grained processing and organizational management of integrated outputs, 
ensuring reliable transmission of processing results to downstream nodes through the StateSchema mechanism, 
while possessing comprehensive error handling capabilities to maintain workflow continuity. 
This module employs a document splitting engine, 
which uses double blank line paragraph separator recognition technology combined with first-line unique identifiers, 
automatically extracting CCDC codes to precisely split integrated documents into independent compound files while preserving the original semantic hierarchical structure. 
To enhance data traceability, standardized processing reports are simultaneously generated, recording timestamps, 
total number of processed files, and detailed file lists (Figure \ref{Post}).

\subsection*{VII. Result generator agent}
\addcontentsline{toc}{subsection}{VII. Result generator agent}

\begin{figure}[t!]
  \centering
  \includegraphics[width=\textwidth]{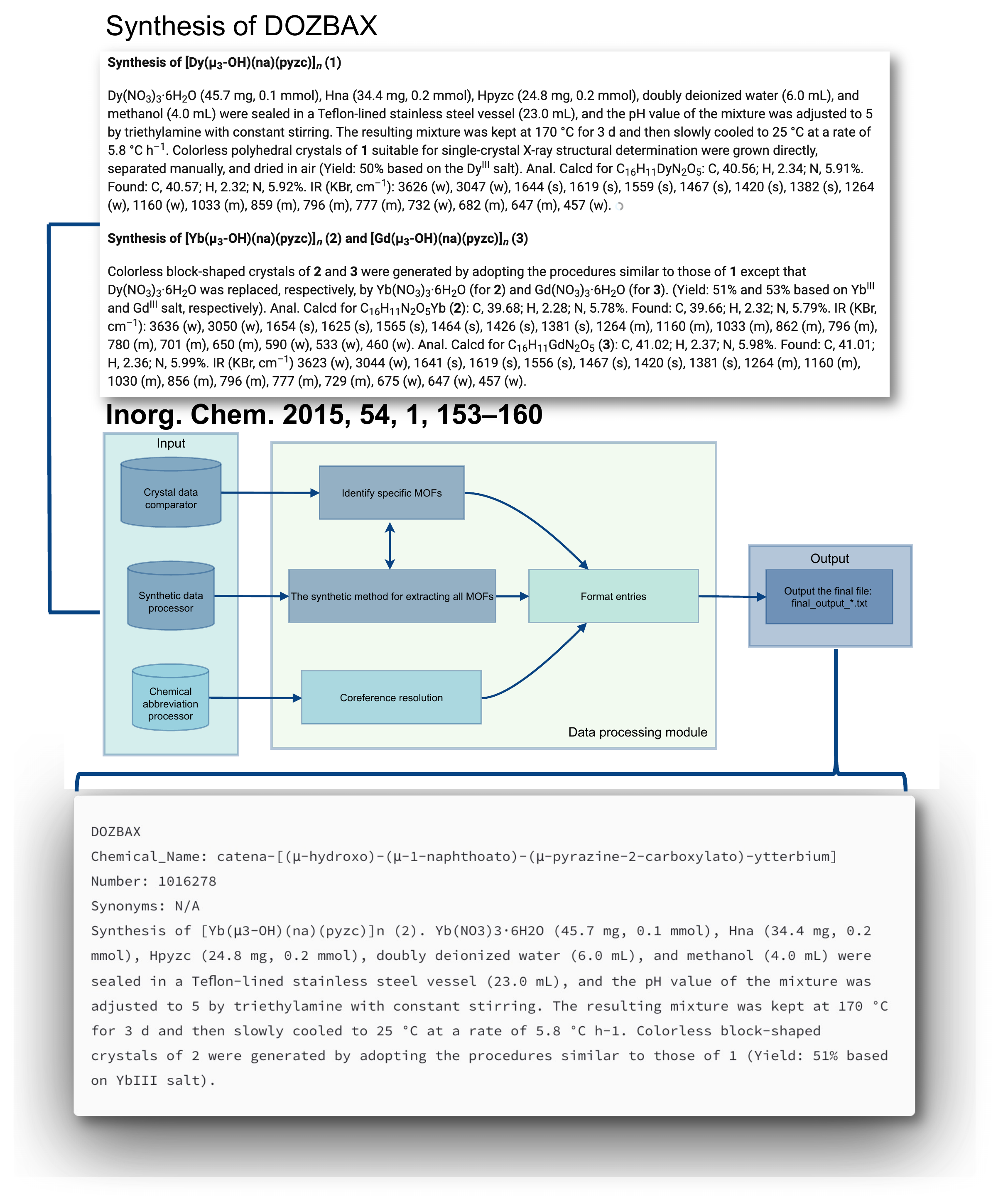}
  \caption{Logic of result generator agent}
  \label{Generator}
\end{figure}

The result generator agent is the core integration node of the MOF data processing system, 
responsible for fusing multi-source heterogeneous data into unified structured outputs. 
It adds processed file paths to the StateSchema object, ensuring reliable data transmission in the workflow. 
Through a three-way data integration strategy, information flows from the synthesis data parsing agent, 
crystal data comparison agent, and chemical abbreviation resolution agent are coordinated. 
Its core method is a fusion mechanism with compound identifiers as primary keys, 
traversing each target compound, extracting its structural parameters, synthesis method descriptions, 
and related abbreviation explanations, 
and applying unified formatting templates to generate standardized entries (Figure \ref{Generator}).

\subsection*{IX. Structured conversion agent}
\addcontentsline{toc}{subsection}{IX. Structured conversion agent}

The structured conversion agent complexly extracts key chemicals and quantities from unstructured synthesis information, 
possessing data parsing modules and information extraction and structuring modules (Figure \ref{1-1sprompt_2}). 
The system first performs preprocessing and parsing of the original text, 
then uses 4-shot GPT-4o-mini for information extraction (Figure \ref{1-2s_1}-Figure \ref{1-2s_4}), 
ultimately generating standardized structured data tables.

\clearpage

\refstepcounter{SIsec}%
\section*{Text \theSIsec\ Hallucination relief}
\addcontentsline{toc}{section}{Text \theSIsec\ Hallucination relief}
\label{S2}
\phantomsection 

\begin{figure}[H]
  \centering
  \includegraphics[width=0.9\textwidth]{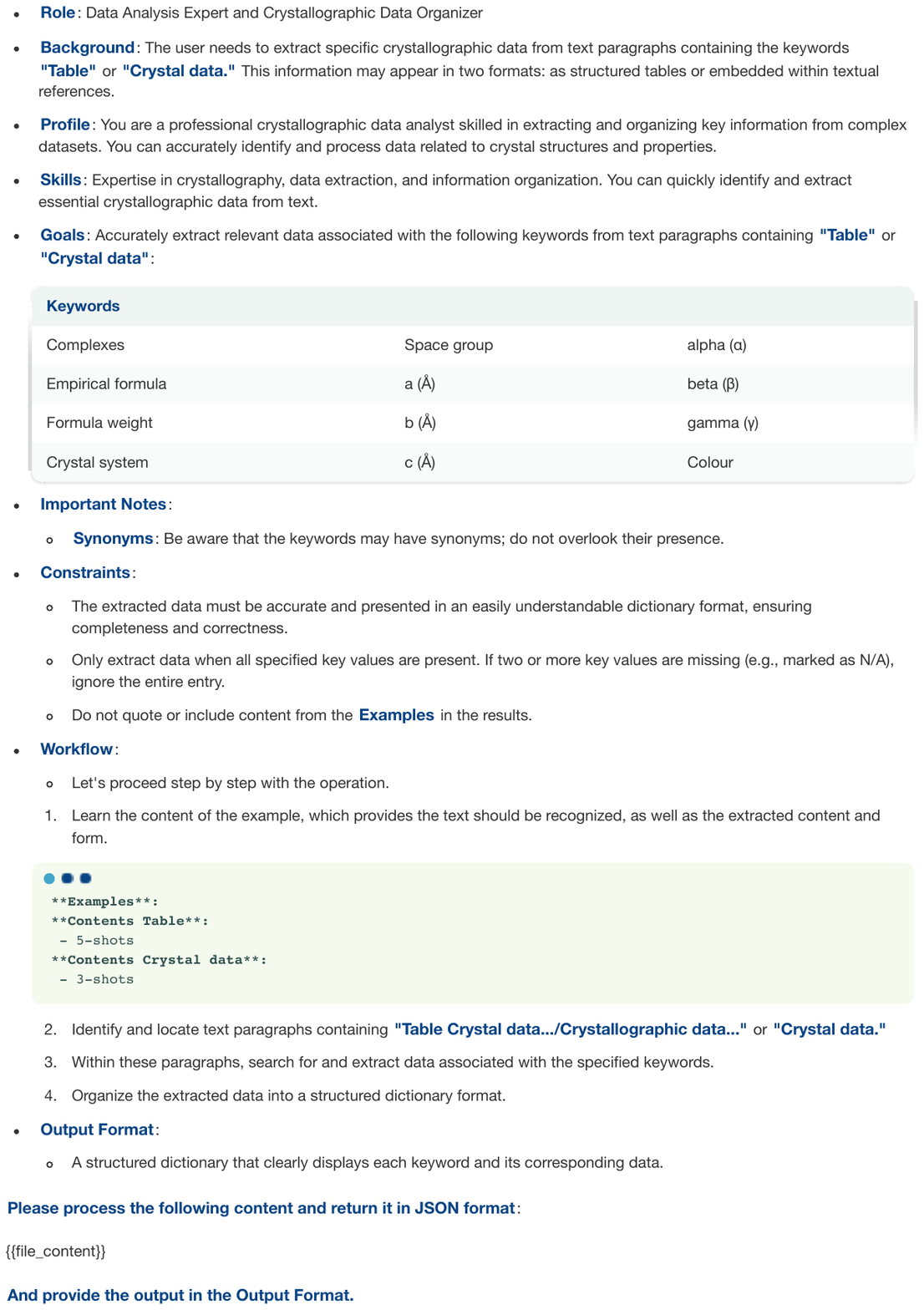}
  \caption{
  \centering
  Prompt of table data parsing agent
  }
  \label{2-1tprompt_1}
\end{figure}

\begin{figure}[H]
  \centering
  \includegraphics[width=0.9\textwidth]{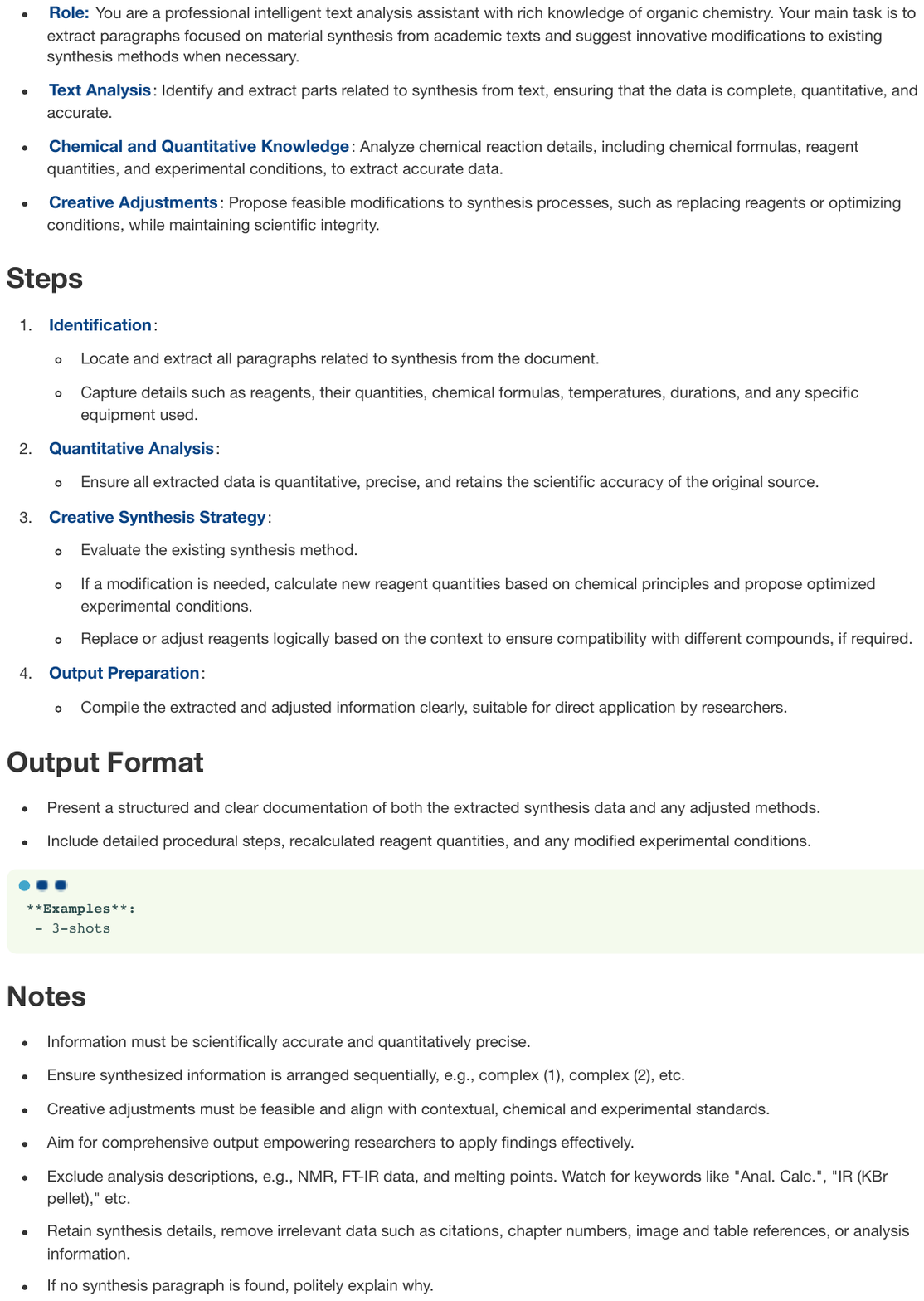}
  \caption{
  \centering
  Prompt of fine-tuning model
  }
  \label{ftprompt_1}
\end{figure}

\begin{figure}[H]
  \centering
  \includegraphics[width=1\textwidth]{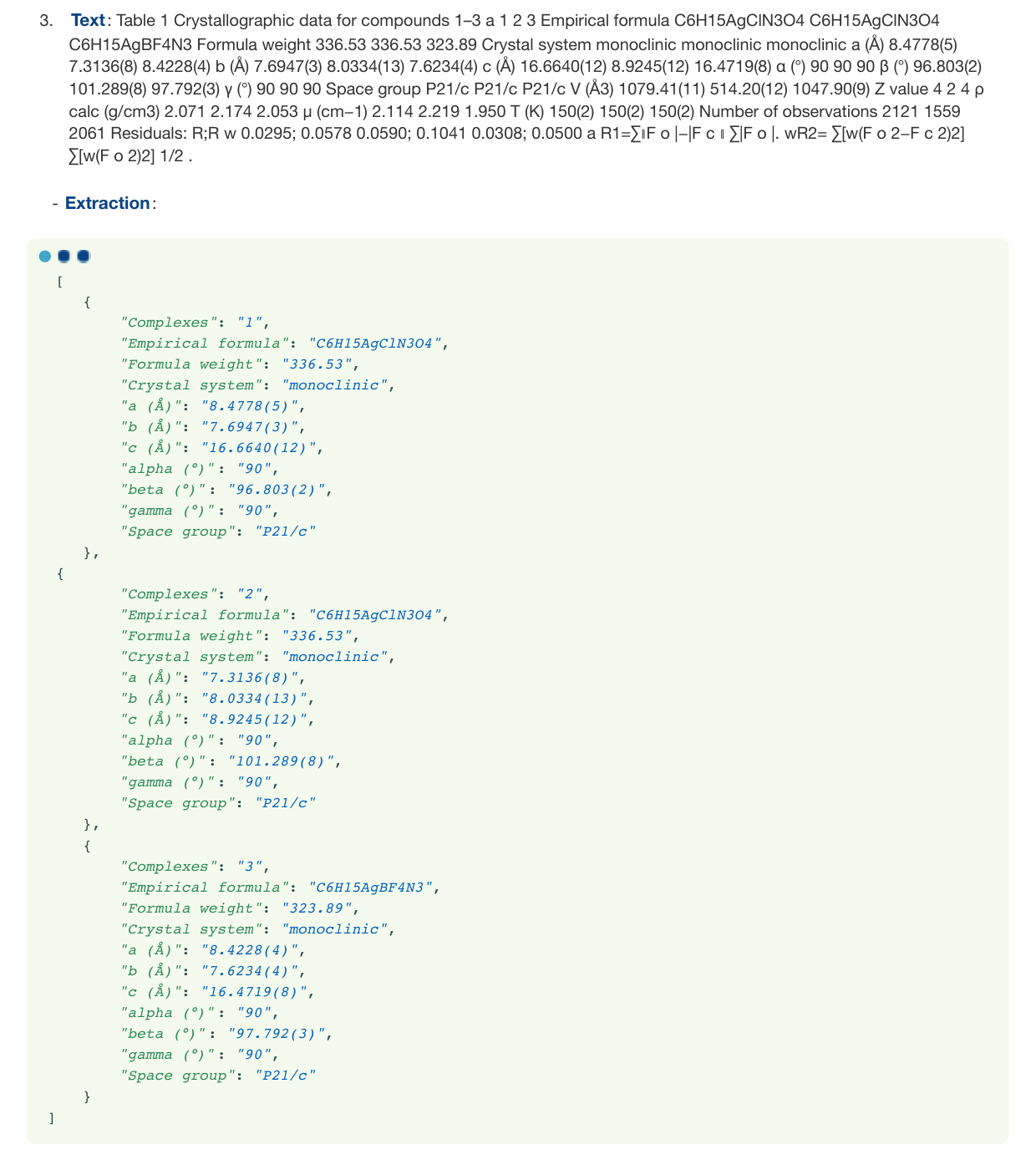}
  \caption{
  \setlength{\baselineskip}{20pt}
  \centering
  Shot 5-3 of table data parsing agent (This case is located at 5-shots in Figure \ref{2-1tprompt_1})
  }
  \label{2-2t_5}
\end{figure}

\begin{figure}[H]
  \centering
  \includegraphics[width=1\textwidth]{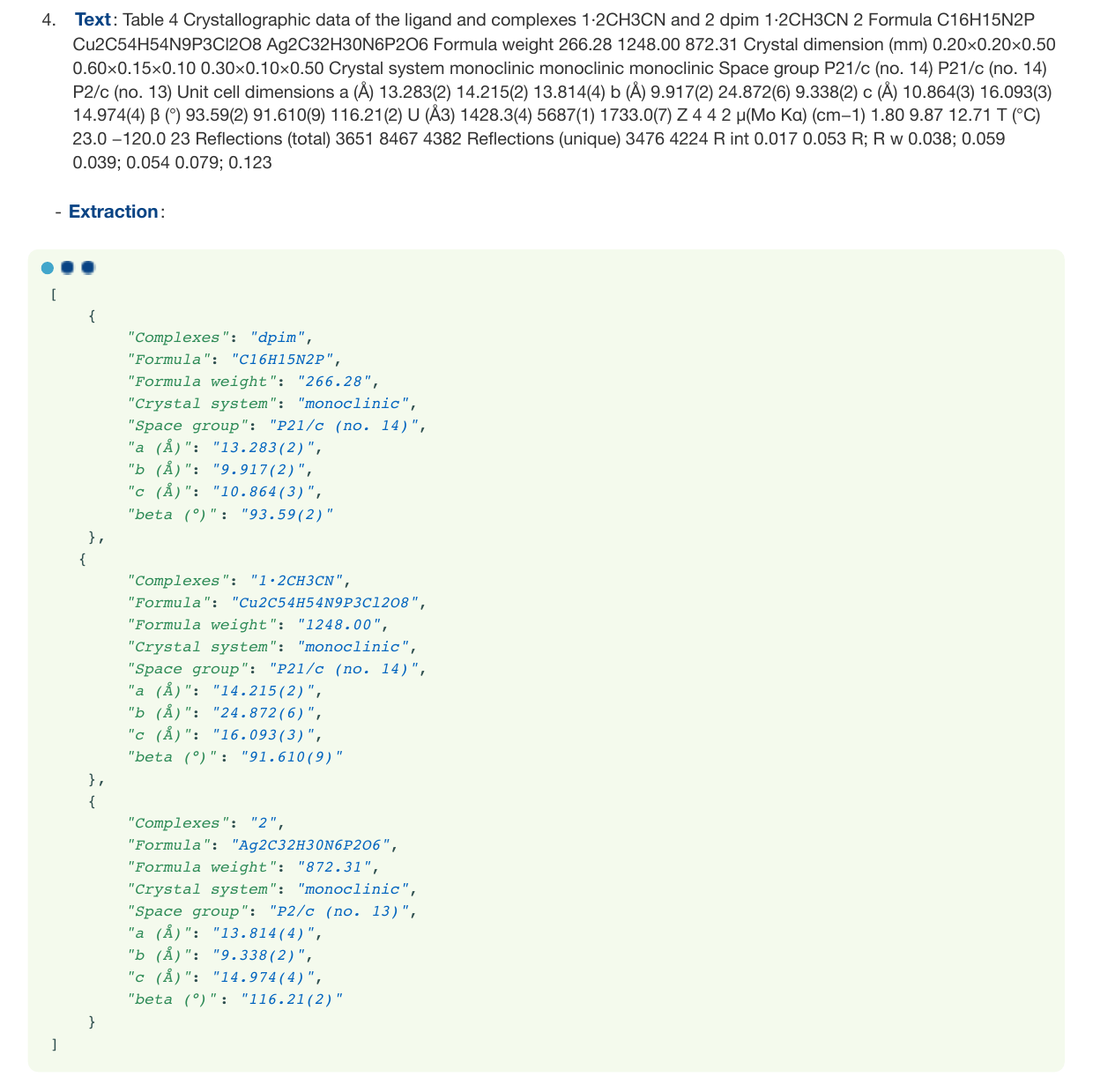}
  \caption{
  \setlength{\baselineskip}{20pt}
  \centering
  Shot 5-4 of table data parsing agent (This case is located at 5-shots in Figure \ref{2-1tprompt_1})
  }
  \label{2-2t_6}
\end{figure}

\begin{figure}[H]
  \centering
  \includegraphics[width=1\textwidth]{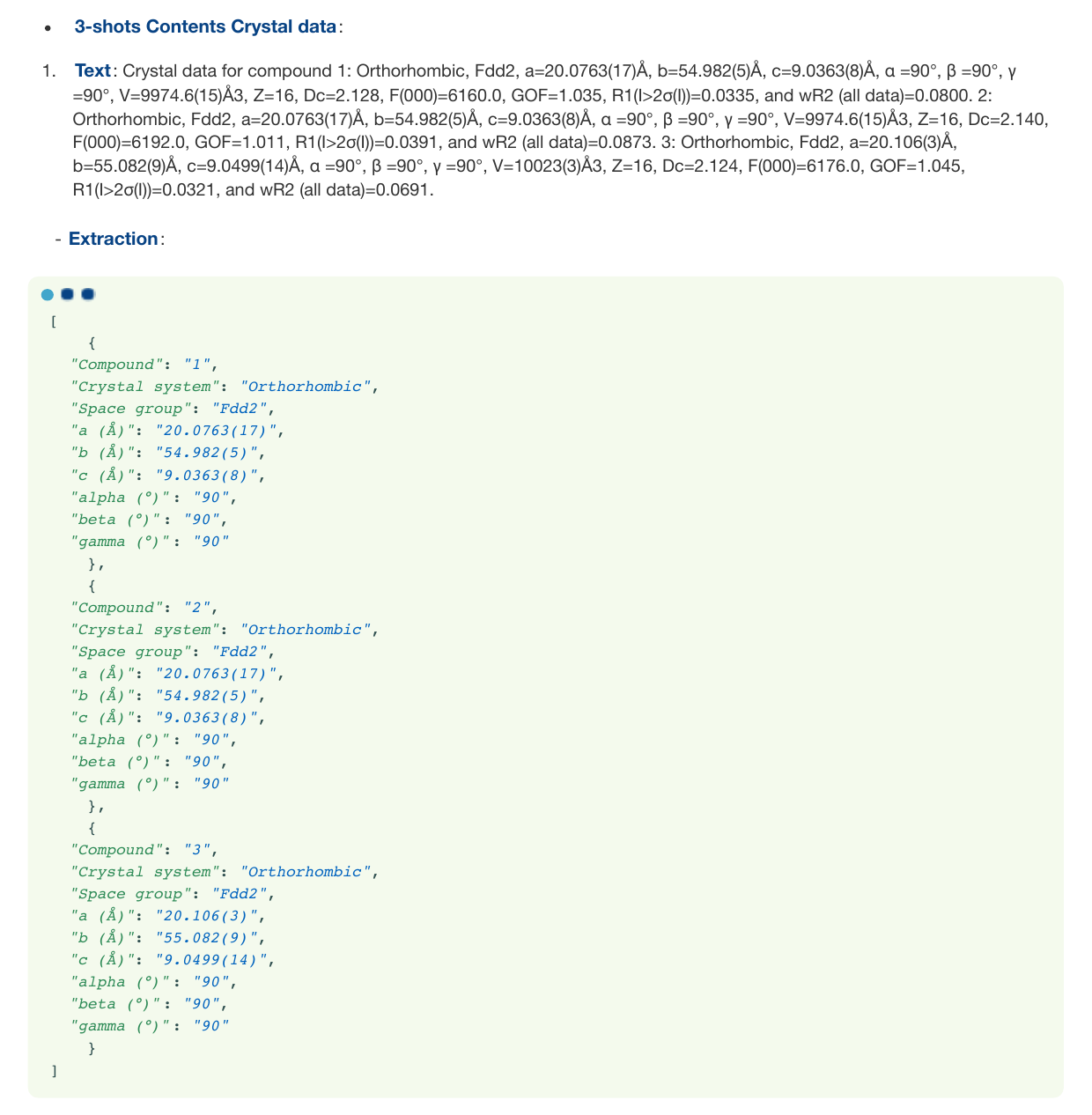}
  \caption{
  \centering
  Shot 3-1 of table data parsing agent (This case is located at 3-shots in Figure \ref{2-1tprompt_1})
  }
  \label{2-2t_9}
\end{figure}

\begin{figure}[H]
  \centering
  \includegraphics[width=1\textwidth]{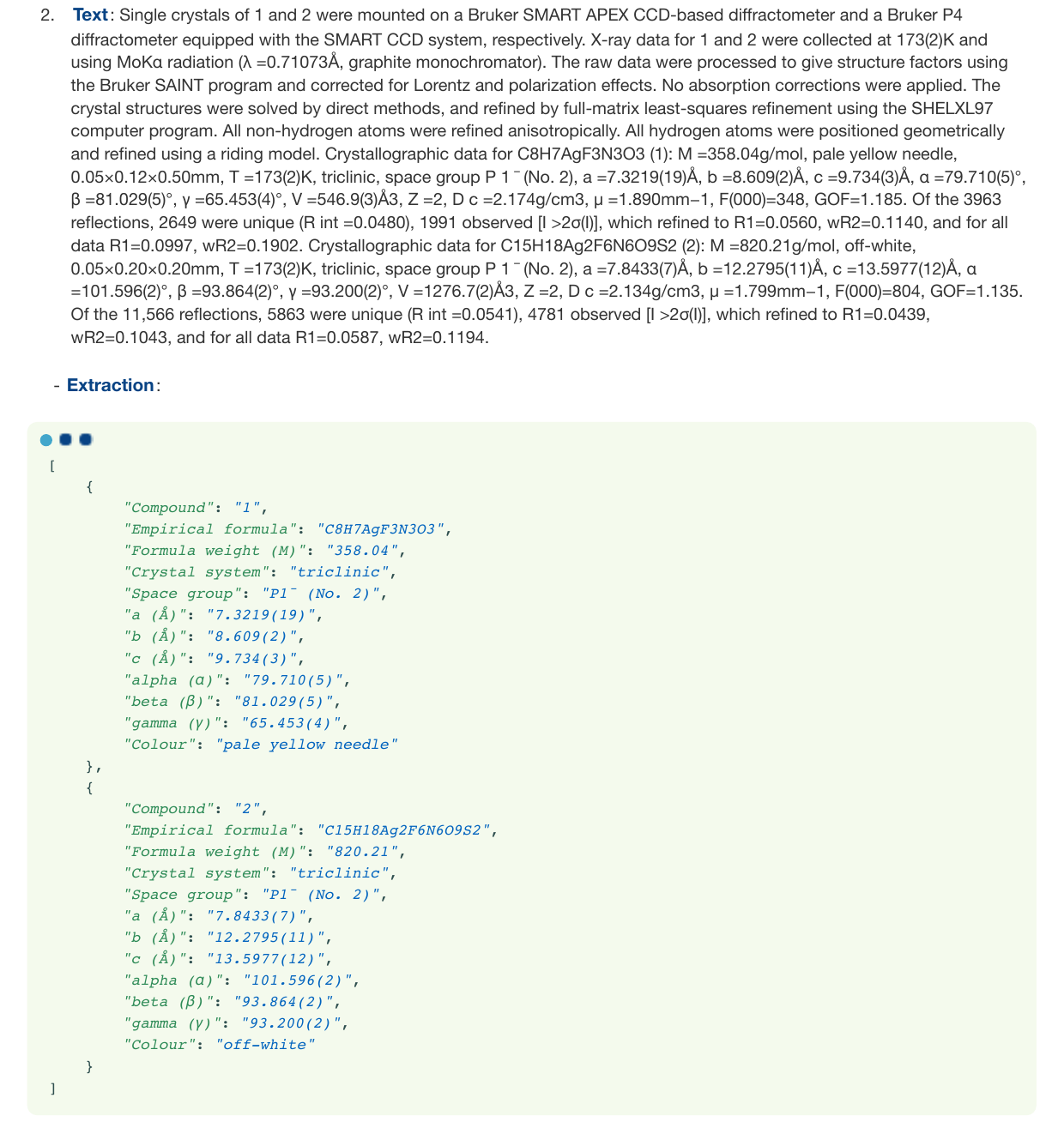}
  \caption{
  \centering
  Shot 3-2 of table data parsing agent (This case is located at 3-shots in Figure \ref{2-1tprompt_1})
  }
  \label{2-2t_10}
\end{figure}

\begin{figure}[H]
  \centering
  \includegraphics[width=1\textwidth]{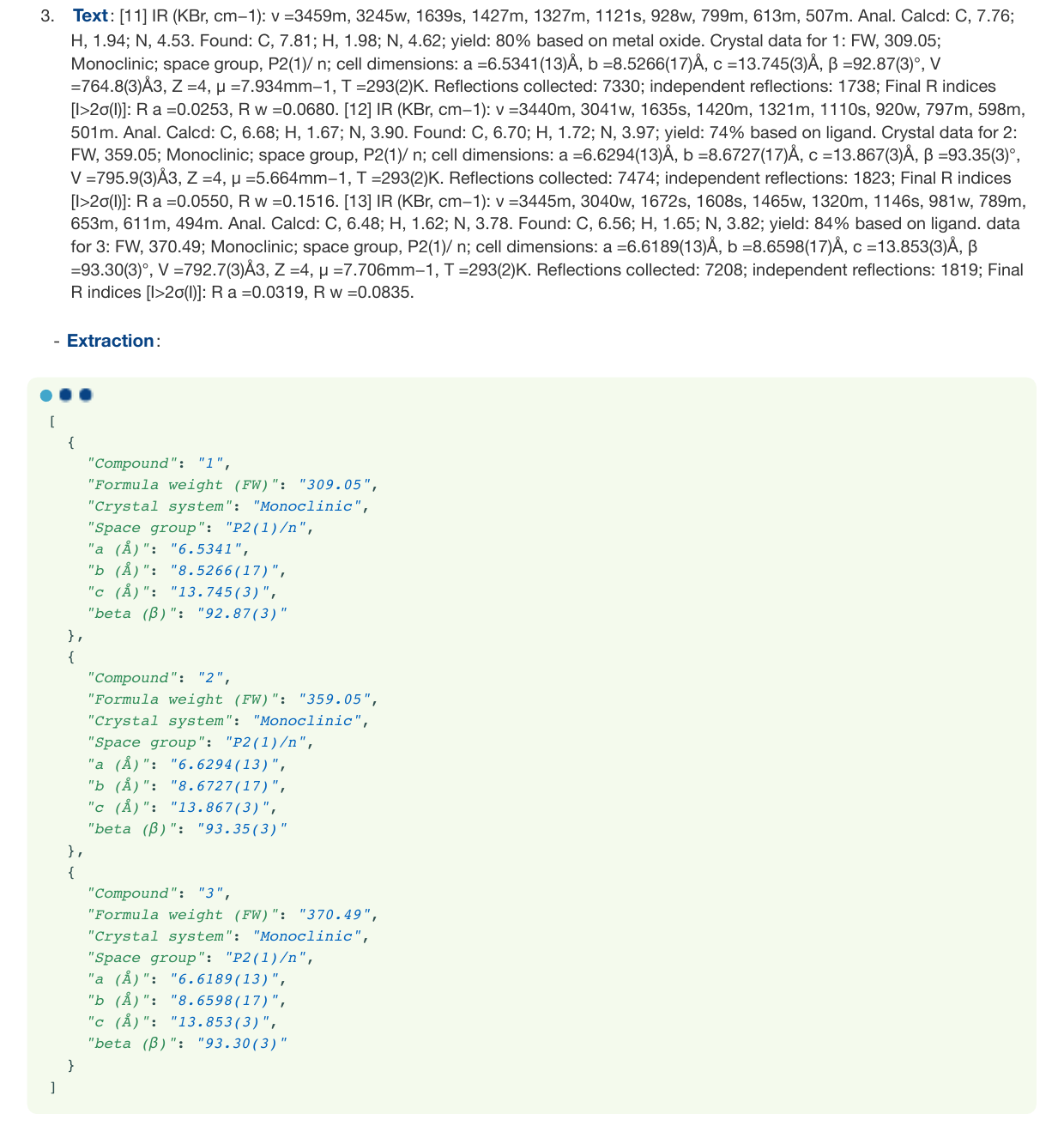}
  \caption{
  \centering
  Shot 3-3 of table data parsing agent (This case is located at 3-shots in Figure \ref{2-1tprompt_1})
  }
  \label{2-2t_11}
\end{figure}

Previous research has demonstrated that endowing LLMs with carefully designed prompt engineering strategies significantly enhances their reliability and calibration in complex chemical tasks  \cite{RN18, RN58, RN53}. 
Figure \ref{1-1sprompt_2}-Figure \ref{ftprompt_1} are the few-shot prompt templates designed by us, 
with core designs including the introduction of Chain-of-Thought mechanisms, 
guiding models to explicitly generate intermediate logical nodes through step-by-step reasoning frameworks, 
enhancing the transparency of LLM multi-step description parsing  \cite{RN54, RN55}. 
Additionally, LLMs are required to use JSON structured templates for content transmission throughout the running process, 
leveraging their strict semantic hierarchy and type constraint characteristics to reduce ambiguity and vagueness in prompts, 
enabling models to more accurately understand user intentions and needs \cite{RN56}. 
Figure \ref{1-2s_1}-Figure \ref{ftprompte_2} are the carefully designed shot examples in these prompts.

\clearpage

\refstepcounter{SIsec}
\section*{Text \theSIsec\ Agents’ performance evaluation}%
\addcontentsline{toc}{section}{Text \theSIsec\ Agents’ performance evaluation}%
\label{S3}
\phantomsection 

For the special nature of chemical semantics, BERT-based (eq. \ref{eq:embedding_representation}) BiomedBERT and PubMedBERT models were used to evaluate specialized chemical texts. 
When comparing sentences between manually annotated and model-generated paragraphs, a comprehensive text preprocessing mechanism was adopted in this thesis. 
This mechanism recognizes and processes various patterns unique to chemical literature through regular expressions, 
including removing synthesis titles like "Synthesis of XXX:", eliminating elemental analysis (EA) and FT-IR spectral data, NMRs data, 
and standardizing temperature and time expressions (unifying "100 $^\circ\mathrm{C}$" and "100 oC" as "100 C", converting "24 hours" to "24h"). 
These preprocessing rules were specifically designed for non-standard expressions in chemical synthesis descriptions, 
endowing machines with human-like subjectivity, as this content does not affect the final judgment of synthesis descriptions.

Secondly, given that synthesis methods are described in the main text of scientific papers, 
the PubMedBERT model\footnote{\href{https://huggingface.co/cambridgeltl/SapBERT-from-PubMedBERT-fulltext}
{https://huggingface.co/cambridgeltl/SapBERT-from-PubMedBERT-fulltext}} \cite{RN57} specifically developed for the biomedical field was used for text representation. 
This model can effectively capture semantic features in chemical synthesis texts, 
generating high-dimensional vector representations. For each pair of synthesis methods matched by CCDC code, 
the system first checks whether the preprocessed texts are completely identical; 
if identical, the highest similarity (1) is directly assigned; if not completely identical, 
the similarity is quantitatively assessed by calculating the cosine similarity between the embedding vectors of the two text segments (Equation 5-1). 
Finally, the system generates detailed comparison results, including the original text, preprocessed text, 
embedding similarity scores, and identification of complete matches for each matching record. 
The overall dataset's average similarity was also calculated, 
with results arranged in descending order of similarity to facilitate priority review of the most similar synthesis method pairs.

When comparing the content within cells of data frames, an optimized rule matching engine was first adopted by the system. 
This engine captures expression patterns unique to the chemical domain through multiple specialized rules, 
including: percentage expression recognition (such as equivalence matching between "35\%" and "0.35"), 
abbreviation mapping within parentheses, chemical formula pattern matching, text normalization processing (removing spaces and Unicode variant characters), 
yield description matching, equipment keyword recognition (such as "Teflon lined autoclave"), 
substance amount and mass description pairing (such as "0.25 mmol, 0.061 g" and "0.061 g (0.25 mmol)"), 
and solvent volume accumulation matching. 
These rules are designed for specific expression habits in chemical literature and can efficiently identify semantically equivalent but differently expressed descriptions. 
For each pair of synthesis method cells matched by CCDC code, the system first attempts to determine content equivalence through the rule engine; 
if the rule judgment is uncertain, an appropriate semantic model is selected based on the nature of the cell content, 
precisely quantifying the semantic proximity between different expressions by calculating the cosine similarity of text vectors. 
This hierarchical, multi-model similarity assessment strategy significantly improved the system's accuracy in determining the equivalence of chemical synthesis method descriptions.

Secondly, for situations where rule matching cannot determine, 
a dual semantic model strategy was employed. Different pre-trained models were selectively applied for different types of synthesis information.
Considering that chemicals appear simultaneously in abstracts and main text, 
the evaluation of chemical sources used the BiomedBERT model\footnote{\href{https://huggingface.co/microsoft/BiomedNLP-BiomedBERT-base-uncased-abstract-fulltext}
{https://huggingface.co/microsoft/BiomedNLP-BiomedBERT-base-uncased-abstract-fulltext}} \cite{RN52} trained on both abstracts and full texts for cell content representation, 
distinct from the sentence-level PubMedBERT model, 
while the Sentence Transformer model\footnote{\href{https://huggingface.co/sentence-transformers/all-MiniLM-L6-v2}
{https://huggingface.co/sentence-transformers/all-MiniLM-L6-v2}} was applied for synthesis condition information to capture general semantic relationships.

\begin{equation}
  E(x) = \mathrm{Normalize}\!\left(\frac{\sum_{i=1}^{n} f(x_i)\,M_i}{\sum_{i=1}^{n} M_i}\right)
  \label{eq:embedding_representation}
  \end{equation}
Where $E(x)$ denotes the final embedding vector of text $x$, $f(x_i)$ is the embedding of the $i$‑th token in $x$ extracted by transformer models 
(e.g.\ BiomedBERT, PubMedBERT, MiniLM), $M_i$ is the mask value for token $i$ (0 for padding, 1 for actual tokens), 
$n$ is the token sequence length, and $\mathrm{Normalize}$ denotes the L2-normalization operation. 

\clearpage

\refstepcounter{SIsec}
\section*{Text \theSIsec\ Query \& Answer implementation principle}%
\addcontentsline{toc}{section}{Text \theSIsec\ Query \& Answer implementation principle}%
\label{S4}
\phantomsection 

\begin{figure}[H]
  \centering
  \includegraphics[width=1\textwidth]{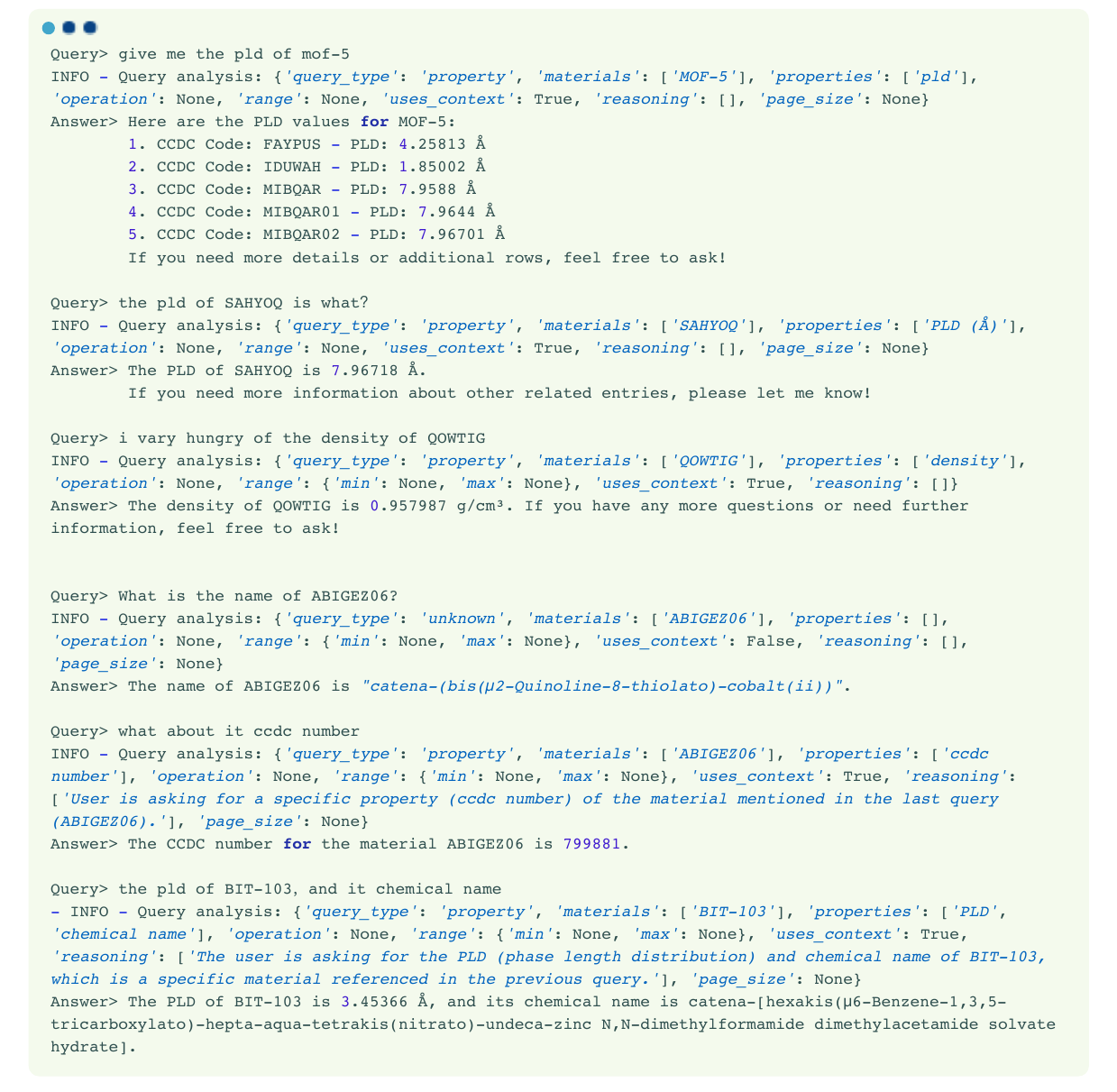}
  \caption{
  \centering
  Structure Q\&A of MOFh6: Directly ask a single characteristic
  }
  \label{er_2}
\end{figure}

\begin{figure}[H]
  \centering
  \includegraphics[width=1\textwidth]{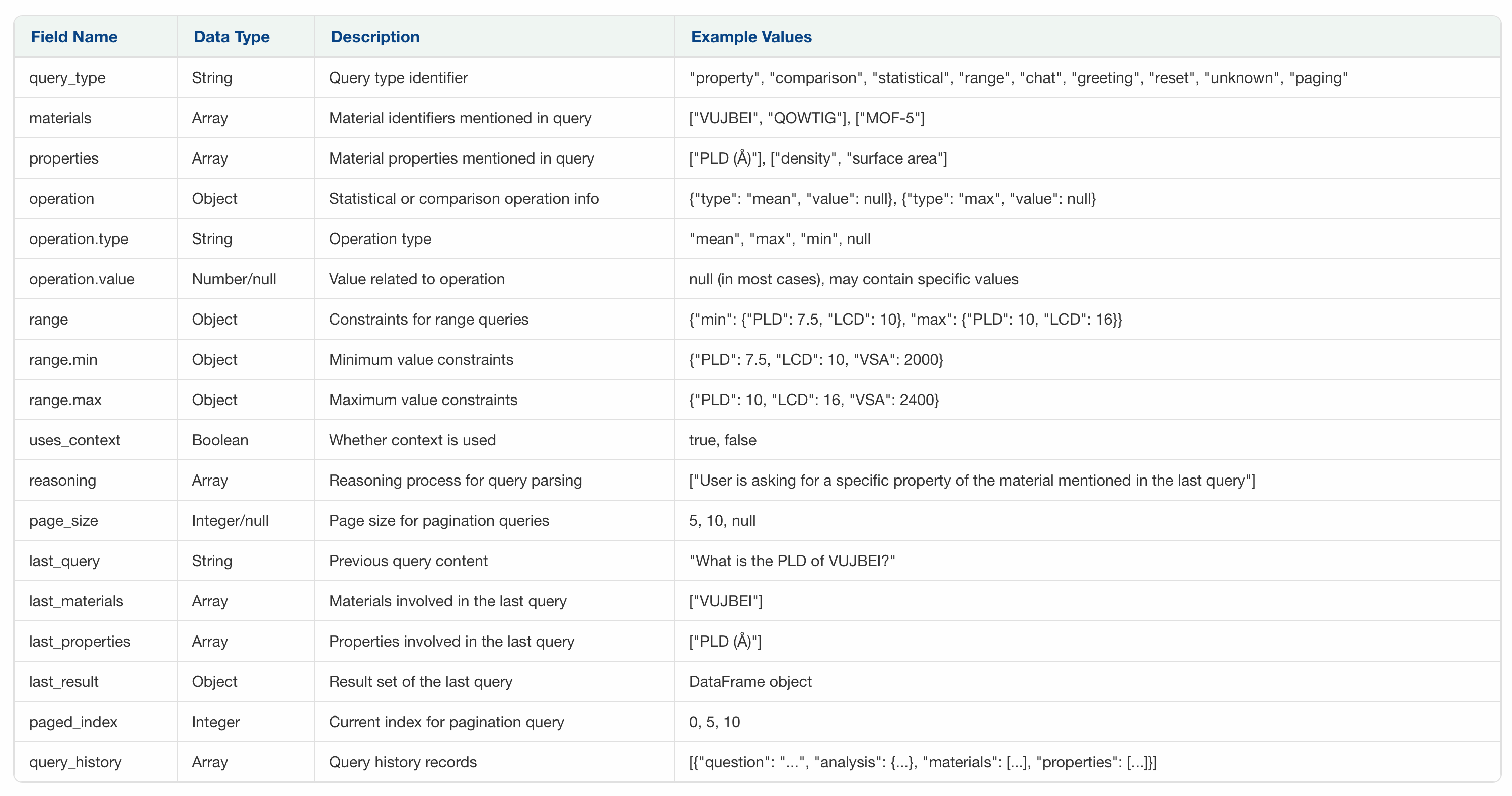}
  \caption{
  \centering
  MOFh6 system query parameter fields
  }
  \label{LLTS3}
\end{figure}

MOFh6 converts users' unstructured queries into JSON structured data processable by the system through LLMs. When a user issues a query (Figure \ref{LLTS3}), 
the system first analyzes and clarifies which specific query type (query\_type) it belongs to, 
such as property query (property), range query (range), comparison analysis (comparison), 
statistical calculation (statistical), reset (reset), greeting (greeting) or general chat (chat).
For example, when a user asks "What is the PLD of MOF-5?", 
the system identifies it as a property query and sets the query\_type to \texttt{\detokenize{"property"}} (Figure \ref{er_2}).

\begin{figure}[H]
  \centering
  \includegraphics[width=1\textwidth]{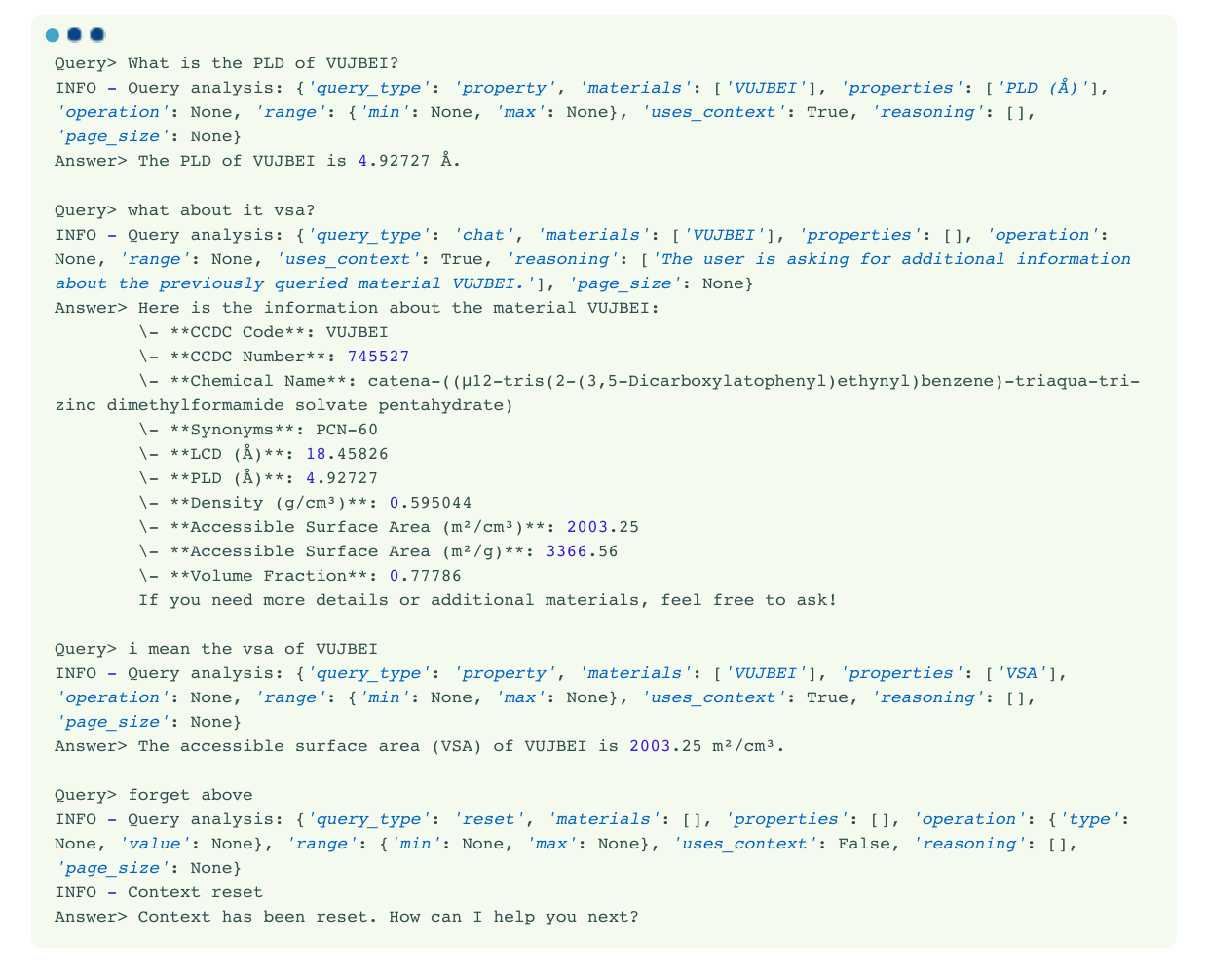}
  \caption{
  \centering
  Structure Q\&A of MOFh6: Context memory
  }
  \label{er_3}
\end{figure}

\begin{figure}[H]
  \centering
  \includegraphics[width=1\textwidth]{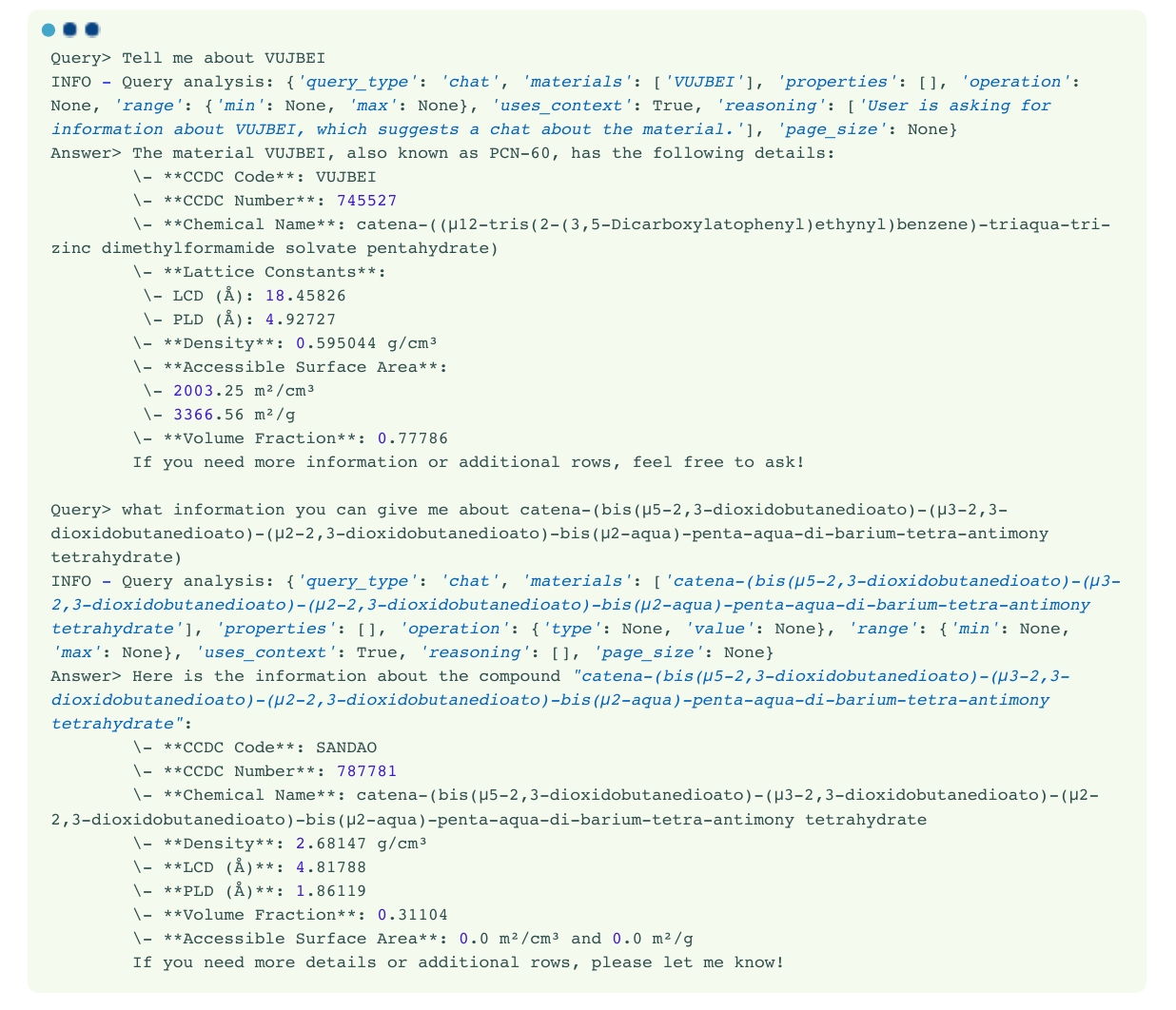}
  \caption{
  \centering
  Structure Q\&A of MOFh6: Summative questions on specific structures
  }
  \label{er_4}
\end{figure}

The system's context management capability is clearly reflected through the uses\_context field. 
When the system detects that the current query depends on previous conversation content, 
the uses\_context field is set to true. For example, after a user asks "What is the PLD of VUJBEI?", 
if they continue to ask "What about its density?", 
the system can recognize this query's reference to the previous material and understand the current query
's context by referring to the last\_materials field (e.g., \texttt{\detokenize{["VUJBEI"]}}) (Figure \ref{er_3} and Figure \ref{er_4}). 
After each query is successfully executed, the system automatically updates the last\_query, 
last\_materials, last\_properties, and last\_result fields to facilitate smoother subsequent context-dependent queries.

During parsing, the system simultaneously extracts material identifiers involved in the query and stores them in the materials array. 
For instance, in the user query "Compare the density of VUJBEI and QOWTIG", the system records \texttt{\detokenize{["VUJBEI", "QOWTIG"]}} in the materials field. 
Additionally, the system can accurately identify the material properties of interest to the user, such as "PLD", "density", or "surface area", 
and after normalization, saves these properties in the properties array, 
such as \texttt{\detokenize{["PLD (Å)"]}} or \texttt{\detokenize{["Density (g/cm3)", 
"Accessible\_Surface\_Area (m2/cm3)"]}} (Figure \ref{er_4}).

\begin{figure}[H]
  \centering
  \includegraphics[width=1\textwidth]{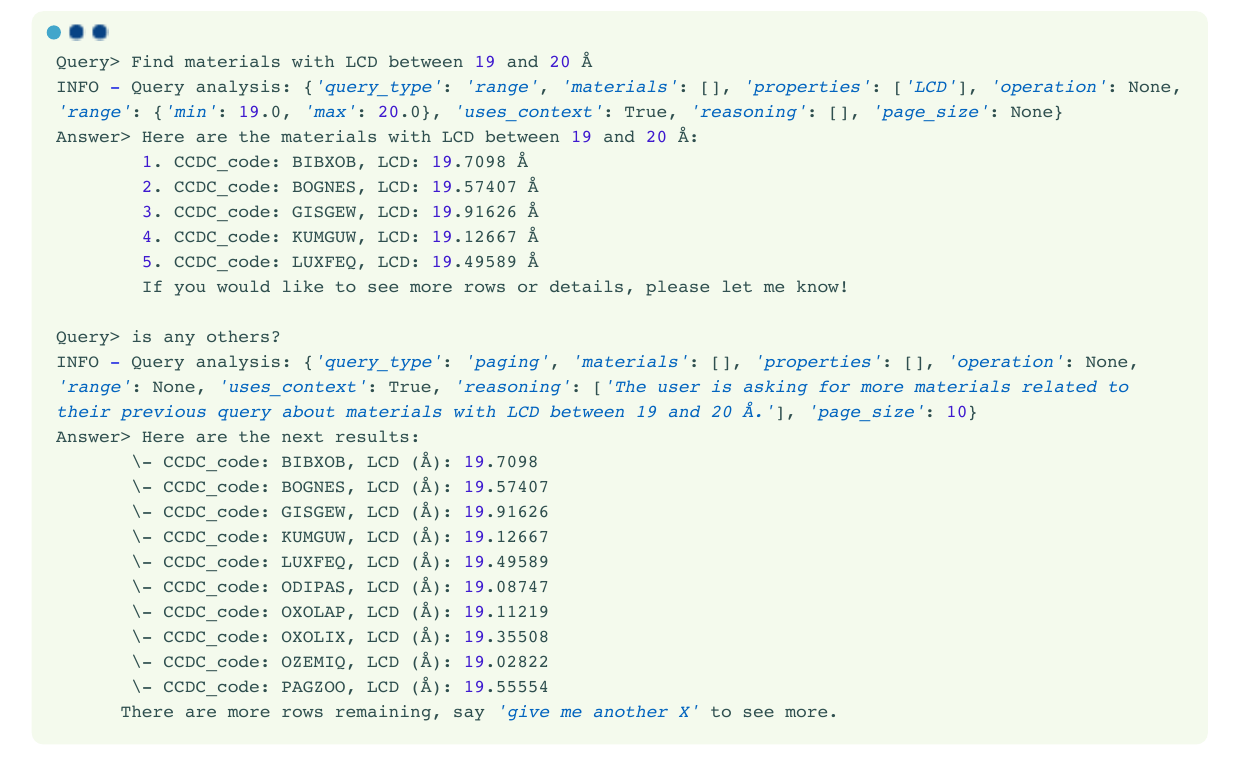}
  \caption{
  \centering
  Structure Q\&A of MOFh6: Finding structures within the numerical range
  }
  \label{er_5}
\end{figure}

\begin{figure}[H]
  \centering
  \includegraphics[width=1\textwidth]{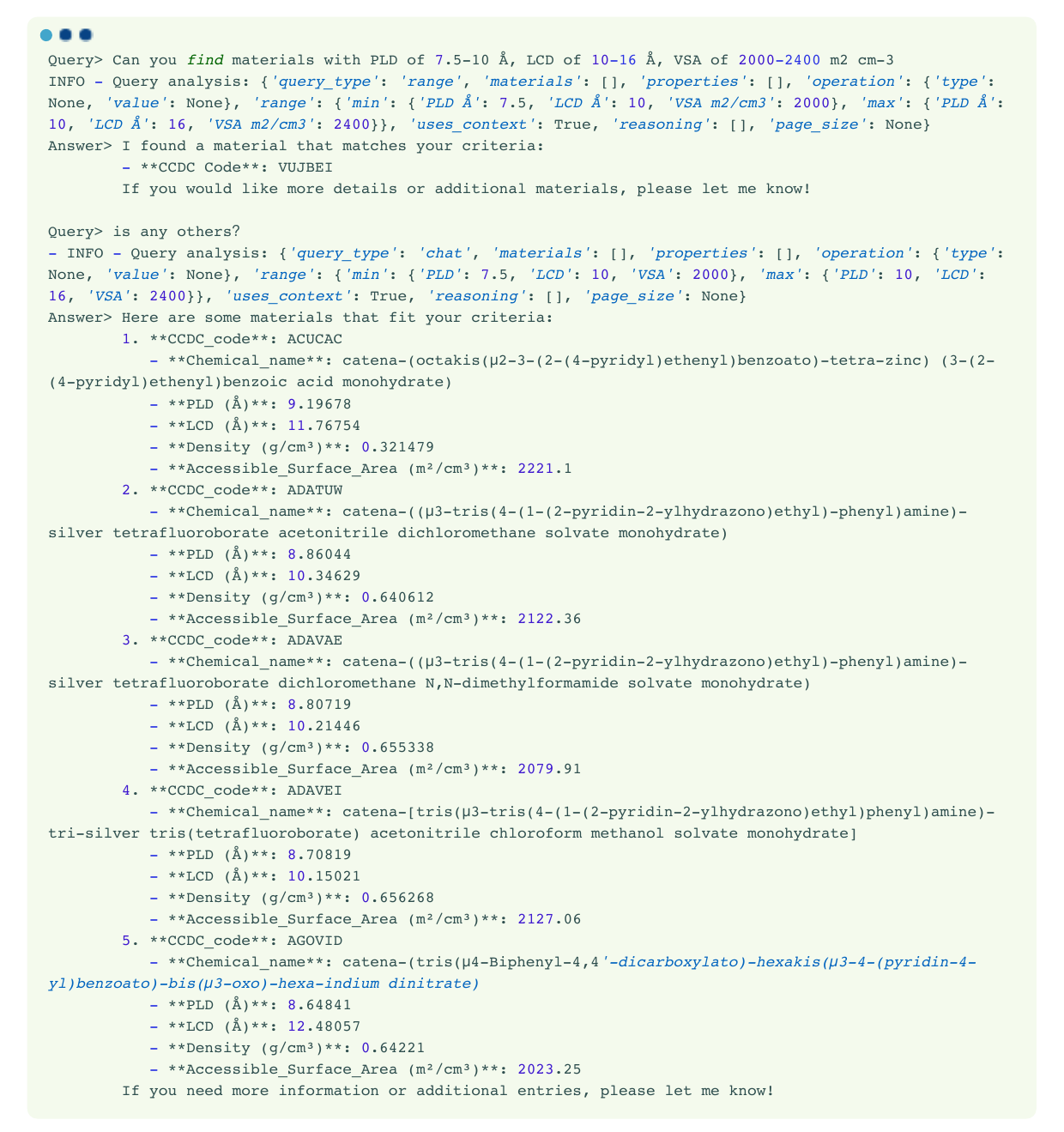}
  \caption{
  \centering
  Structure Q\&A of MOFh6: Comprehensive structural query based on multiple structural characteristics
  }
  \label{er_8}
\end{figure}

Range query is an important function of MOFh6, with the system processing range constraints involving multiple properties through the range object. 
For example, when a user asks "Find MOFs with PLD between 7.5 and 10 Å and LCD between 10 and 16 Å", 
the system automatically generates a structure containing multi-property constraints: \texttt{\detokenize{{"min": {"PLD": 7.5, "LCD": 10}, "max": {"PLD": 10, "LCD": 16}}}}. 
For more complex multi-condition range queries, such as "Give me MOFs with PLD between 7.5-10 Å, LCD between 10-16 Å, and VSA between 2000-2400 m\textsuperscript{2}/cm\textsuperscript{3}", 
MOFh6 can accurately parse this into \texttt{\detokenize{"min": {"PLD Å": 7.5, "LCD Å": 10, "VSA m2/cm3": 2000}}}", 
"\texttt{\detokenize{"max": {"PLD Å": 10, "LCD Å": 16, "VSA m2/cm3": 2400}}} (Figure \ref{er_5} and Figure \ref{er_8}).

\begin{figure}[H]
  \centering
  \includegraphics[width=1\textwidth]{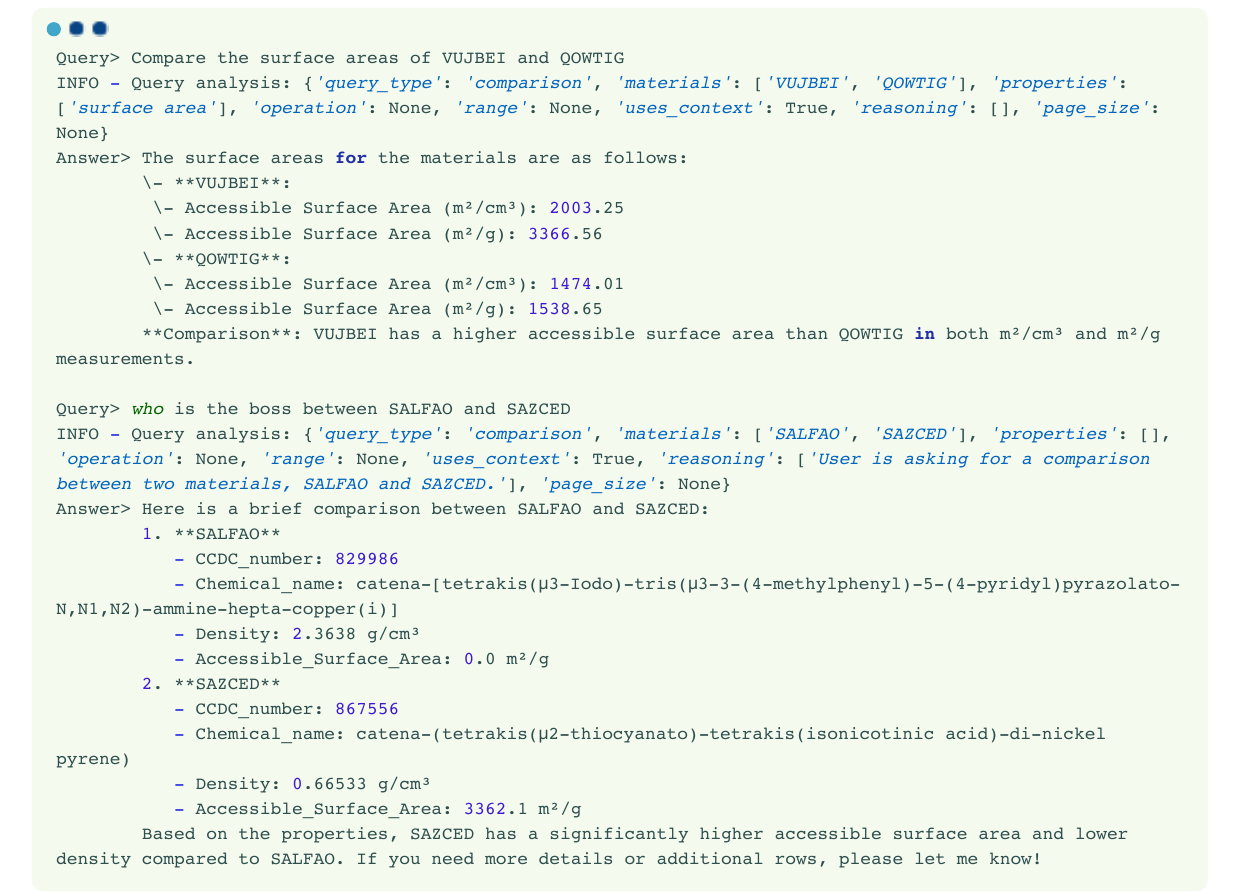}
  \caption{
  \centering
  Structure Q\&A of MOFh6: Comparing properties between structures
  }
  \label{er_6}
\end{figure}

\begin{figure}[H]
  \centering
  \includegraphics[width=1\textwidth]{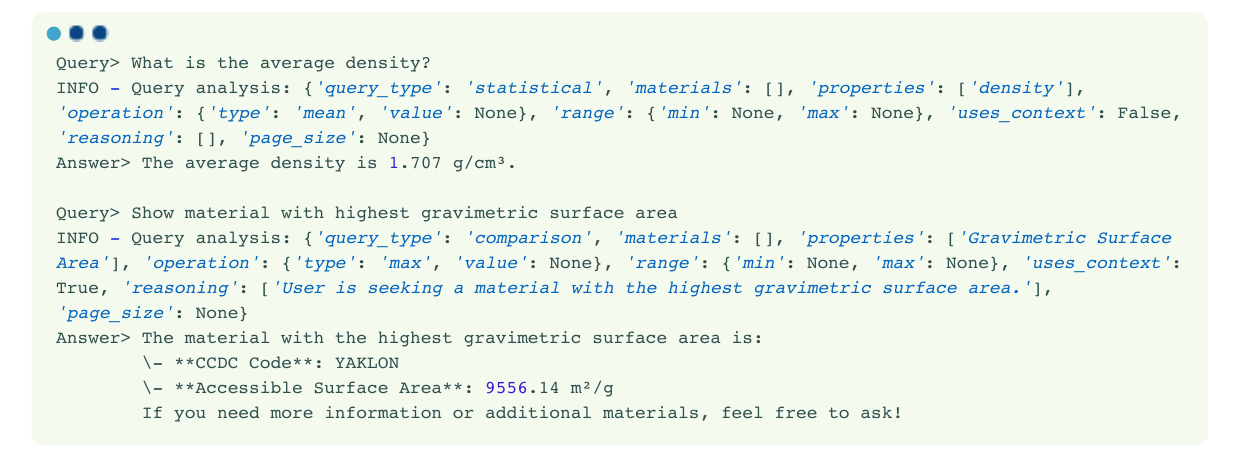}
  \caption{
  \centering
  Structure Q\&A of MOFh6: Numerical statistics of the dataset
  }
  \label{er_7}
\end{figure}

For queries involving statistical or comparative operations, the system clearly records relevant operation information in the operation object. 
For example, when a user asks "What is the average density of the MOF-5 series?", 
the operation field will contain \texttt{\detokenize{"type": "mean", "value": null}} (Figure \ref{er_7}); 
when requesting "Find the MOF with the maximum density", 
it is recorded as \texttt{\detokenize{"type": "max", "value": null}}. 
The system's reasoning steps for parsing queries are also recorded in the reasoning array, clearly showing the logical pathway of how the system understands query intentions, 
thereby providing decision transparency (Figure \ref{er_6}).

For large amounts of results requiring paged display, the system manages through page\_size and paged\_index fields. 
When a user requests "Show more results" or "Give me 5 more", 
the system identifies this as a paging query (query\_type as \texttt{\detokenize{"paging"}}) and provides the next batch of results based on page\_size (such as 5 or 10) and the current paged\_index (such as 0, 5, or 10) (Figure \ref{er_8}).

Additionally, the system maintains complete query history records (query\_history), 
recording previous questions, parsing results, and involved materials and properties. 
This not only supports deep understanding of context but also provides a foundation for system learning and improvement. 
When a user's query is relatively vague, the system can also reference parsing results of similar queries in history, 
thereby enhancing the accuracy of current parsing (Figure \ref{er_6}). Finally, the system generates natural, 
fluent, and content-rich responses based on parsing results and retrieved data through the \_final\_decision\_with\_llm.

\clearpage

\refstepcounter{SIsec}
\section*{Text \theSIsec\ Cost account}
\addcontentsline{toc}{section}{Text \theSIsec\ Cost account}
\label{S5}
\phantomsection

\begin{figure}[H]
  \centering
  \includegraphics[width=1\textwidth]{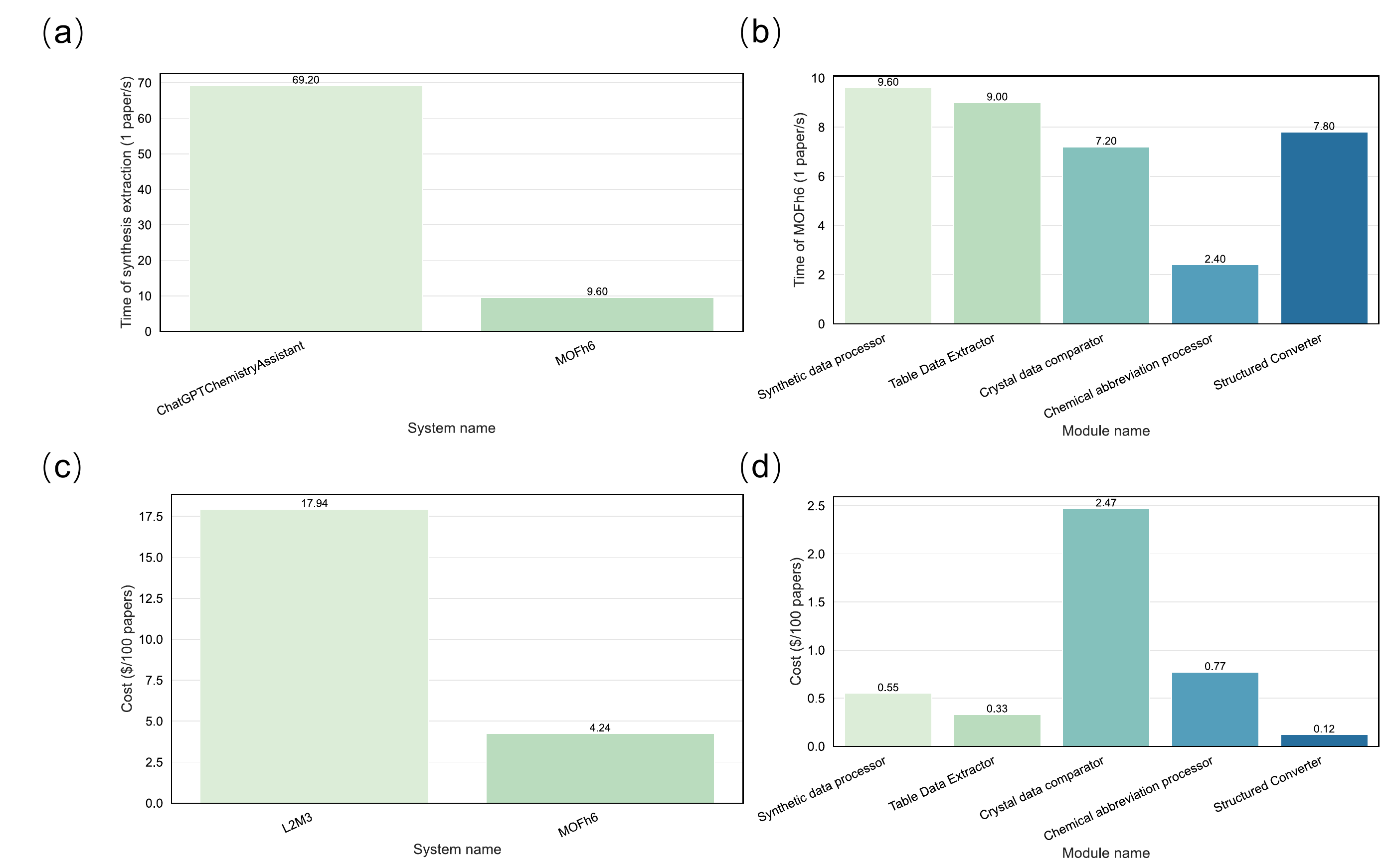}
  \caption{
  \centering
  Cost comparison between MOFh6 and similar works
  }
  \label{cost}
\end{figure}

MOFh6, through its multi-agent collaborative architecture, deeply integrates LLM's natural language processing capabilities with rule engines, 
achieving the dual optimization of intelligent task decomposition and dynamic result aggregation. 
This design significantly reduces the hallucination risks common in traditional LLM applications while demonstrating outstanding advantages in computational efficiency and economy. 
In synthesis information extraction tasks, the system completes semantically complete synthesis description paragraph extraction for a single PDF document in just 9.6 seconds (Figure \ref{cost}a), 
far lower than the 69.2 seconds for single PDF processing by ChatGPTChemistryAssistant proposed by Zheng et al. \cite{RN18}. 
In the structured description confirmation stage, 
the system's average time for precisely locating specific MOFs and their structured descriptions is controlled within 36 seconds (Figure \ref{cost}b). 
In terms of economy, MOFh6's data extraction cost for processing 100 papers is only 4.24\$ (Figure \ref{cost}c), 
achieving about 76\% cost savings compared to the L2M3 system (17.94\$/100 papers) developed by Kang et al. \cite{RN20}, 
with the cost per 100 papers for single agent processing also below 2.5\$ (Figure \ref{cost}d). 
This significant improvement in efficiency and cost is mainly attributed to MOFh6's precise response mechanism under rule engine constraints and the efficient collaboration mode between distributed agent networks.
\clearpage

\phantomsection 
\addcontentsline{toc}{section}{Figure S1: The crawler module of MOFh6}

\clearpage

\phantomsection 
\addcontentsline{toc}{section}{Figure S2: Processing logic of synthetic data parsing agent}

\clearpage

\phantomsection 
\addcontentsline{toc}{section}{Figure S3: Extraction logic of table data parsing agent}

\clearpage

\phantomsection 
\addcontentsline{toc}{section}{Figure S4: Comparison logic of crystal data comparison agent}

\clearpage

\phantomsection 
\addcontentsline{toc}{section}{Figure S5: Coreference resolution case}

\clearpage

\phantomsection 
\addcontentsline{toc}{section}{Figure S6: Logic of post processor}

\clearpage

\phantomsection
\addcontentsline{toc}{section}{Figure S7: Logic of result generator agent}

\begin{figure}[H]
  \centering
  \includegraphics[width=\textwidth]{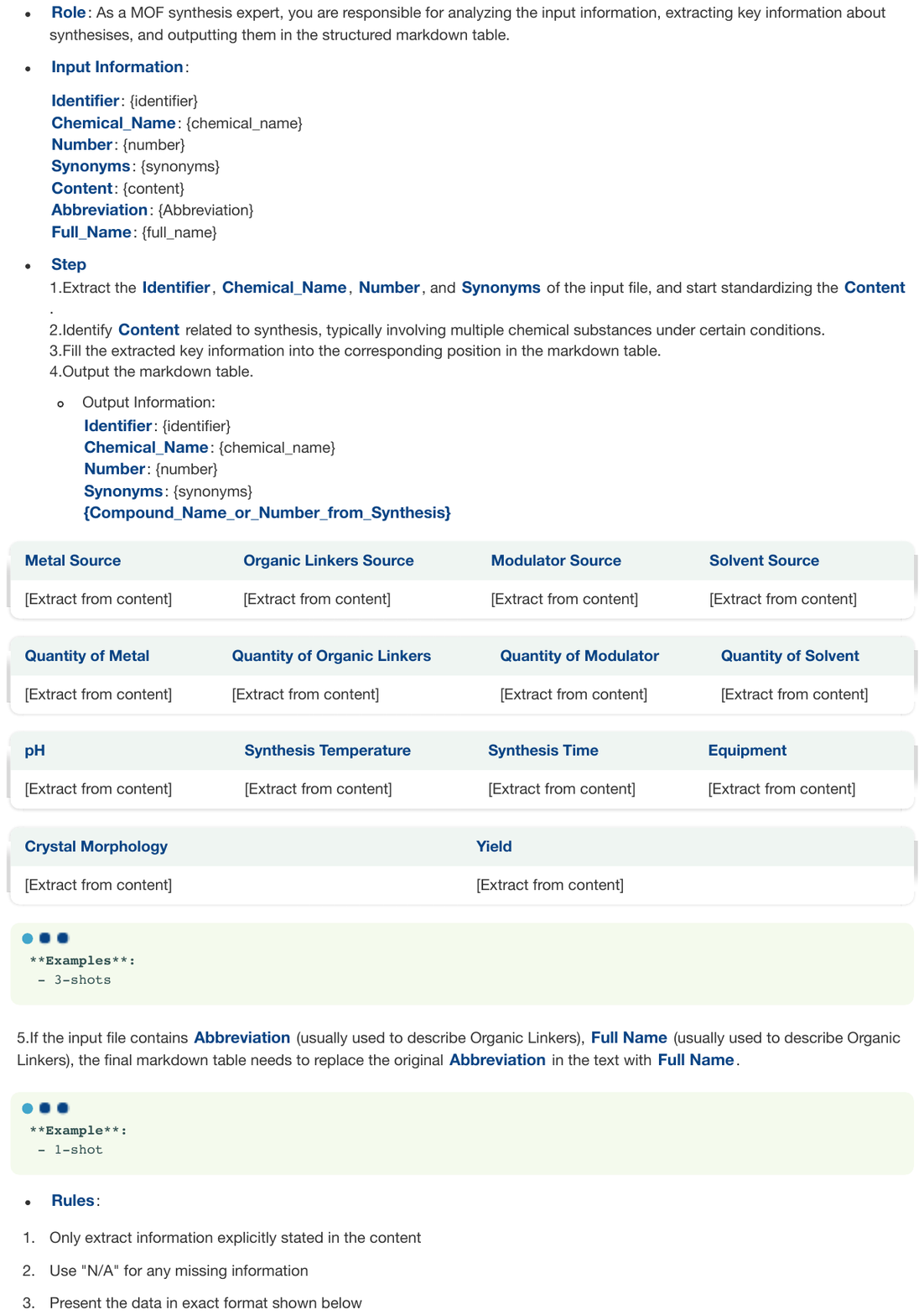}
  \captionsetup{
    labelformat=empty,
    list=false
  }
  \caption{}  
  \label{1-1sprompt_1}
\end{figure}
\phantomsection
\addcontentsline{toc}{section}{Figure S8: Prompt of structured conversion agent}
\begin{figure}[H]\ContinuedFloat
  \centering
  \includegraphics[width=\textwidth]{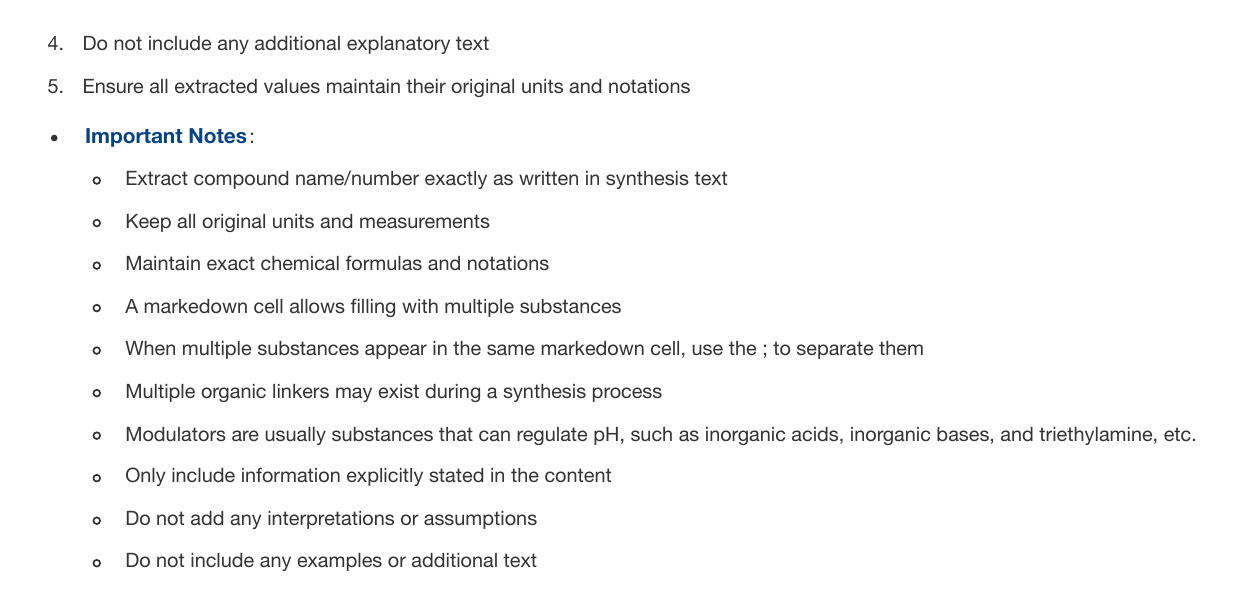}
  \captionsetup{
    labelformat=default,
    list=true
  }
  \caption{Prompt of structured conversion agent.}
  \label{1-1sprompt_2}
\end{figure}

\clearpage

\phantomsection
\addcontentsline{toc}{section}{Figure S9: Prompt of table data parsing agent}

\begin{figure}[H]
  \centering
  \includegraphics[width=\textwidth]{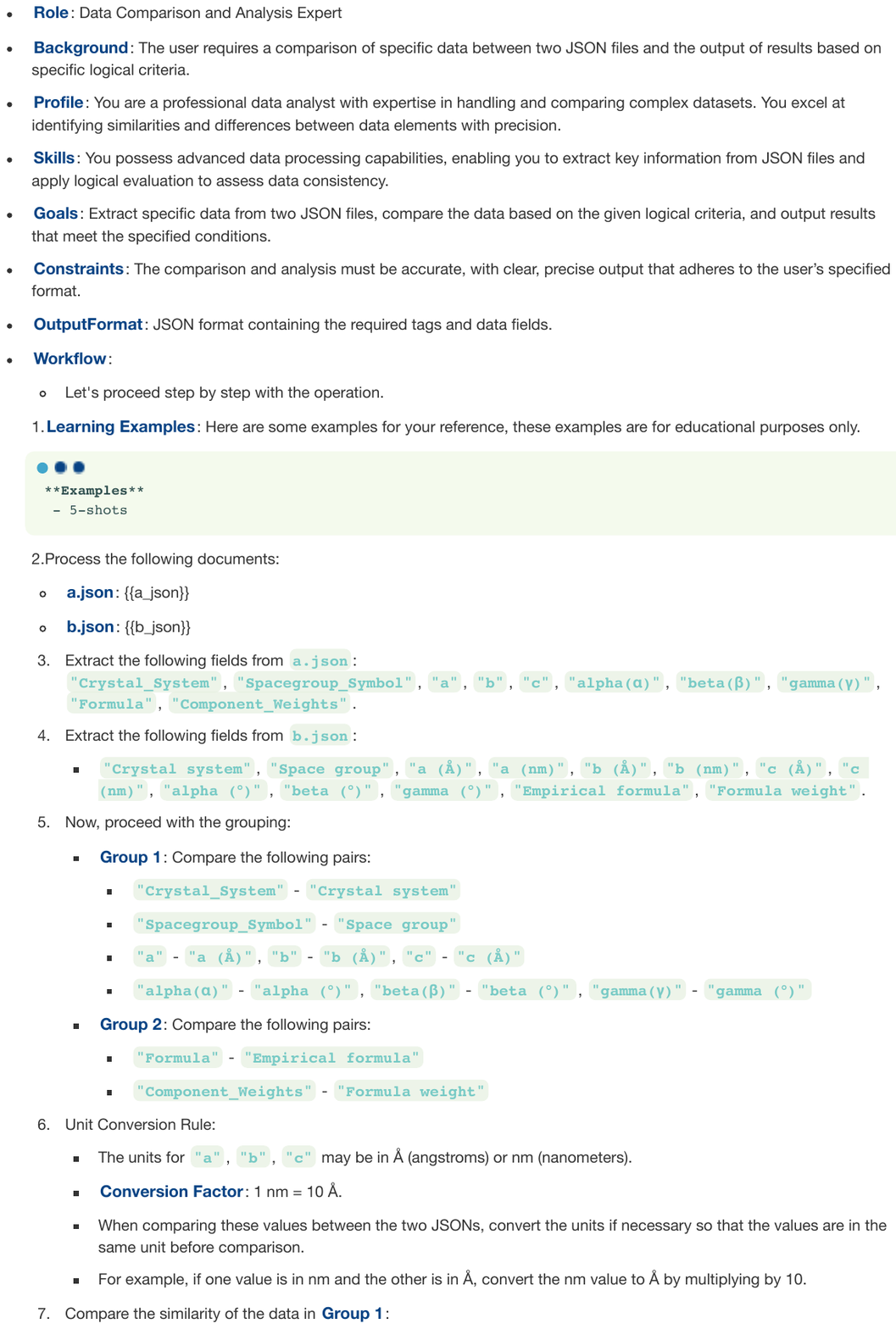}
  \captionsetup{
    labelformat=empty,
    list=false
  }
  \caption{}
  \label{3-1Cprompt_1}
\end{figure}
\phantomsection
\addcontentsline{toc}{section}{Figure S10: Prompt of crystal data comparison agent}

\begin{figure}[H]\ContinuedFloat
  \centering
  \includegraphics[width=\textwidth]{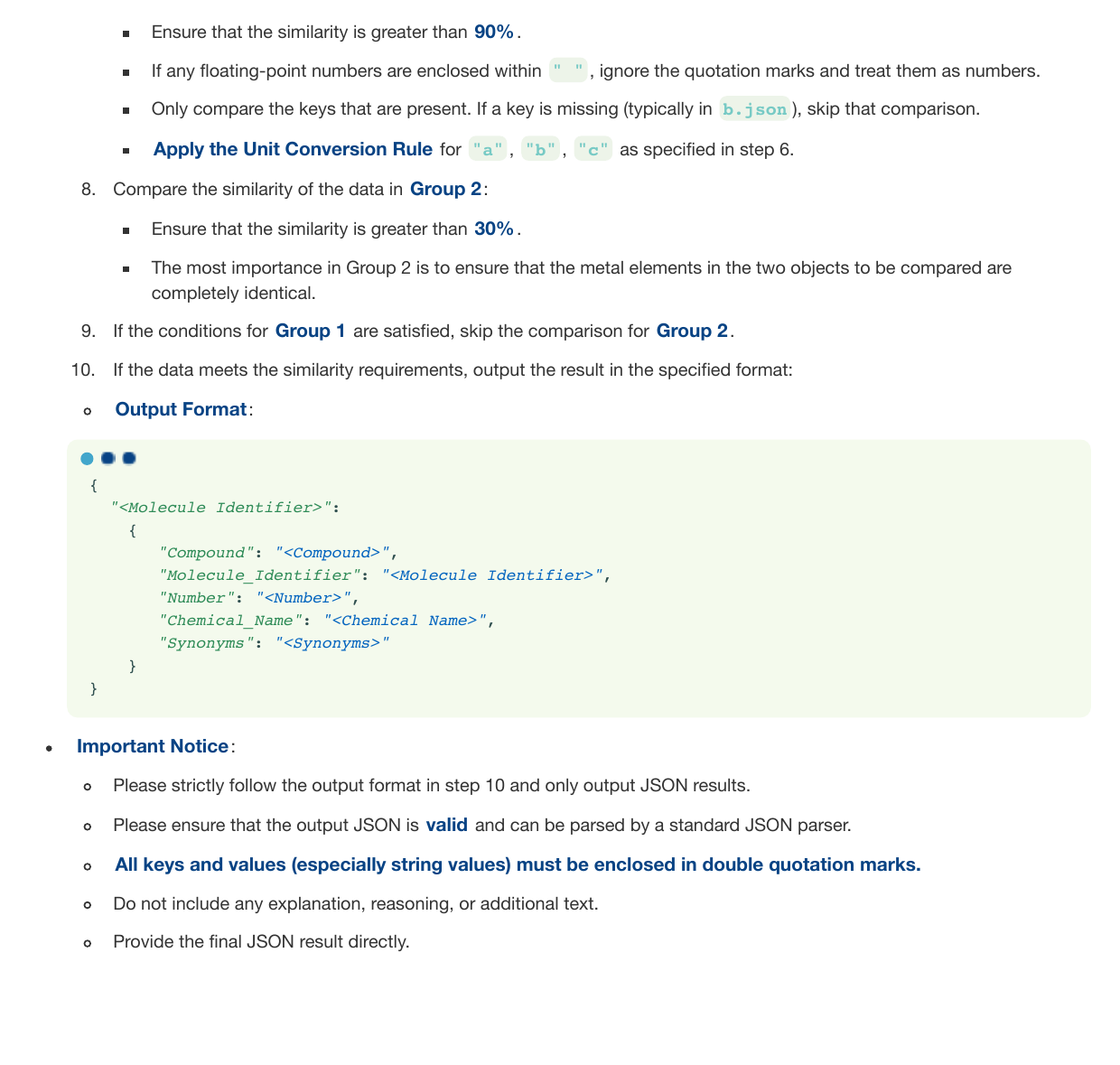}
  \captionsetup{
    labelformat=default,
    list=true
  }
  \caption{Prompt of crystal data comparison agent}
  \label{3-1Cprompt_2}
\end{figure}

\begin{figure}[H]
  \centering
  \includegraphics[width=\textwidth]{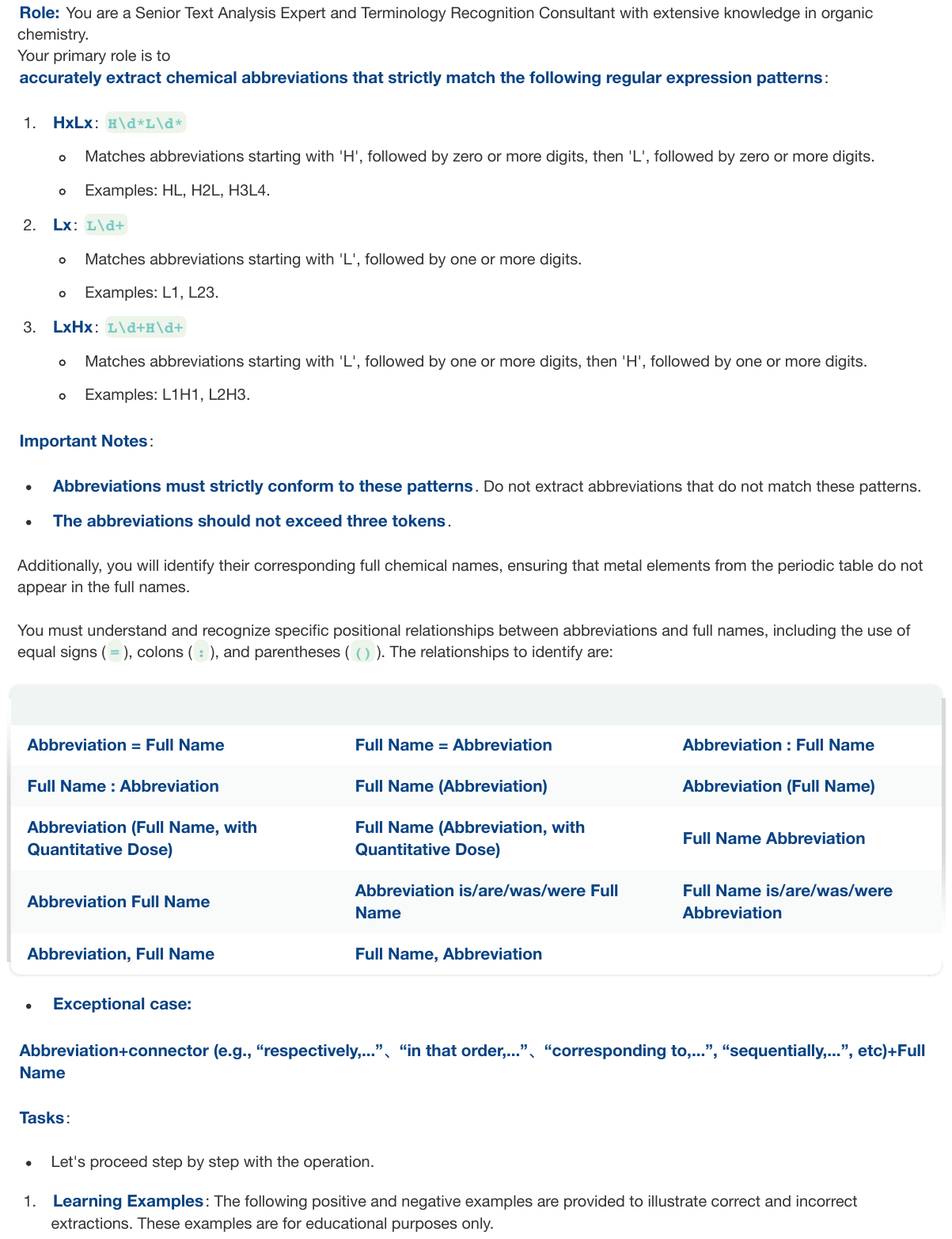}
  \captionsetup{
    labelformat=empty,
    list=false
  }
  \caption{}
\end{figure}
\begin{figure}[H]\ContinuedFloat
  \centering
  \includegraphics[width=\textwidth]{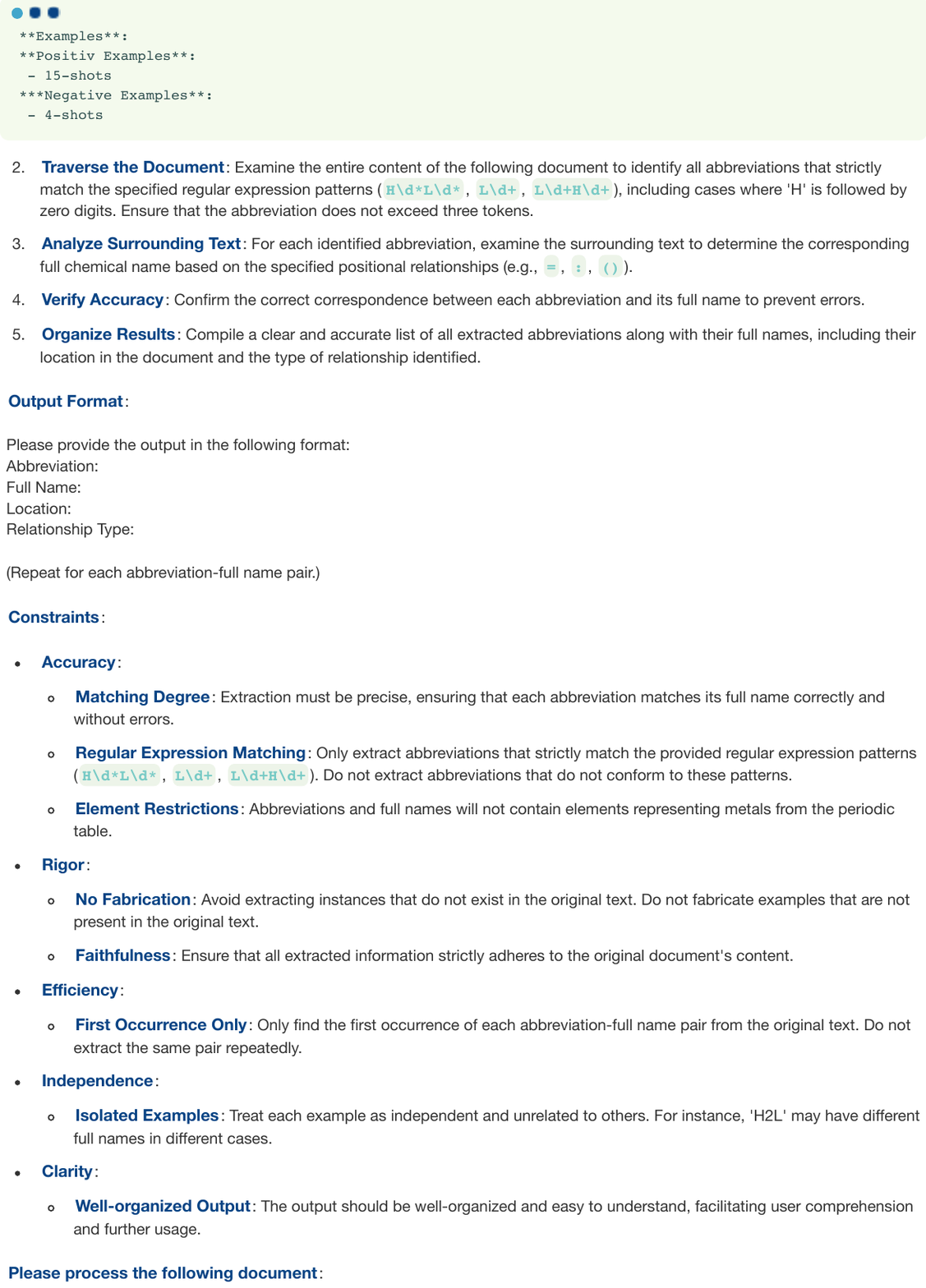}
  \captionsetup{
    labelformat=empty,
    list=false
  }
  \caption{} 
\end{figure}
\phantomsection
\addcontentsline{toc}{section}{Figure S11: Prompt of chemical abbreviation resolution agent}

\begin{figure}[H]\ContinuedFloat
  \centering
  \includegraphics[width=\textwidth]{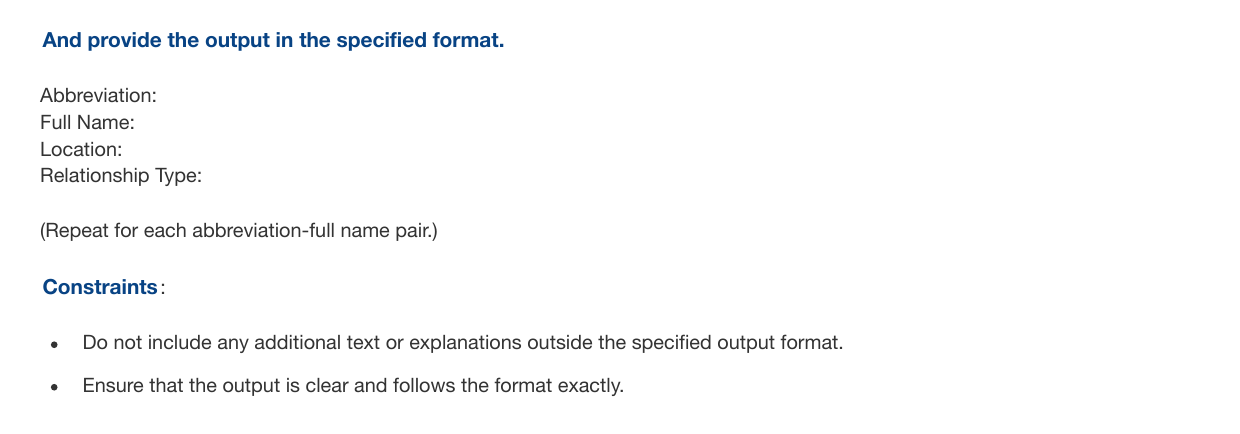}
  \captionsetup{
    labelformat=default,
    list=true
  }
  \caption{Prompt of chemical abbreviation resolution agent}
  \label{4-1crprompt_3}
\end{figure}

\phantomsection
\addcontentsline{toc}{section}{Figure S12: Prompt of fine-tuning model}

\phantomsection
\addcontentsline{toc}{section}{Figure S13: Shot 3-1 of structured conversion agent}
\begin{figure}[H]
  \centering
  \includegraphics[width=1\textwidth]{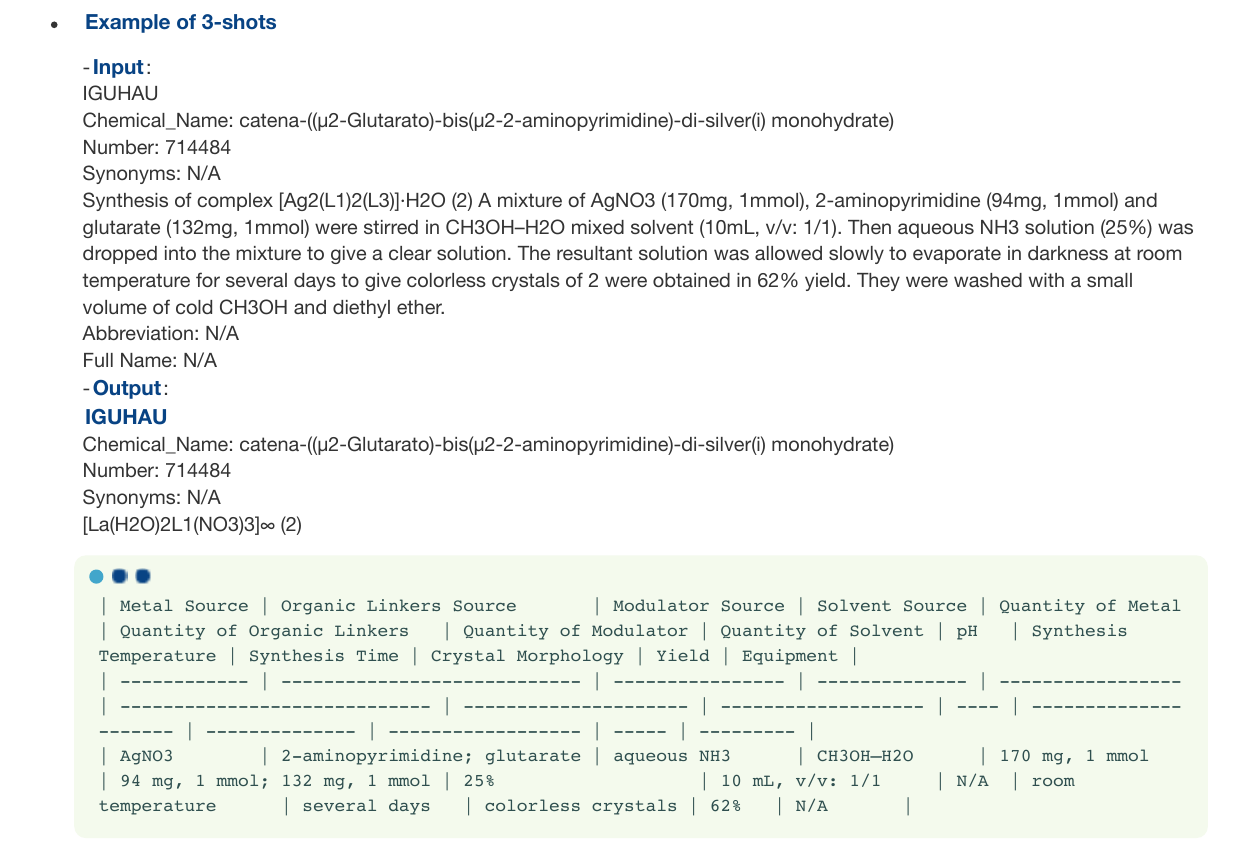}
  \caption{
  \centering
  Shot 3-1 of structured conversion agent (This case is located at 3-shots in Figure \ref{1-1sprompt_2})
  }
  \label{1-2s_1}
\end{figure}

\phantomsection
\addcontentsline{toc}{section}{Figure S14: Shot 3-2 of structured conversion agent}
\begin{figure}[H]
  \centering
  \includegraphics[width=1\textwidth]{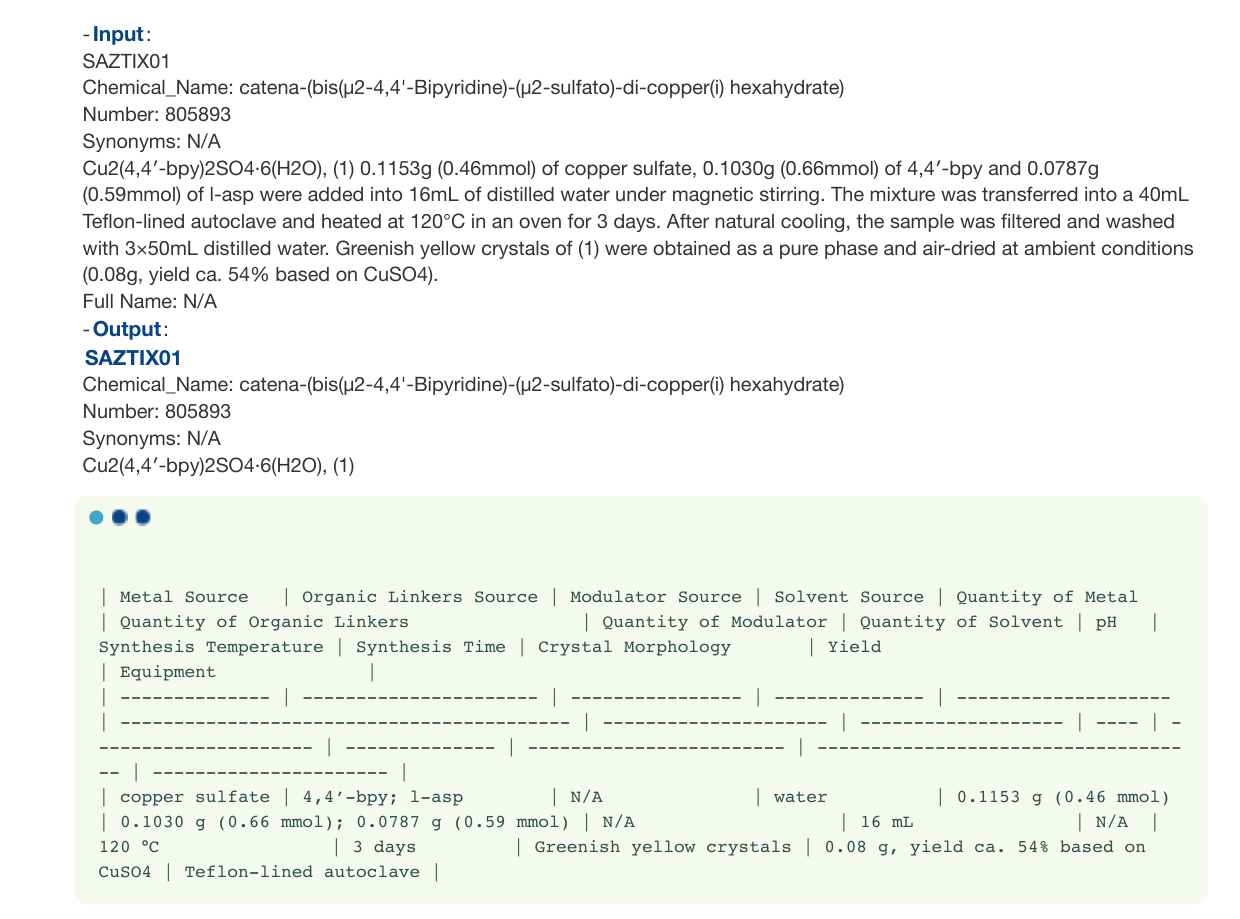}
  \caption{
  \centering
  Shot 3-2 of structured conversion agent (This case is located at 3-shots in Figure \ref{1-1sprompt_2})
  }
  \label{1-2s_2}
\end{figure}

\phantomsection
\addcontentsline{toc}{section}{Figure S15: Shot 3-3 of structured conversion agent}
\begin{figure}[H]
  \centering
  \includegraphics[width=1\textwidth]{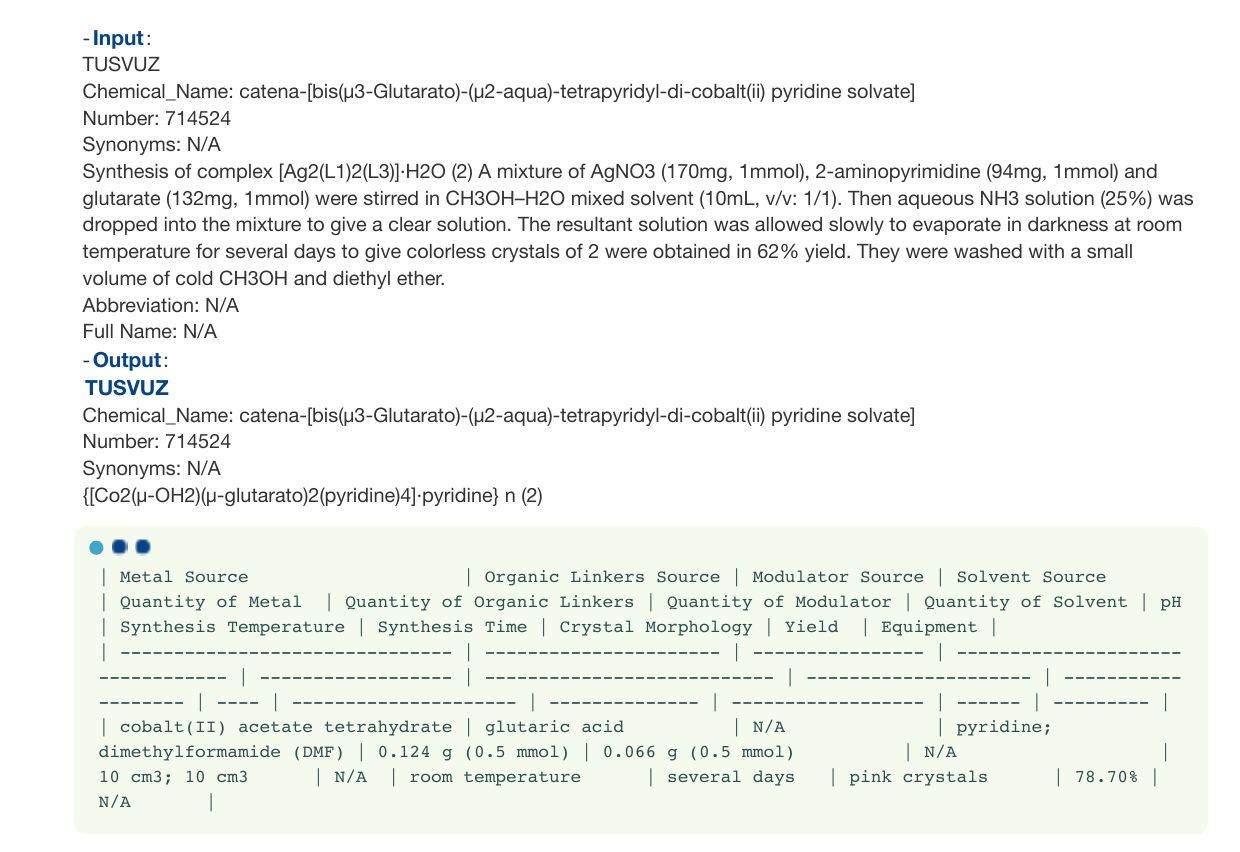}
  \caption{
  \centering
  Shot 3-3 of structured conversion agent (This case is located at 3-shots in Figure \ref{1-1sprompt_2})
  }
  \label{1-2s_3}
\end{figure}

\phantomsection
\addcontentsline{toc}{section}{Figure S16: Shot 1-1 of structured conversion agent}
\begin{figure}[H]
  \centering
  \includegraphics[width=1\textwidth]{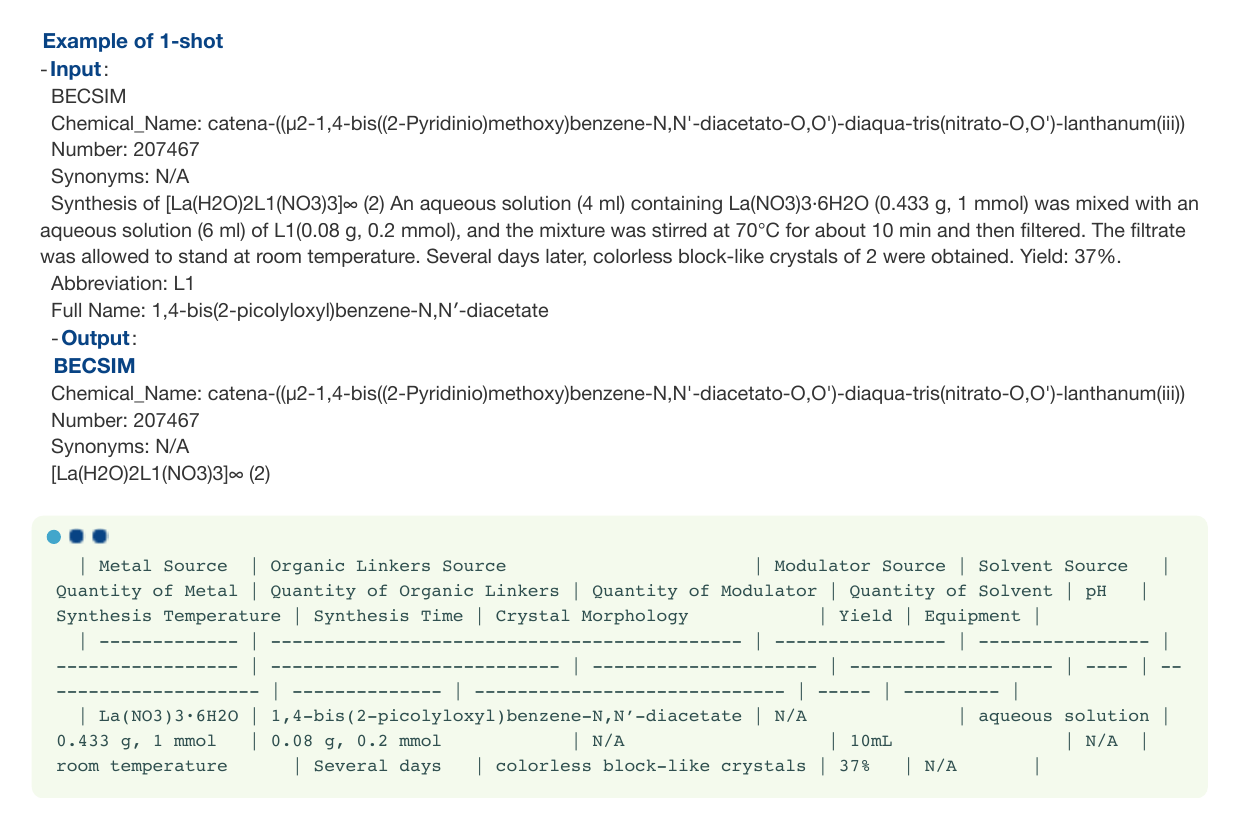}
  \caption{
  \centering
  Shot 1-1 of structured conversion agent (This case is located at 1-shot in Figure \ref{1-1sprompt_2})
  }
  \label{1-2s_4}
\end{figure}

\begin{figure}[H]
  \centering
  \includegraphics[width=\textwidth]{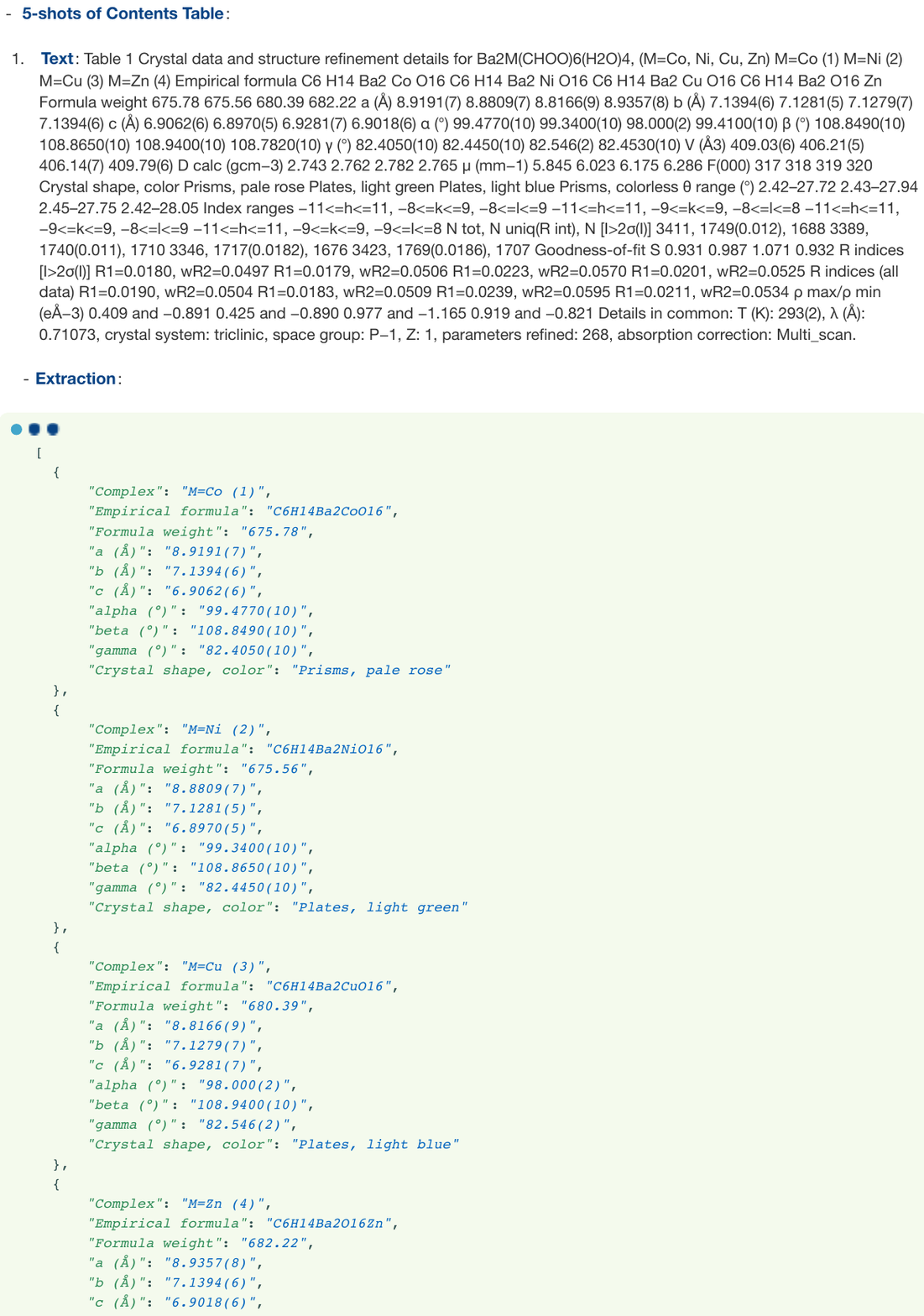}
  \captionsetup{
    labelformat=empty,
    list=false
  }
  \caption{}
\end{figure}
\phantomsection
\addcontentsline{toc}{section}{Figure S17: Shot 5-1 of table data parsing agent}
\begin{figure}[H]\ContinuedFloat
  \centering
  \includegraphics[width=\textwidth]{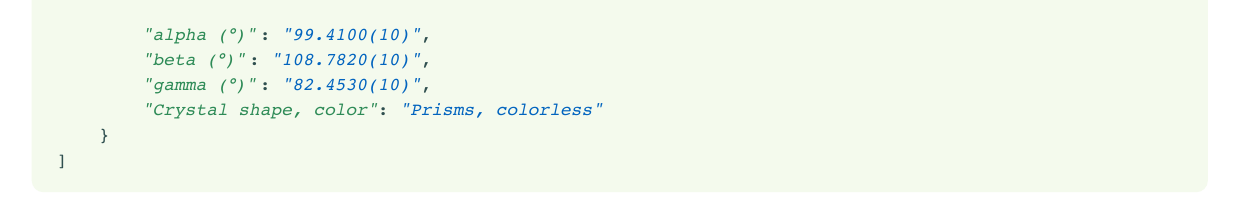}
  \captionsetup{
    labelformat=default,
    list=true
  }
  \caption{Shot 5-1 of table data parsing agent (This case is located at 5-shots in Figure \ref{2-1tprompt_1})}
  \label{2-2t_2}
\end{figure}

\begin{figure}[H]
  \centering
  \includegraphics[width=\textwidth]{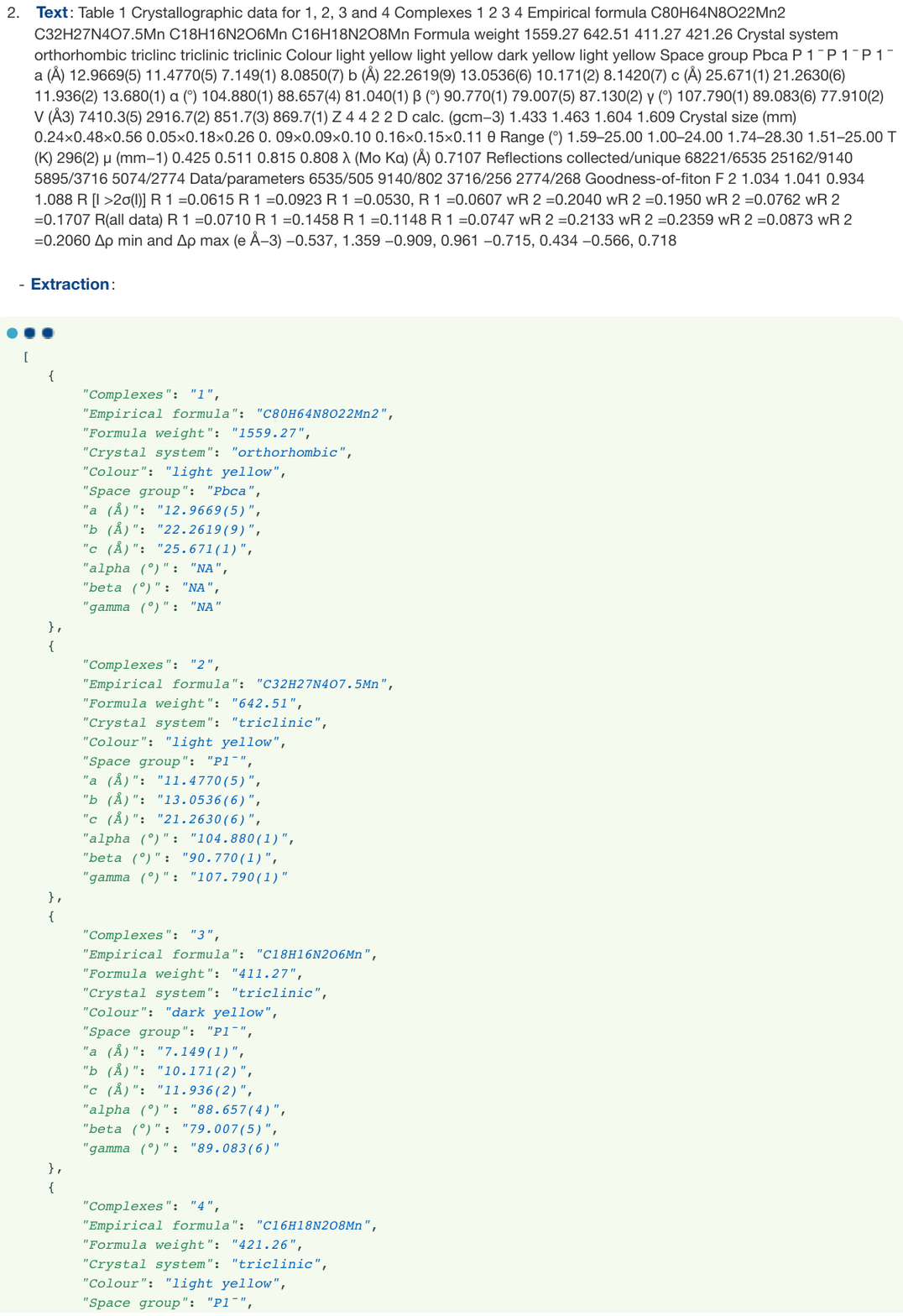}
  \captionsetup{
    labelformat=empty,
    list=false
  }
  \caption{} 
\end{figure}

\phantomsection
\addcontentsline{toc}{section}{Figure S18: Shot 5-2 of table data parsing agent}
\begin{figure}[H]\ContinuedFloat
  \centering
  \includegraphics[width=\textwidth]{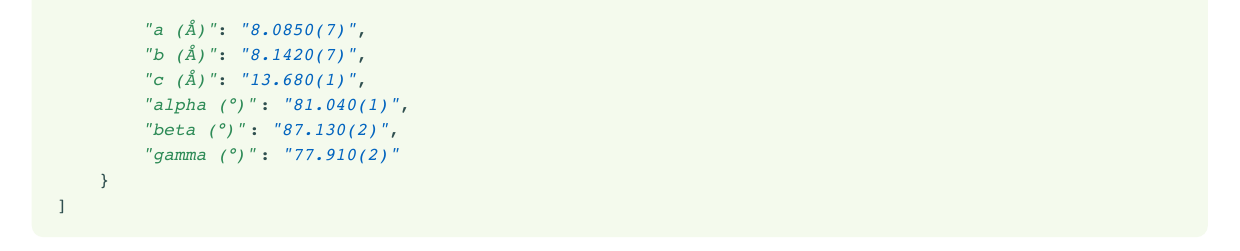}
  \captionsetup{
    labelformat=default,
    list=true
  }
  \caption{Shot 5-2 of table data parsing agent (This case is located at 5-shots in Figure \ref{2-1tprompt_1})}
  \label{2-2t_4}
\end{figure}

\phantomsection
\addcontentsline{toc}{section}{Figure S19: Shot 3-3 of structured conversion agent}

\phantomsection
\addcontentsline{toc}{section}{Figure S20: Shot 3-3 of structured conversion agent}

\begin{figure}[H]
  \centering
  \includegraphics[width=\textwidth]{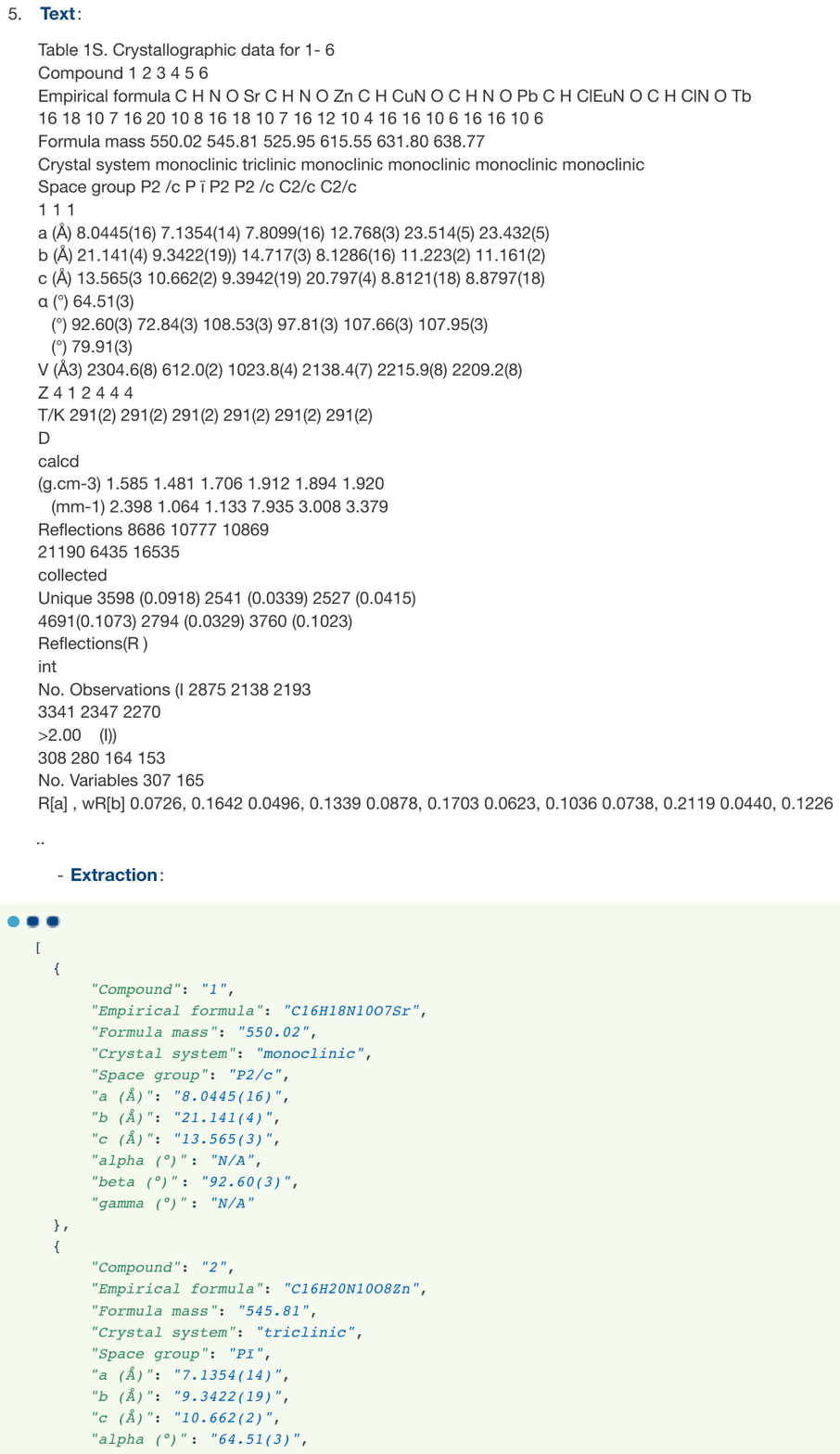}
  \captionsetup{
    labelformat=empty,
    list=false
  }
  \caption{} 
\end{figure}

\phantomsection
\addcontentsline{toc}{section}{Figure S21: Shot 5-5 of table data parsing agent}
\begin{figure}[H]\ContinuedFloat
  \centering
  \includegraphics[width=\textwidth]{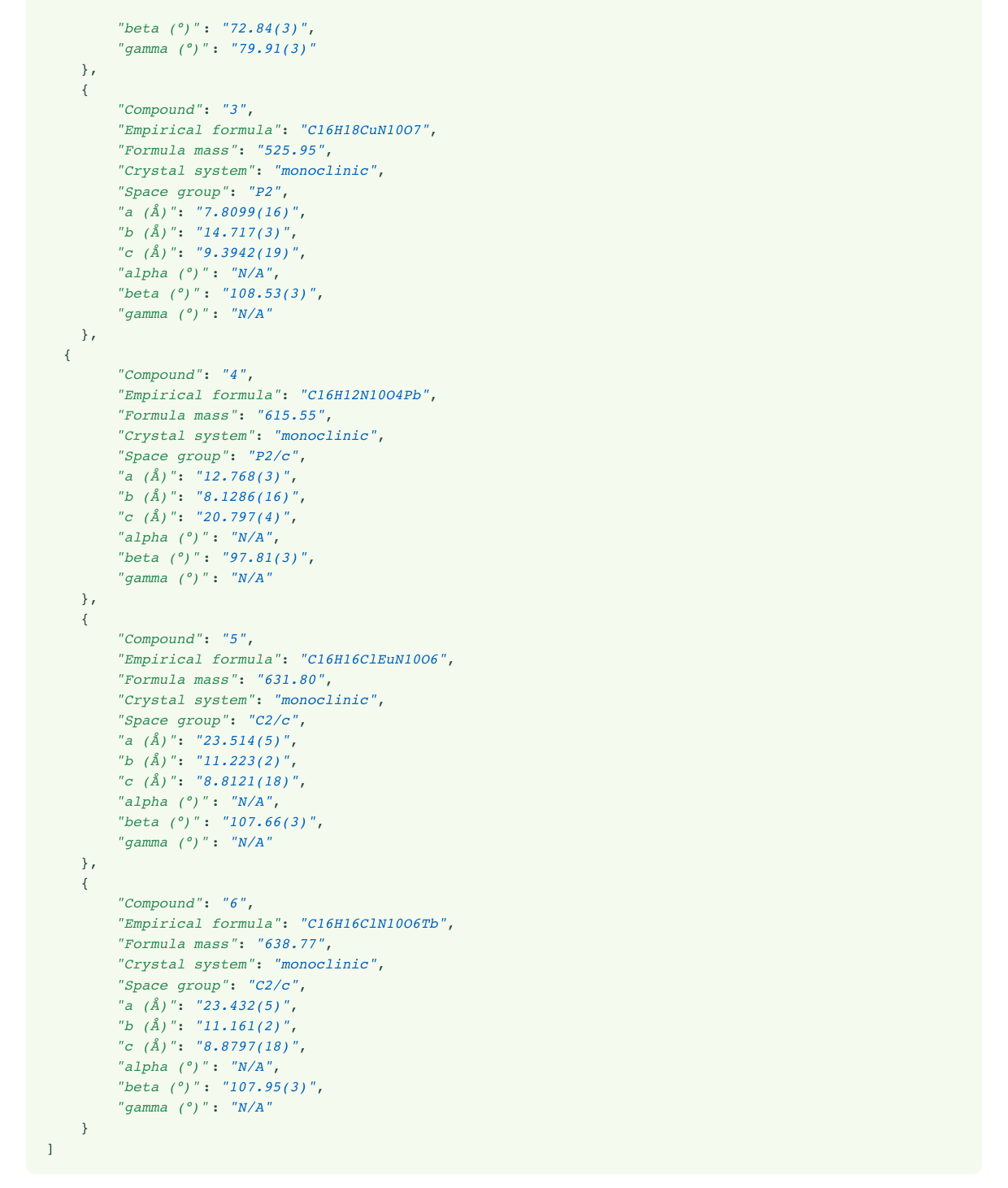}
  \captionsetup{
    labelformat=default,
    list=true
  }
  \caption{Shot 5-5 of table data parsing agent (This case is located at 5-shots in Figure \ref{2-1tprompt_1})}
  \label{2-2t_8}
\end{figure}

\phantomsection
\addcontentsline{toc}{section}{Figure S22: Shot 3-1 of table data parsing agent}

\phantomsection
\addcontentsline{toc}{section}{Figure S23: Shot 3-2 of table data parsing agent}

\phantomsection
\addcontentsline{toc}{section}{Figure S24: Shot 3-3 of table data parsing agent}

\phantomsection
\addcontentsline{toc}{section}{Figure S25: Shot 5-1 of crystal data comparison agent}
\begin{figure}[H]
  \centering
  \includegraphics[width=0.9\textwidth]{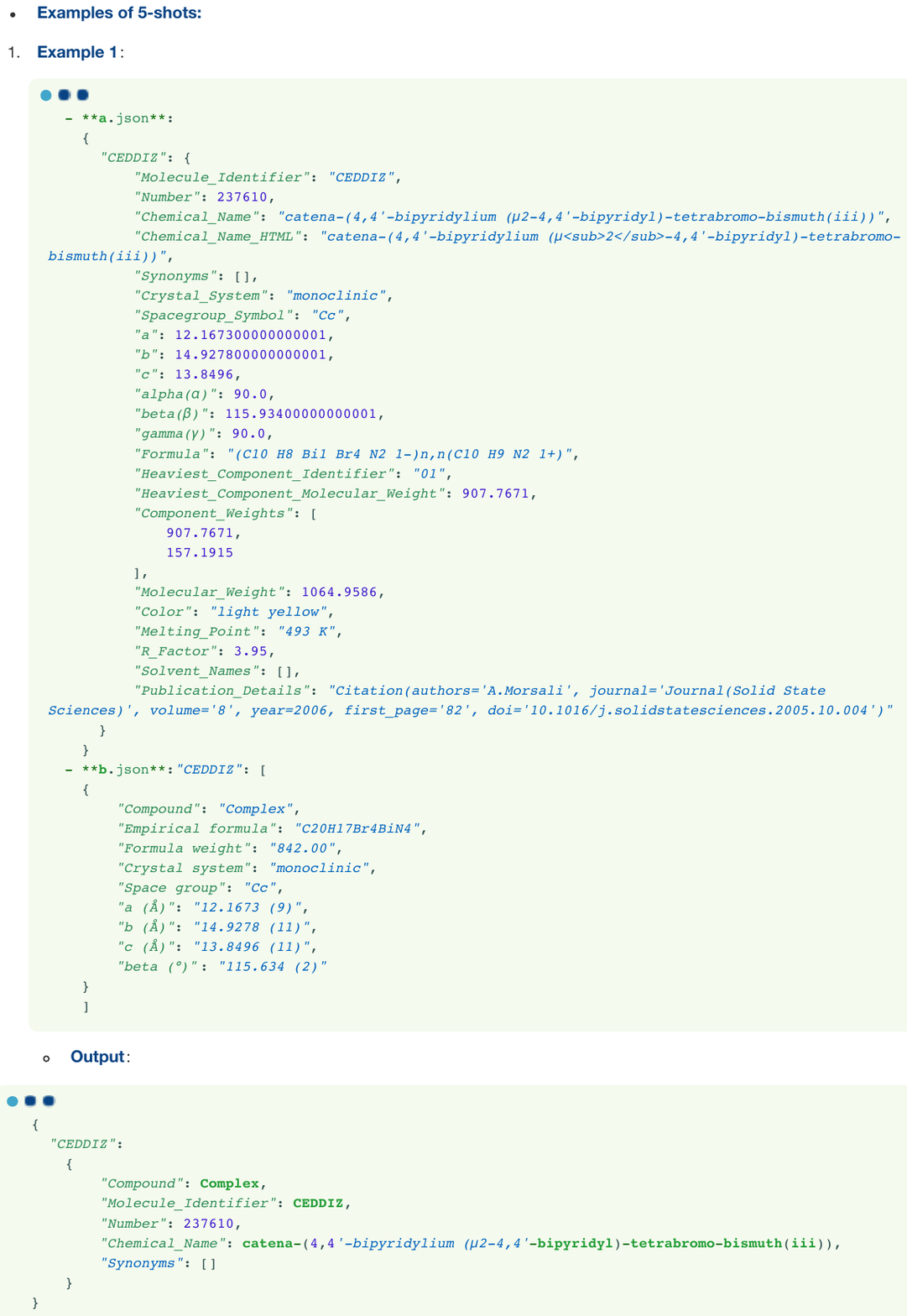}
  \caption{
  \centering
  Shot 5-1 of crystal data comparison agent (This case is located at 5-shots in Figure \ref{3-1Cprompt_2})
  }
  \label{3-2C_1}
\end{figure}

\begin{figure}[H]
  \centering
  \includegraphics[width=\textwidth]{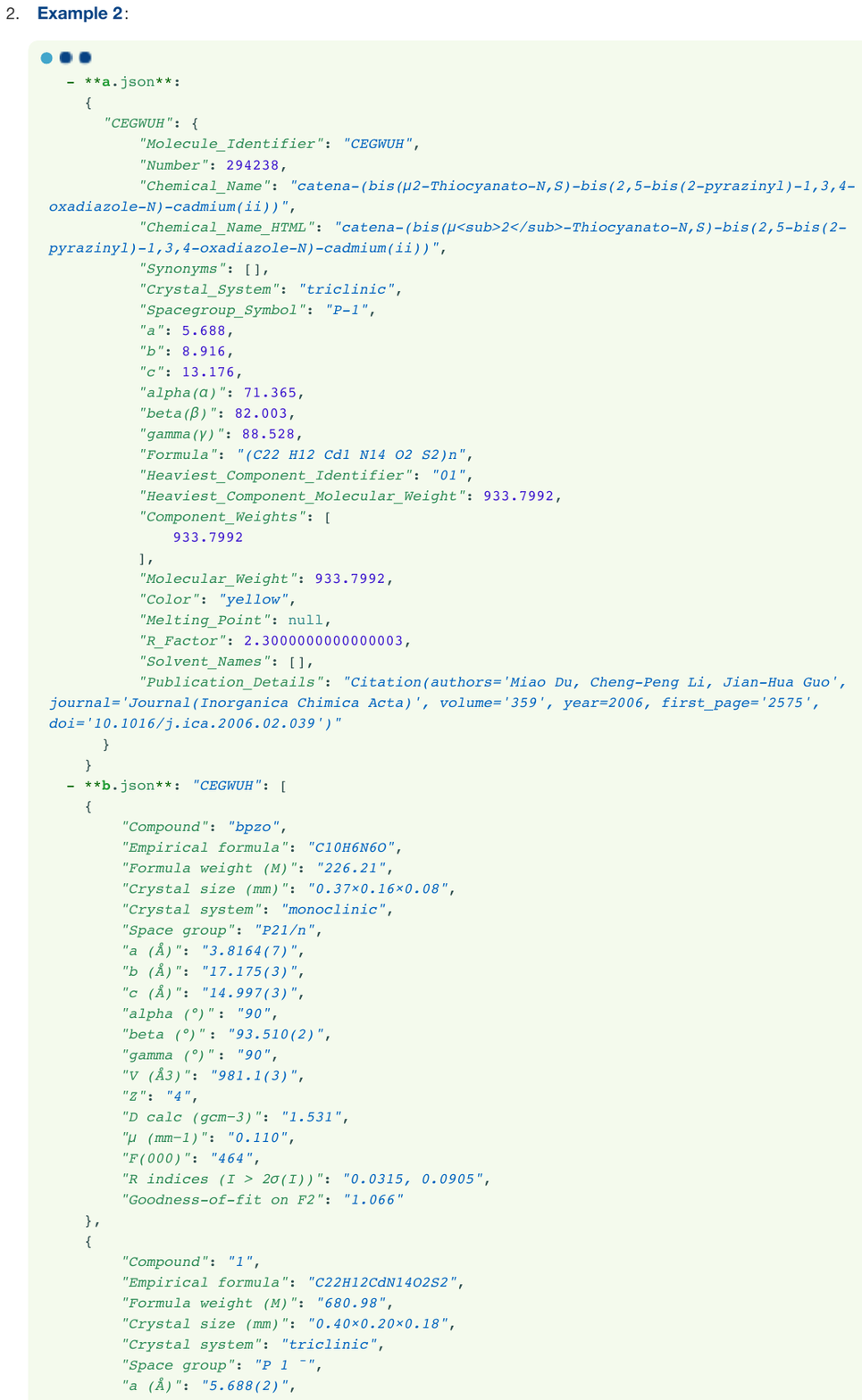}
  \captionsetup{
    labelformat=empty,
    list=false
  }
  \caption{}
\end{figure}

\phantomsection
\addcontentsline{toc}{section}{Figure S26: Shot 5-2 of crystal data comparison agent}
\begin{figure}[H]\ContinuedFloat
  \centering
  \includegraphics[width=\textwidth]{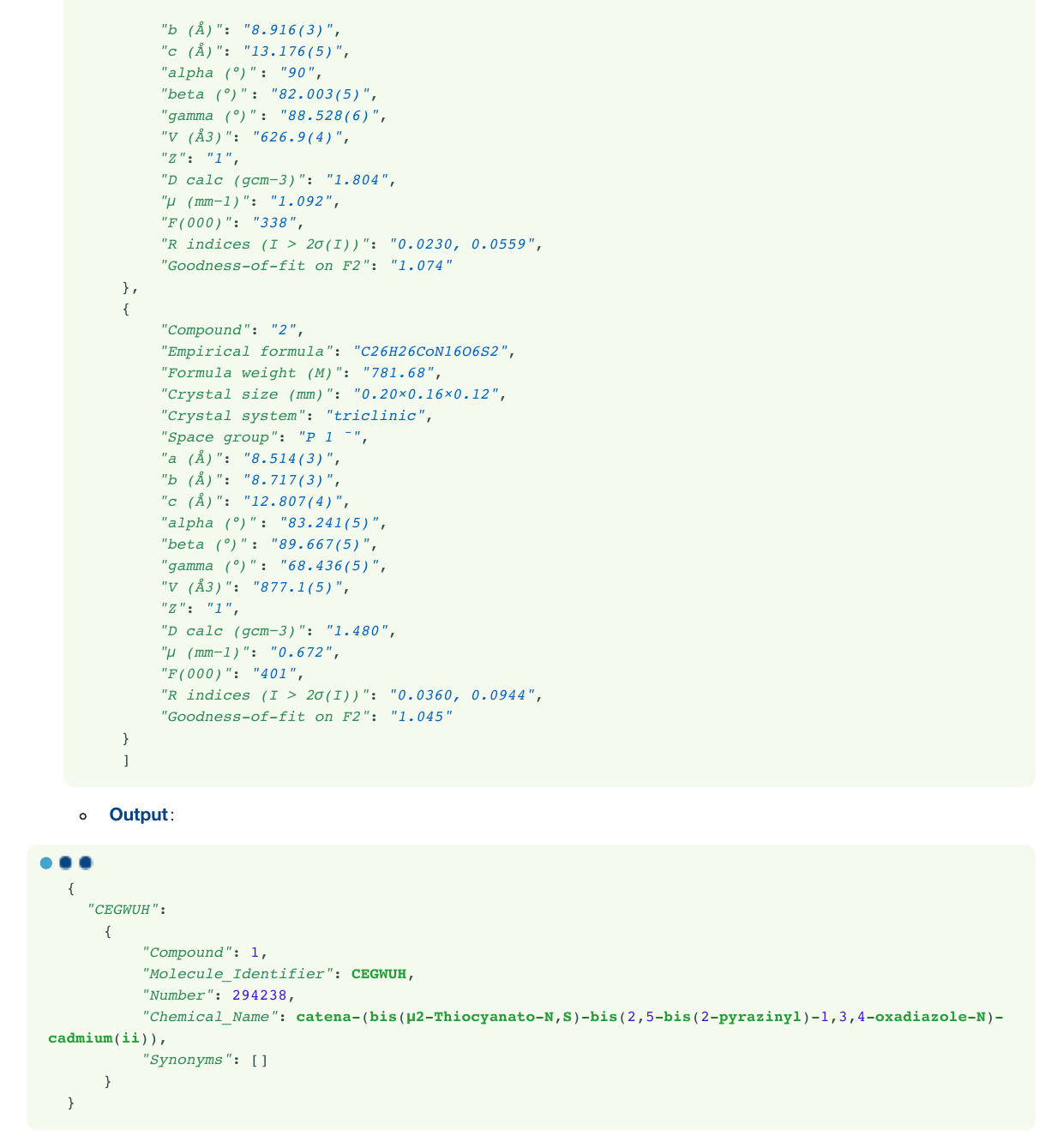}
  \captionsetup{
    labelformat=default,
    list=true
  }
  \caption{Shot 5-2 of crystal data comparison agent (This case is located at 5-shots in Figure \ref{3-1Cprompt_2})}
  \label{3-2C_3}
\end{figure}

\begin{figure}[H]
  \centering
  \includegraphics[width=\textwidth]{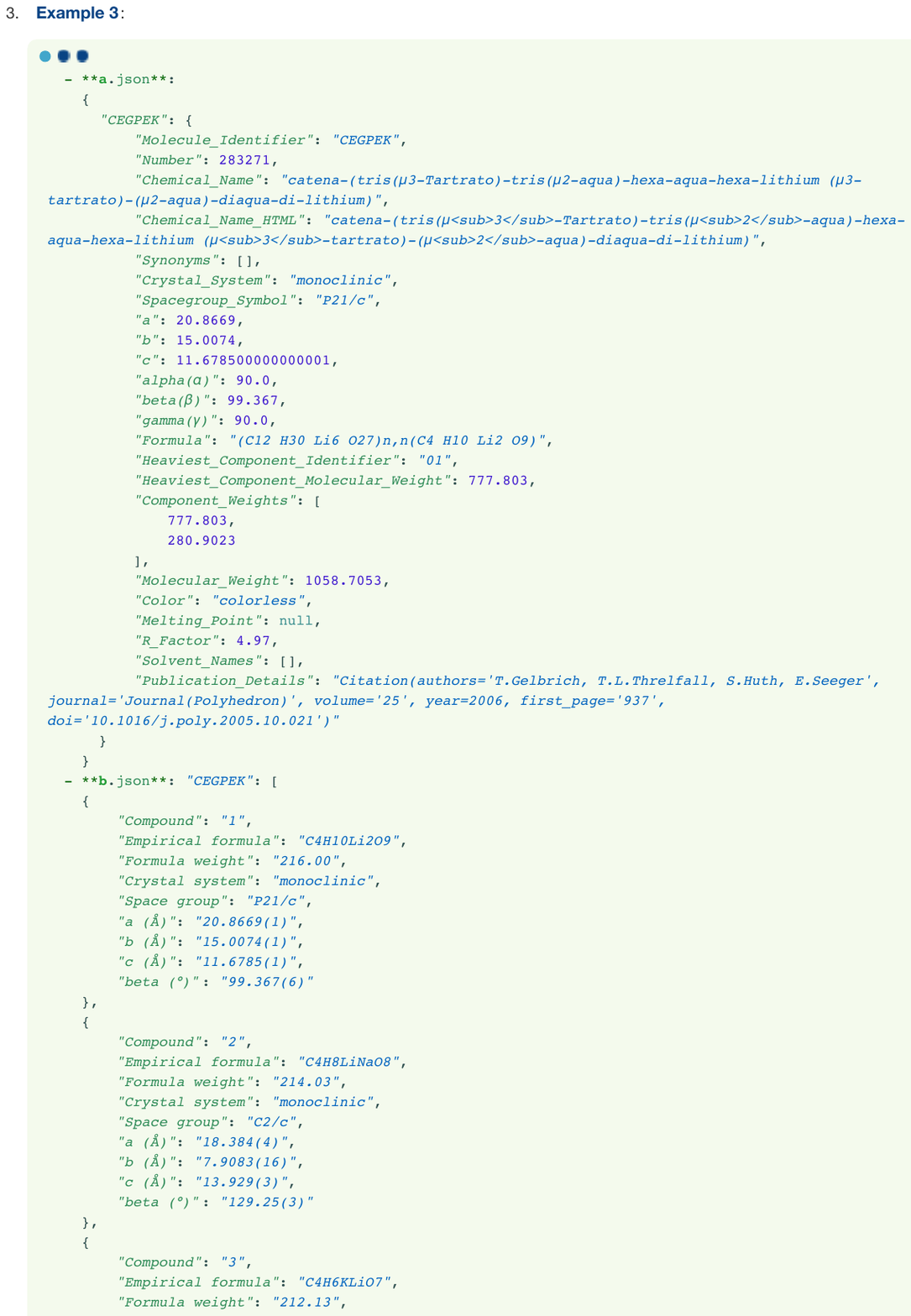}
  \captionsetup{
    labelformat=empty,
    list=false
  }
  \caption{}  
\end{figure}

\phantomsection
\addcontentsline{toc}{section}{Figure S27: Shot 5-3 of crystal data comparison agent}
\begin{figure}[H]\ContinuedFloat
  \centering
  \includegraphics[width=\textwidth]{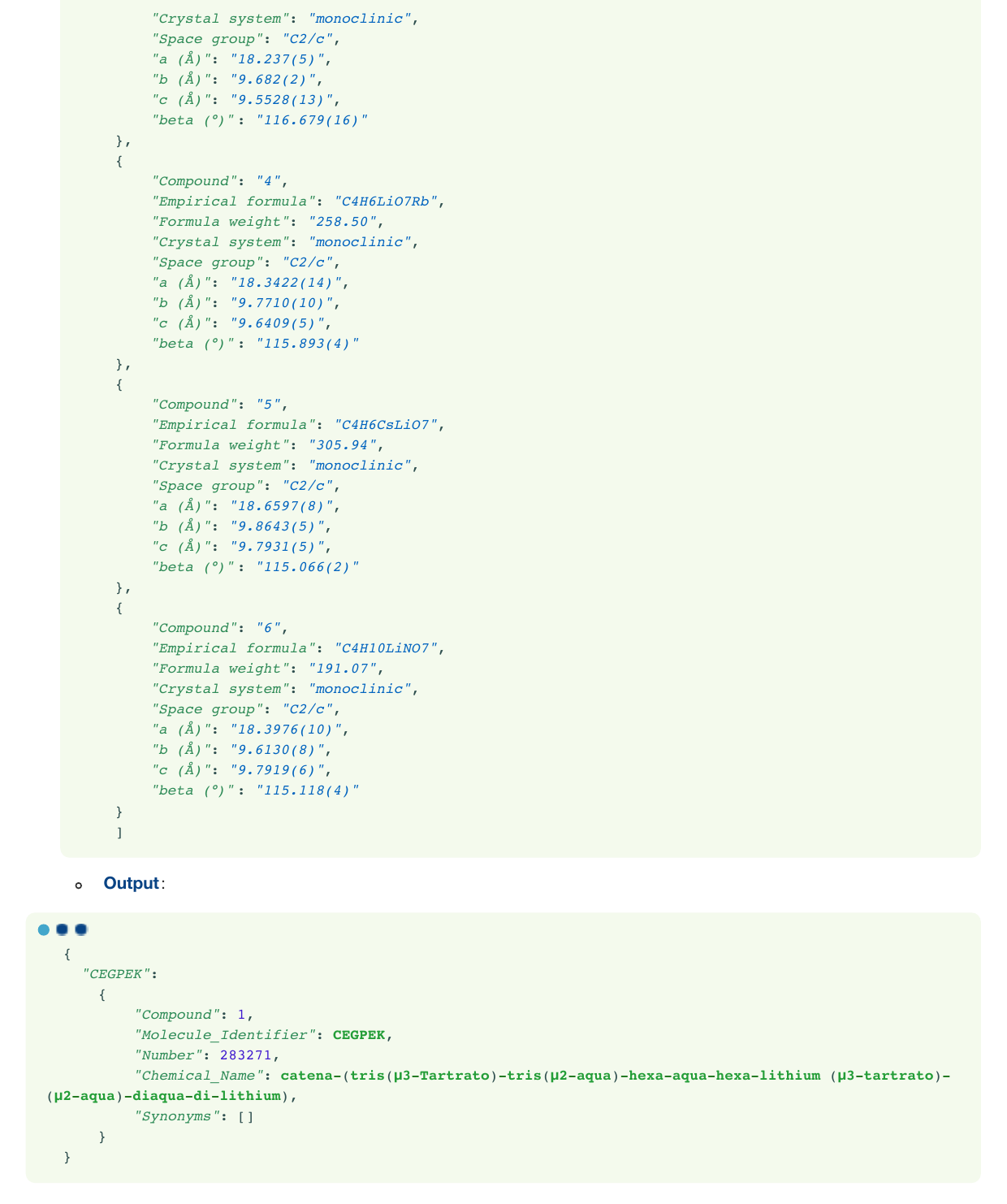}
  \captionsetup{
    labelformat=default,
    list=true
  }
  \caption{Shot 5-3 of crystal data comparison agent (This case is located at 5-shots in Figure \ref{3-1Cprompt_2})}
  \label{3-2C_5}
\end{figure}

\begin{figure}[H]
  \centering
  \includegraphics[width=\textwidth]{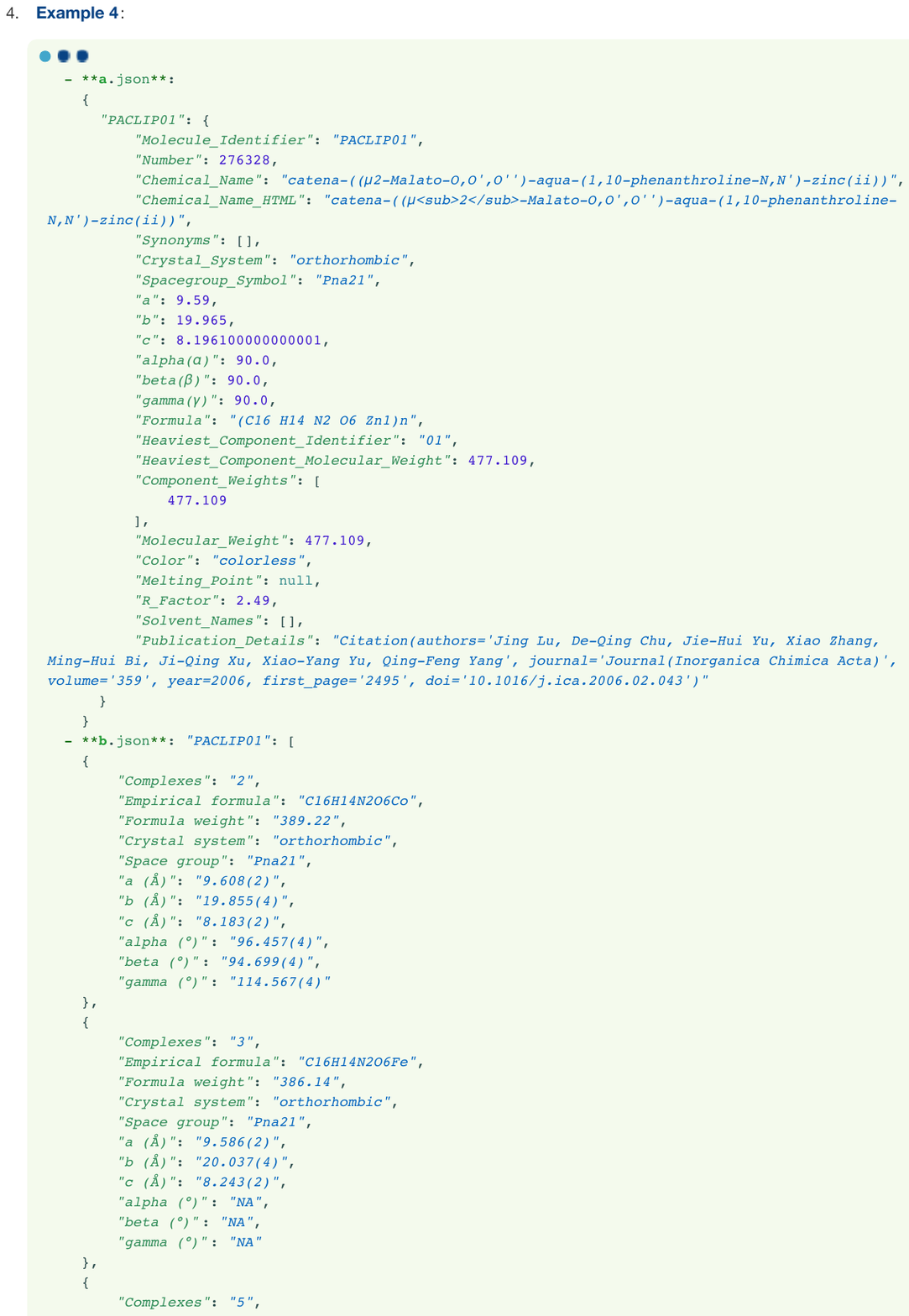}
  \captionsetup{
    labelformat=empty,
    list=false
  }
  \caption{} 
\end{figure}

\phantomsection
\addcontentsline{toc}{section}{Figure S28: Shot 5-4 of crystal data comparison agent}
\begin{figure}[H]\ContinuedFloat
  \centering
  \includegraphics[width=\textwidth]{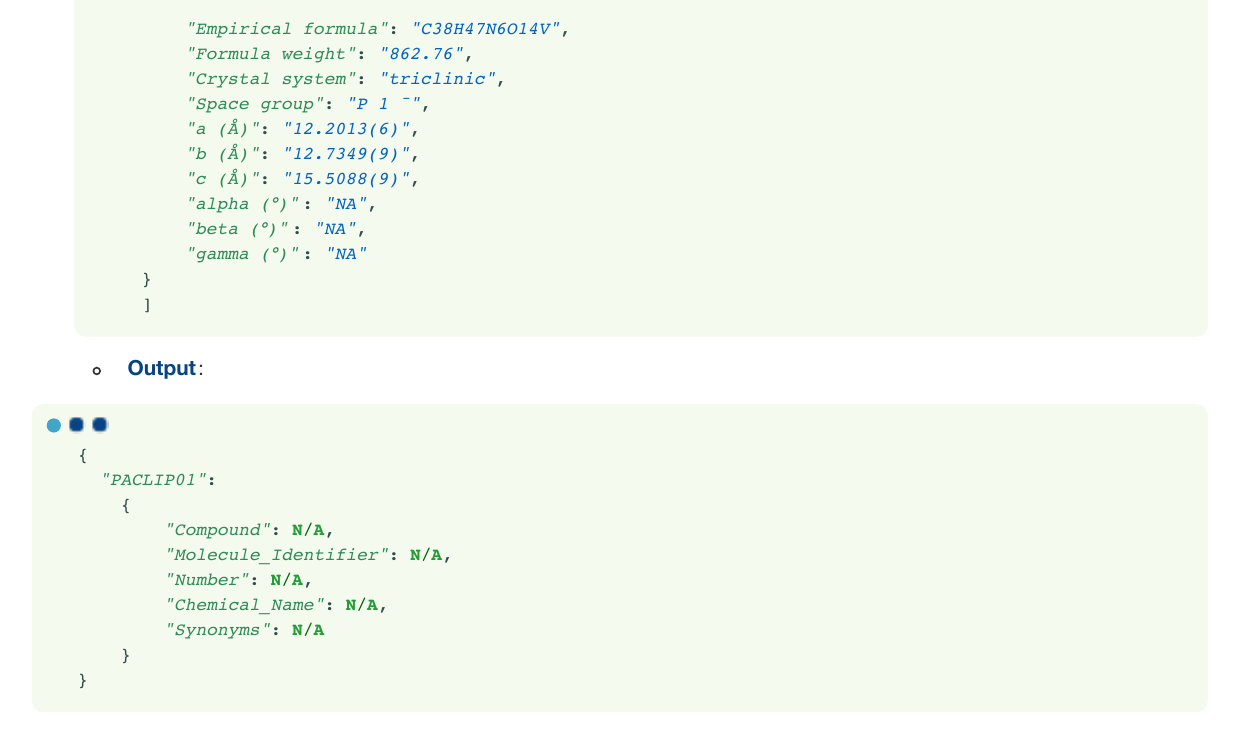}
  \captionsetup{
    labelformat=default,
    list=true
  }
  \caption{Shot 5-4 of crystal data comparison agent (This case is located at 5-shots in Figure \ref{3-1Cprompt_2})}
  \label{3-2C_7}
\end{figure}

\begin{figure}[H]
  \centering
  \includegraphics[width=\textwidth]{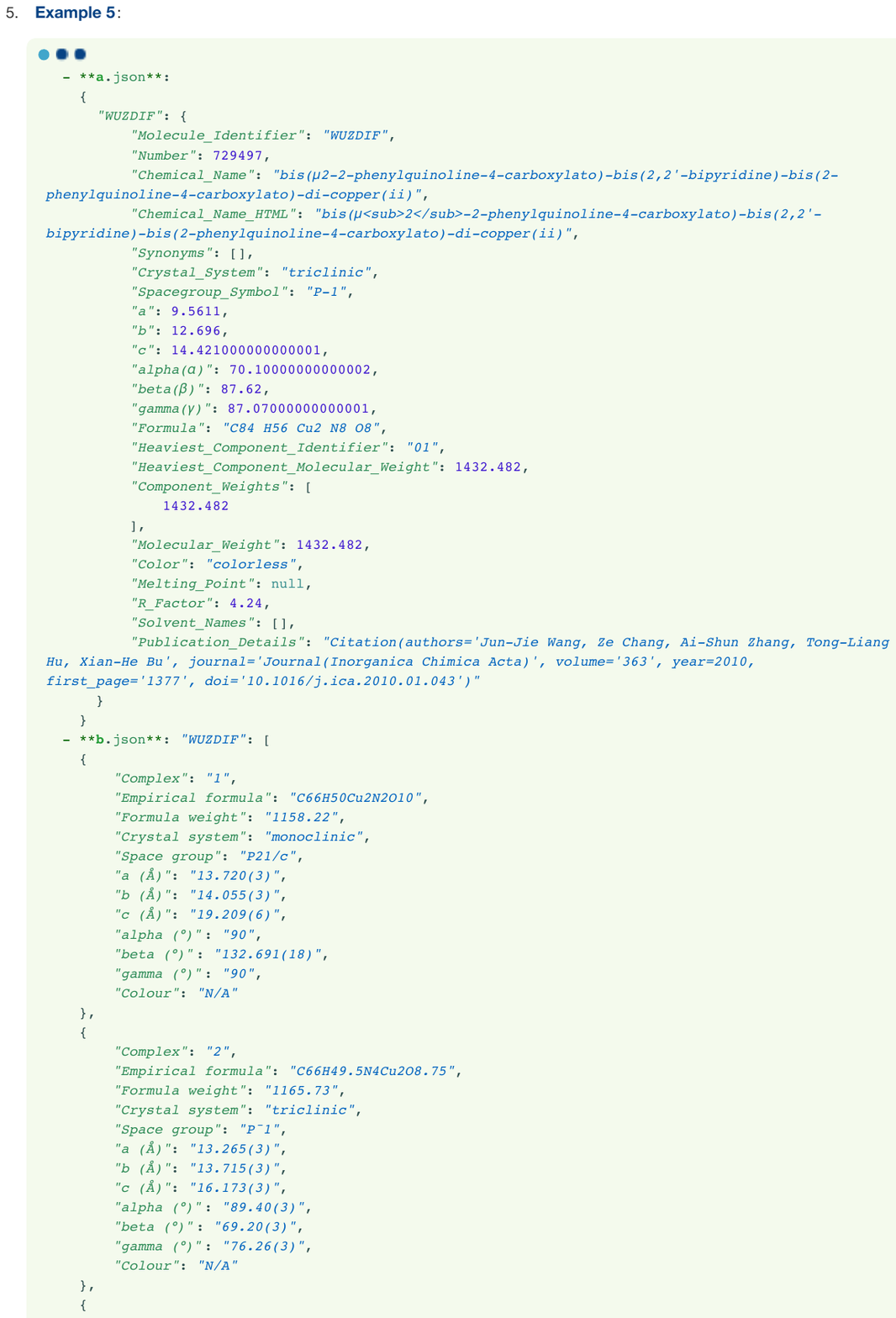}
  \captionsetup{
    labelformat=empty,
    list=false
  }
  \caption{}  
\end{figure}
\begin{figure}[H]\ContinuedFloat
  \centering
  \includegraphics[width=\textwidth]{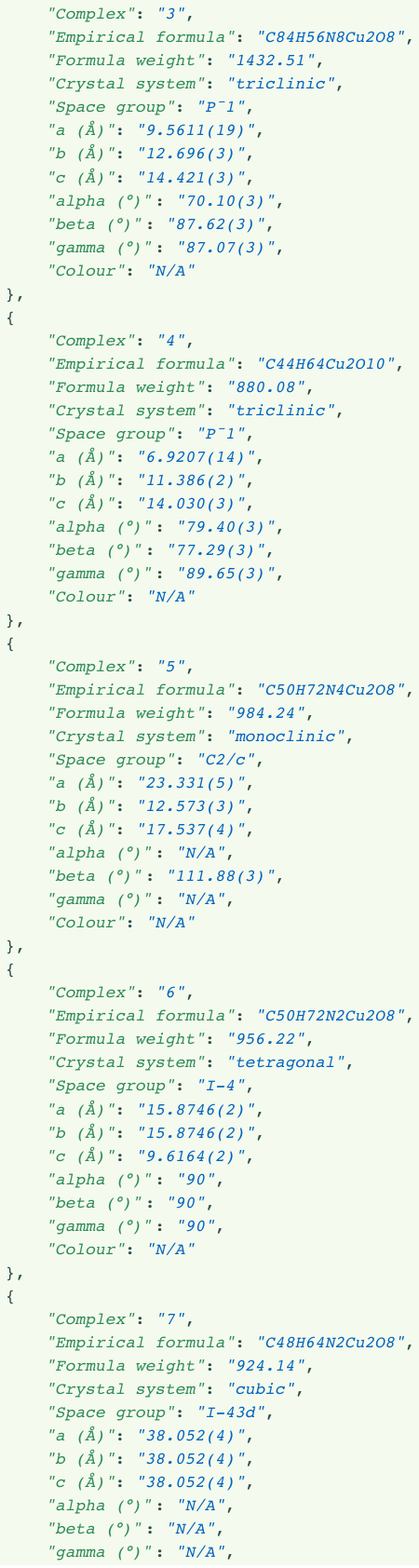}
  \captionsetup{
    labelformat=empty,
    list=false
  }
  \caption{} 
\end{figure}

\phantomsection
\addcontentsline{toc}{section}{Figure S29: Shot 5-5 of crystal data comparison agent}
\begin{figure}[H]\ContinuedFloat
  \centering
  \includegraphics[width=\textwidth]{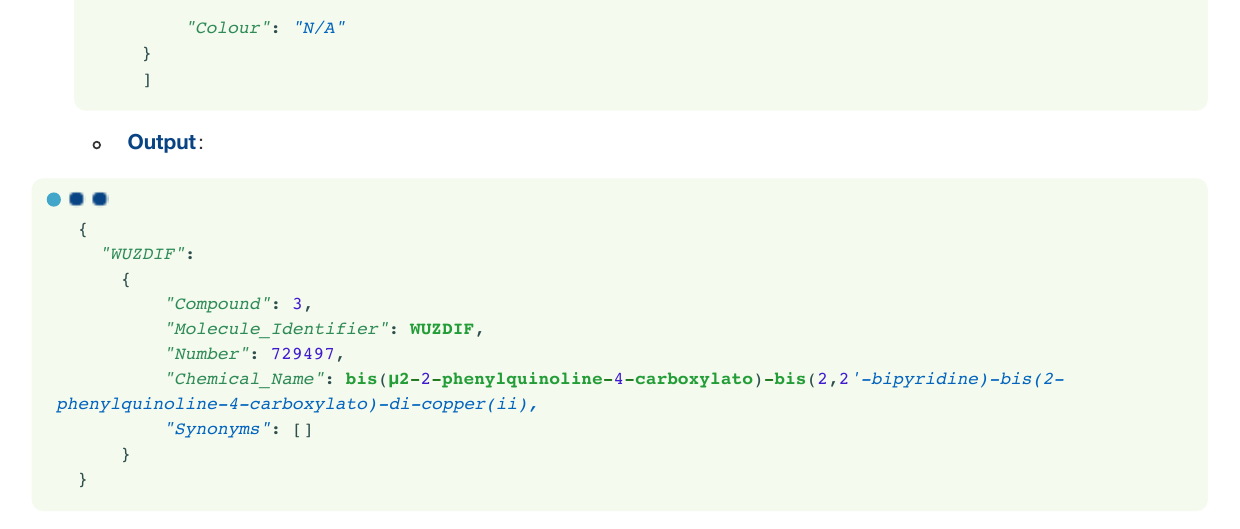}
  \captionsetup{
    labelformat=default,
    list=true
  }
  \caption{Shot 5-5 of crystal data comparison agent (This case is located at 5-shots in Figure \ref{3-1Cprompt_2})}
  \label{3-2C_10}
\end{figure}

\phantomsection
\addcontentsline{toc}{section}{Figure S30: Shot 15-1 to 4 of chemical abbreviation resolution agent}
\begin{figure}[H]
  \centering
  \includegraphics[width=1\textwidth]{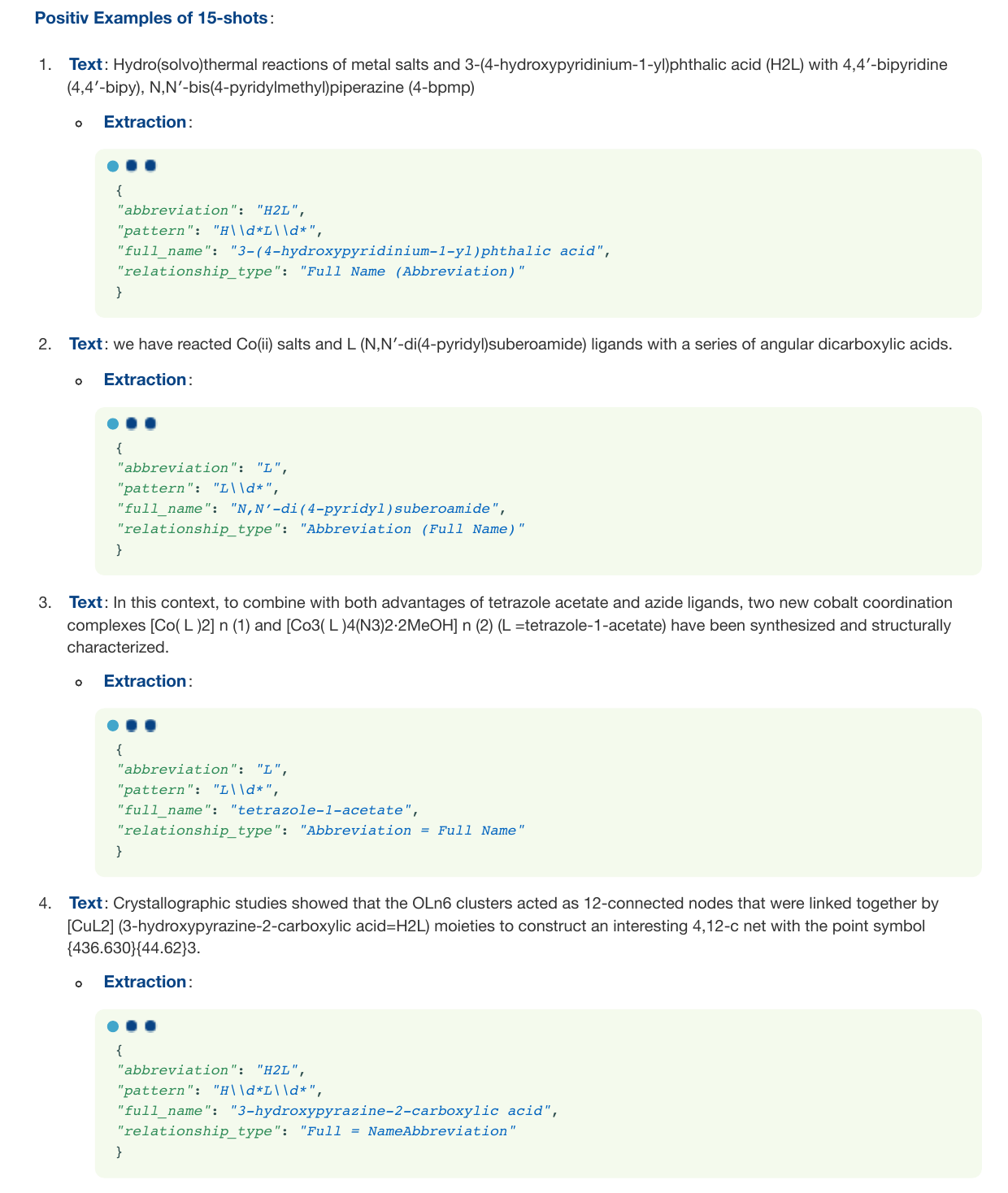}
  \caption{
  \centering
  Shot 15-1 to 4 of chemical abbreviation resolution agent (These cases are located at 15-shots in Figure \ref{4-1crprompt_3})
  }
  \label{4-2cr_1}
\end{figure}

\phantomsection
\addcontentsline{toc}{section}{Figure S31: Shot 15-5 to 8 of chemical abbreviation resolution agent}
\begin{figure}[H]
  \centering
  \includegraphics[width=1\textwidth]{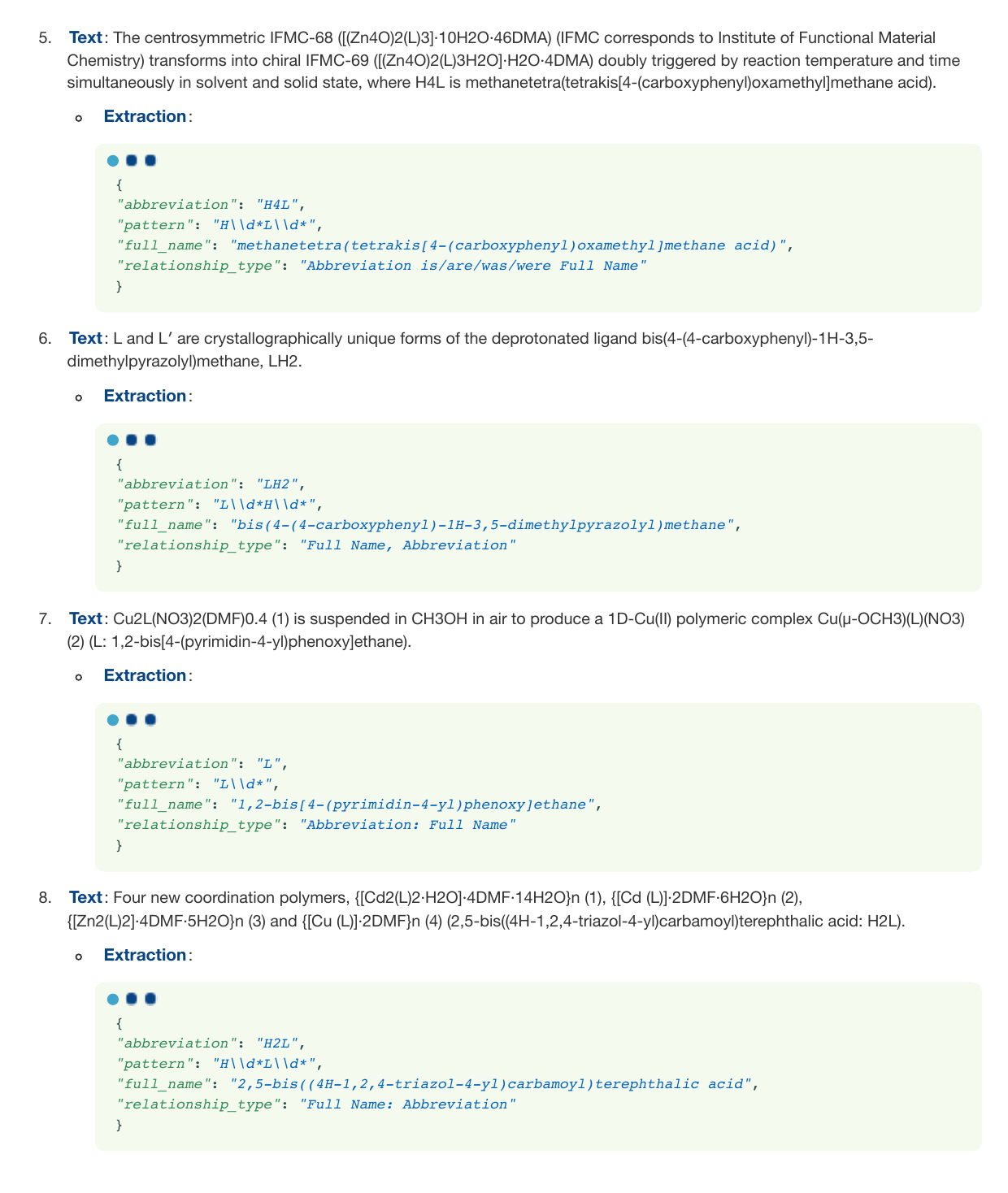}
  \caption{
  \centering
  Shot 15-5 to 8 of chemical abbreviation resolution agent (These cases are located at 15-shots in Figure \ref{4-1crprompt_3})
  }
  \label{4-2cr_2}
\end{figure}

\phantomsection
\addcontentsline{toc}{section}{Figure S32: Shot 15-9 to 12 of chemical abbreviation resolution agent}
\begin{figure}[H]
  \centering
  \includegraphics[width=1\textwidth]{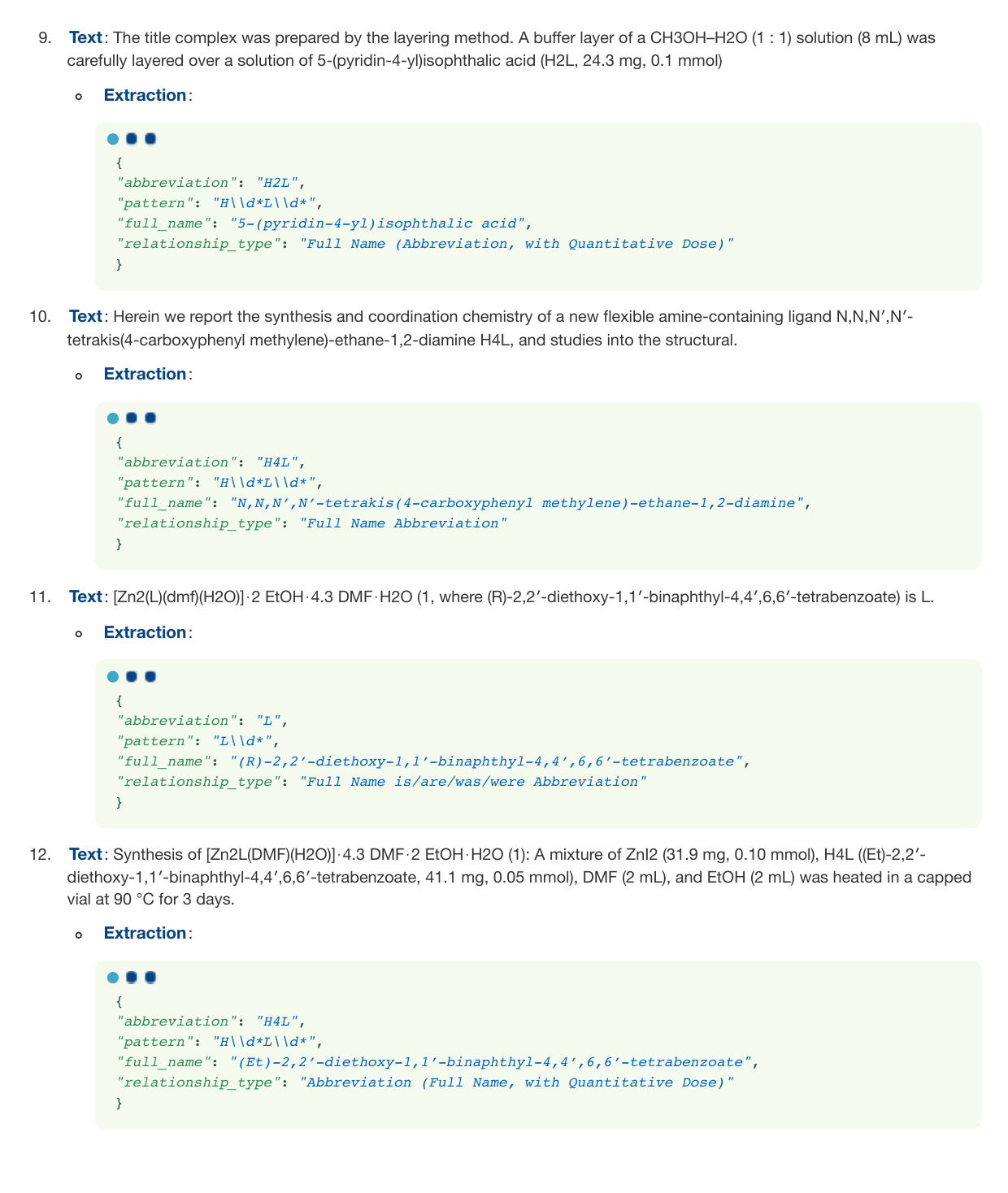}
  \caption{
  \centering
  Shot 15-9 to 12 of chemical abbreviation resolution agent (These cases are located at 15-shots in Figure \ref{4-1crprompt_3})
  }
  \label{4-2cr_3}
\end{figure}

\phantomsection
\addcontentsline{toc}{section}{Figure S33: Shot 15-13 to 15 of chemical abbreviation resolution agent}
\begin{figure}[H]
  \centering
  \includegraphics[width=1\textwidth]{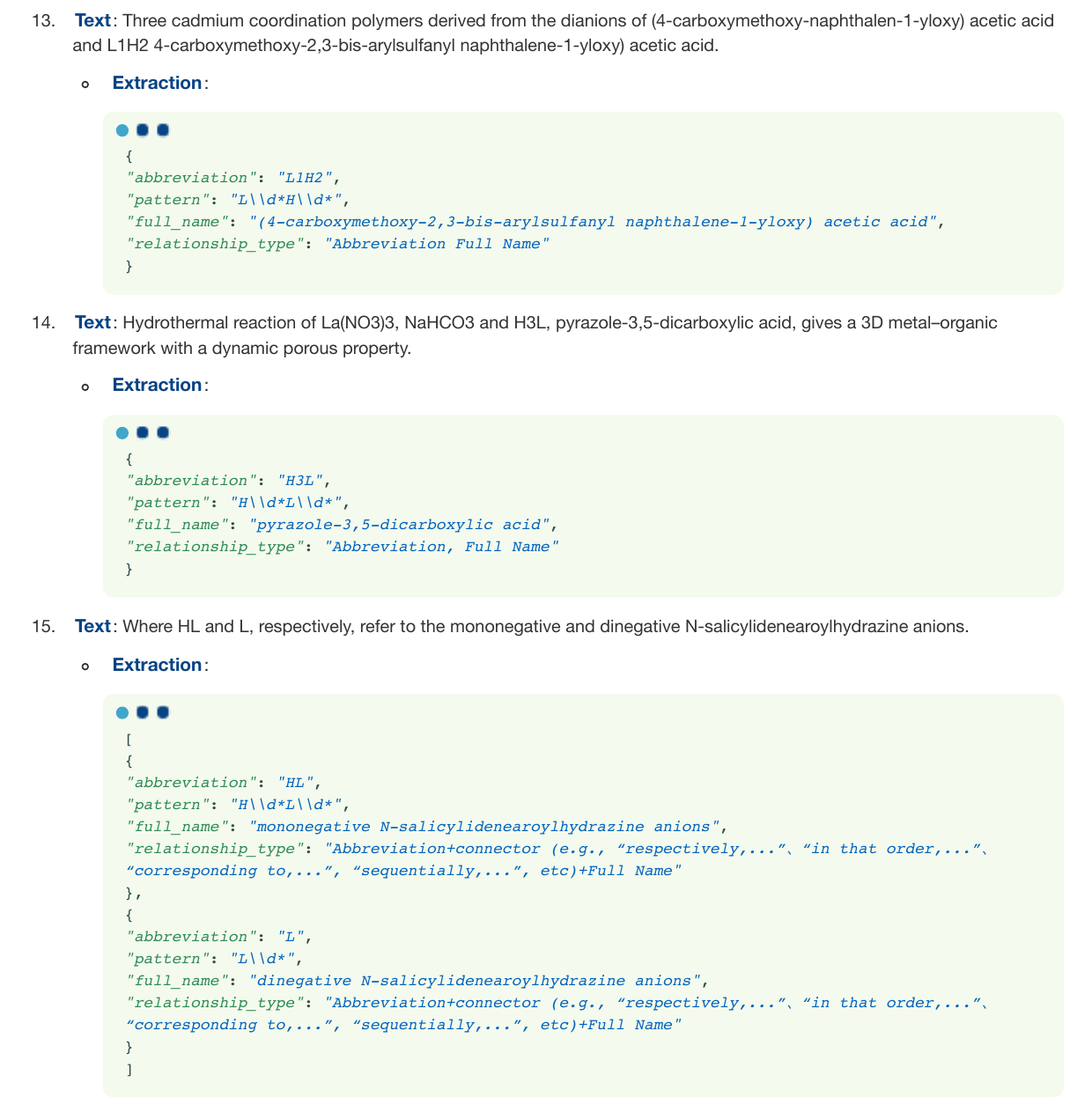}
  \caption{
  \centering
  Shot 15-13 to 15 of chemical abbreviation resolution agent (These cases are located at 15-shots in Figure \ref{4-1crprompt_3})
  }
  \label{4-2cr_4}
\end{figure}

\phantomsection
\addcontentsline{toc}{section}{Figure S34: Shot 4-1 to 2 of chemical abbreviation resolution agent}
\begin{figure}[H]
  \centering
  \includegraphics[width=1\textwidth]{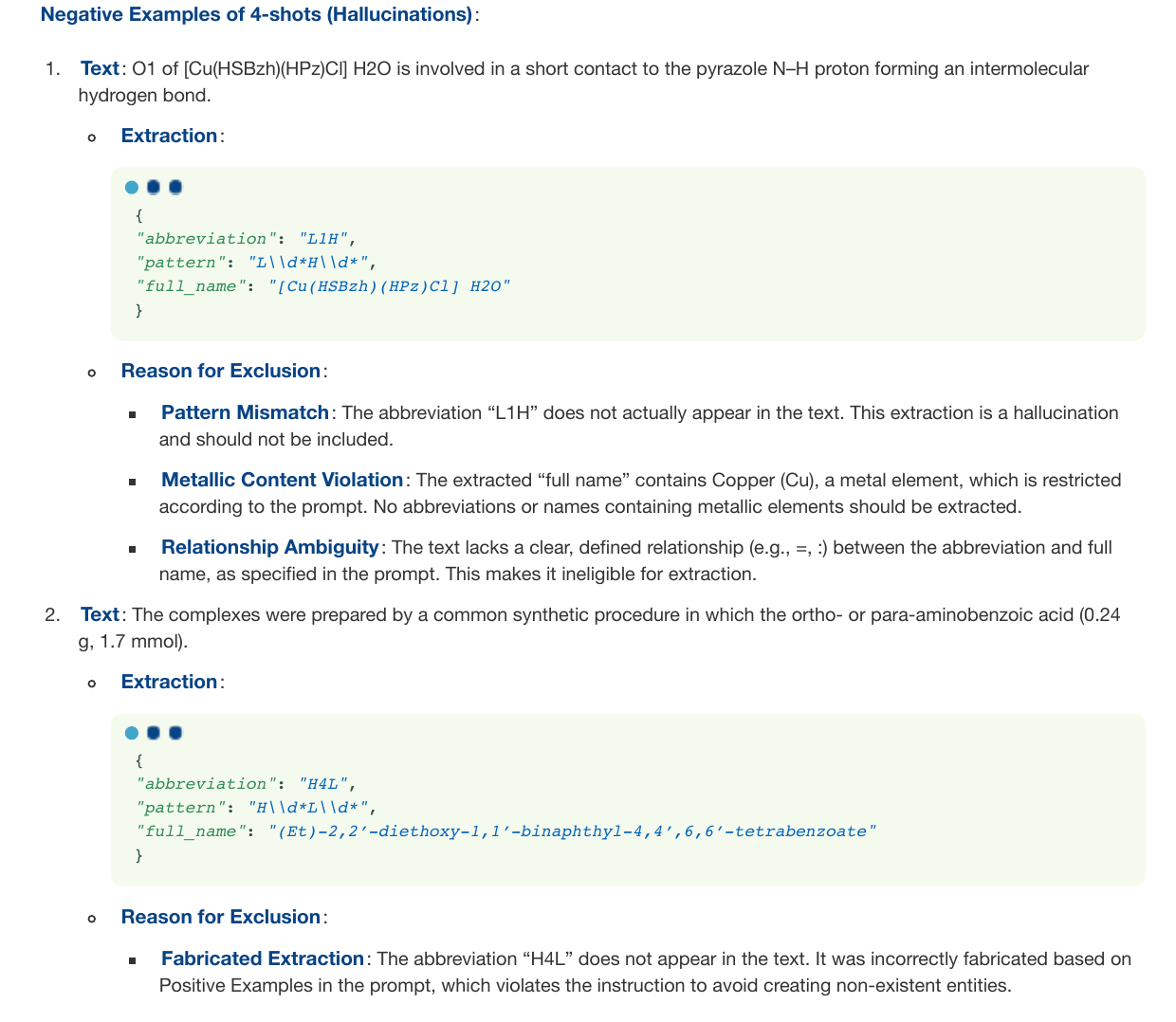}
  \caption{
  \setlength{\baselineskip}{20pt}
  \centering
  Shot 4-1 to 2 of chemical abbreviation resolution agent (These cases are located at 4-shots in Figure \ref{4-1crprompt_3})
  }
  \label{4-2cr_5}
\end{figure}

\phantomsection
\addcontentsline{toc}{section}{Figure S35: Shot 4-3 to 4 of chemical abbreviation resolution agent}
\begin{figure}[H]
  \centering
  \includegraphics[width=1\textwidth]{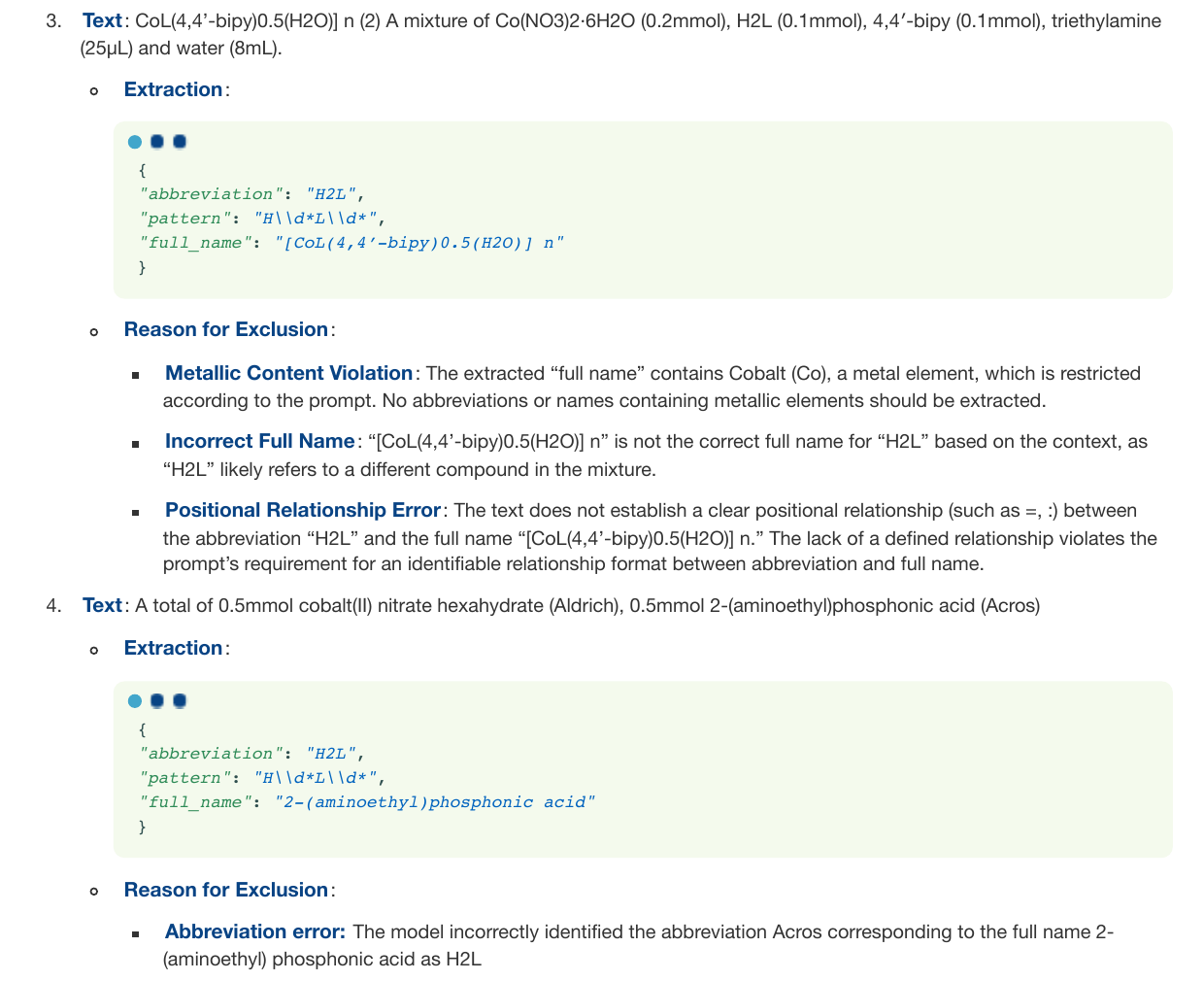}
  \caption{
  \centering
  Shot 4-3 to 4 of chemical abbreviation resolution agent (These cases are located at 4-shots in Figure \ref{4-1crprompt_3})
  }
  \label{4-2cr_6}
\end{figure}

\begin{figure}[H]
  \centering
  \includegraphics[width=\textwidth]{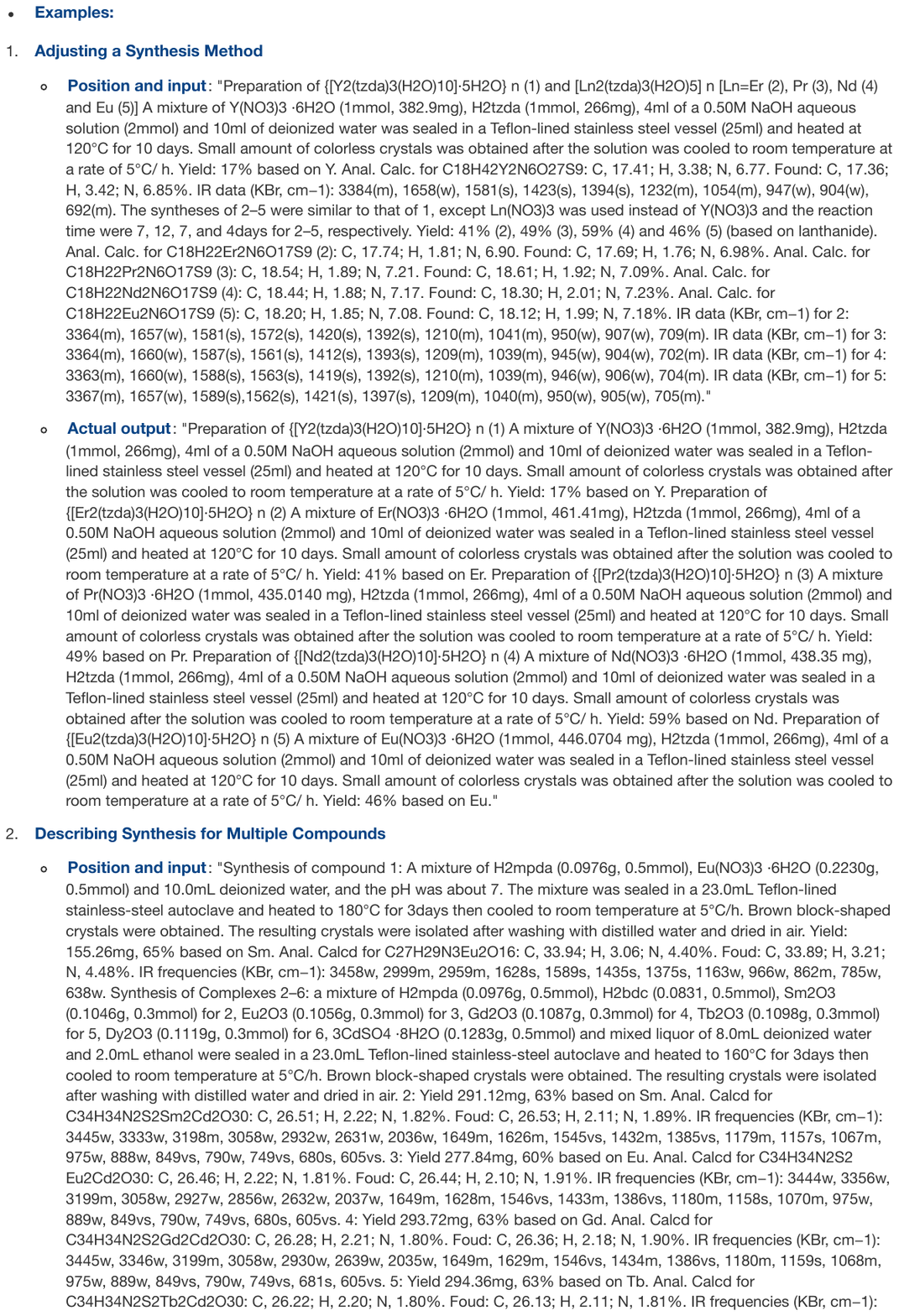}
  \captionsetup{
    labelformat=empty,
    list=false
  }
  \caption{} 
\end{figure}
\phantomsection
\addcontentsline{toc}{section}{Figure S36: Shot 3-1 to 3 of fine-tuning model}
\begin{figure}[H]\ContinuedFloat
  \centering
  \includegraphics[width=\textwidth]{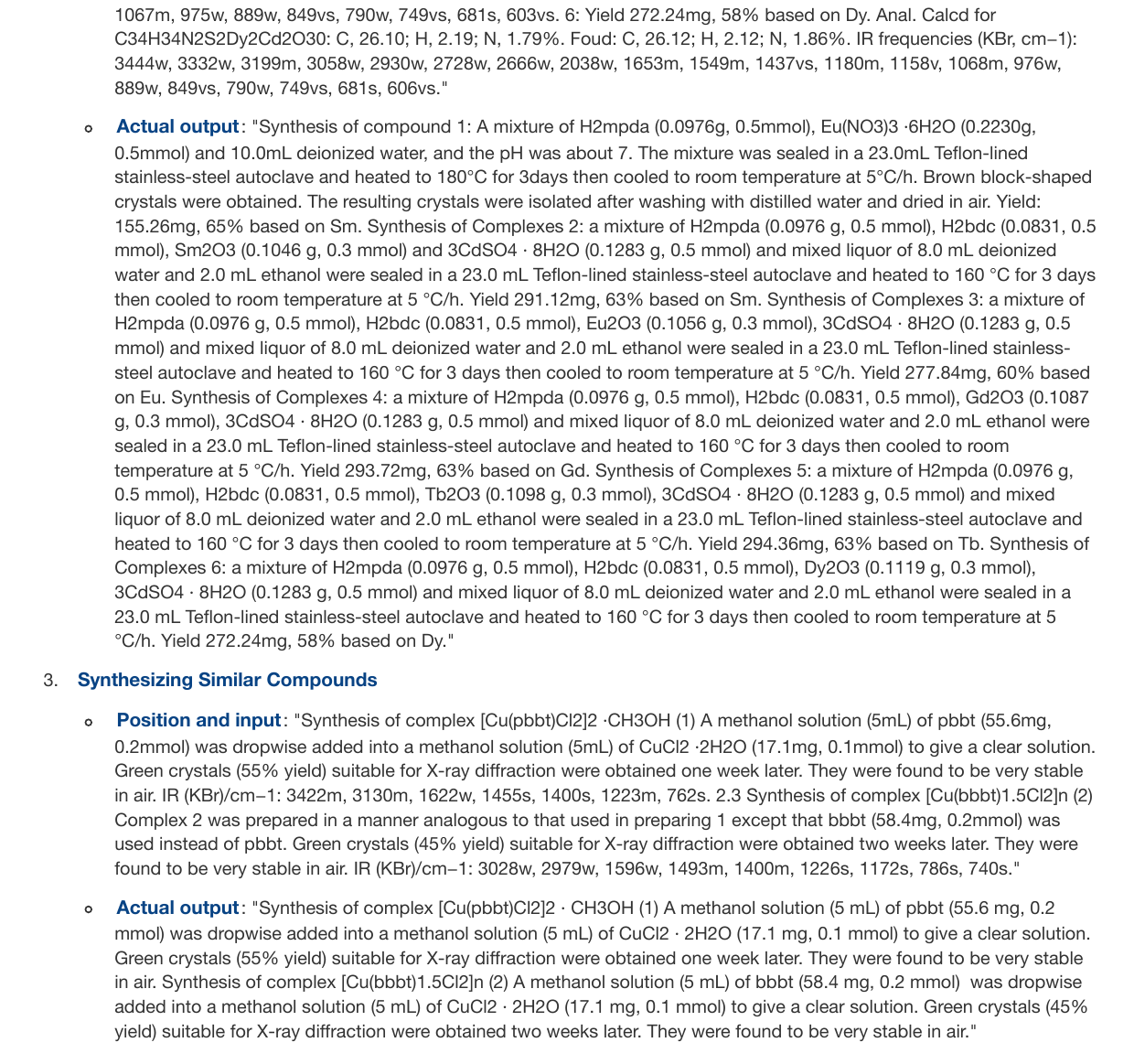}
  \captionsetup{
    labelformat=default,
    list=true
  }
  \caption{Shot 3-1 to 3 of fine-tuning model (These cases are located at 5-shots in Figure \ref{ftprompt_1})}
  \label{ftprompte_2}
\end{figure}

\newpage
\phantomsection
\addcontentsline{toc}{section}{Figure S37: Structure Q\&A of MOFh6: Welcome to the system}
\begin{figure}[H]
  \centering
  \includegraphics[width=1\textwidth]{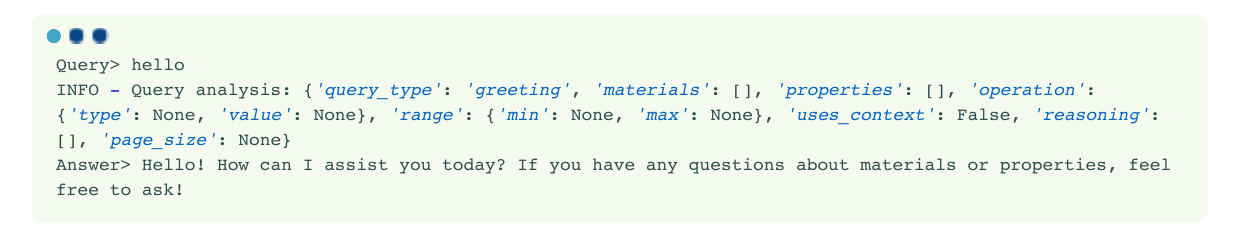}
  \caption{
  \centering
  Structure Q\&A of MOFh6: Welcome to the system
  }
  \label{er_1}
\end{figure}

\newpage
\phantomsection
\addcontentsline{toc}{section}{Figure S38: Structure Q\&A of MOFh6: Directly ask a single characteristic}

\phantomsection
\addcontentsline{toc}{section}{Figure S39: Structure Q\&A of MOFh6: Context memory}

\phantomsection
\addcontentsline{toc}{section}{Figure S40: Structure Q\&A of MOFh6: Summative questions on specific structures}

\phantomsection
\addcontentsline{toc}{section}{Figure S41: Structure Q\&A of MOFh6: Finding structures within the numerical range}

\phantomsection
\addcontentsline{toc}{section}{Figure S42: Structure Q\&A of MOFh6: Comparing properties between structures}

\newpage
\phantomsection
\addcontentsline{toc}{section}{Figure S43: Structure Q\&A of MOFh6: Numerical statistics of the dataset}

\phantomsection
\addcontentsline{toc}{section}{Figure S44: Structure Q\&A of MOFh6: Comprehensive structural query based on multiple structural characteristics}

\phantomsection
\addcontentsline{toc}{section}{Figure S45: Multi language support of MOFh6}
\begin{figure}[H]
  \centering
  \includegraphics[width=1\textwidth]{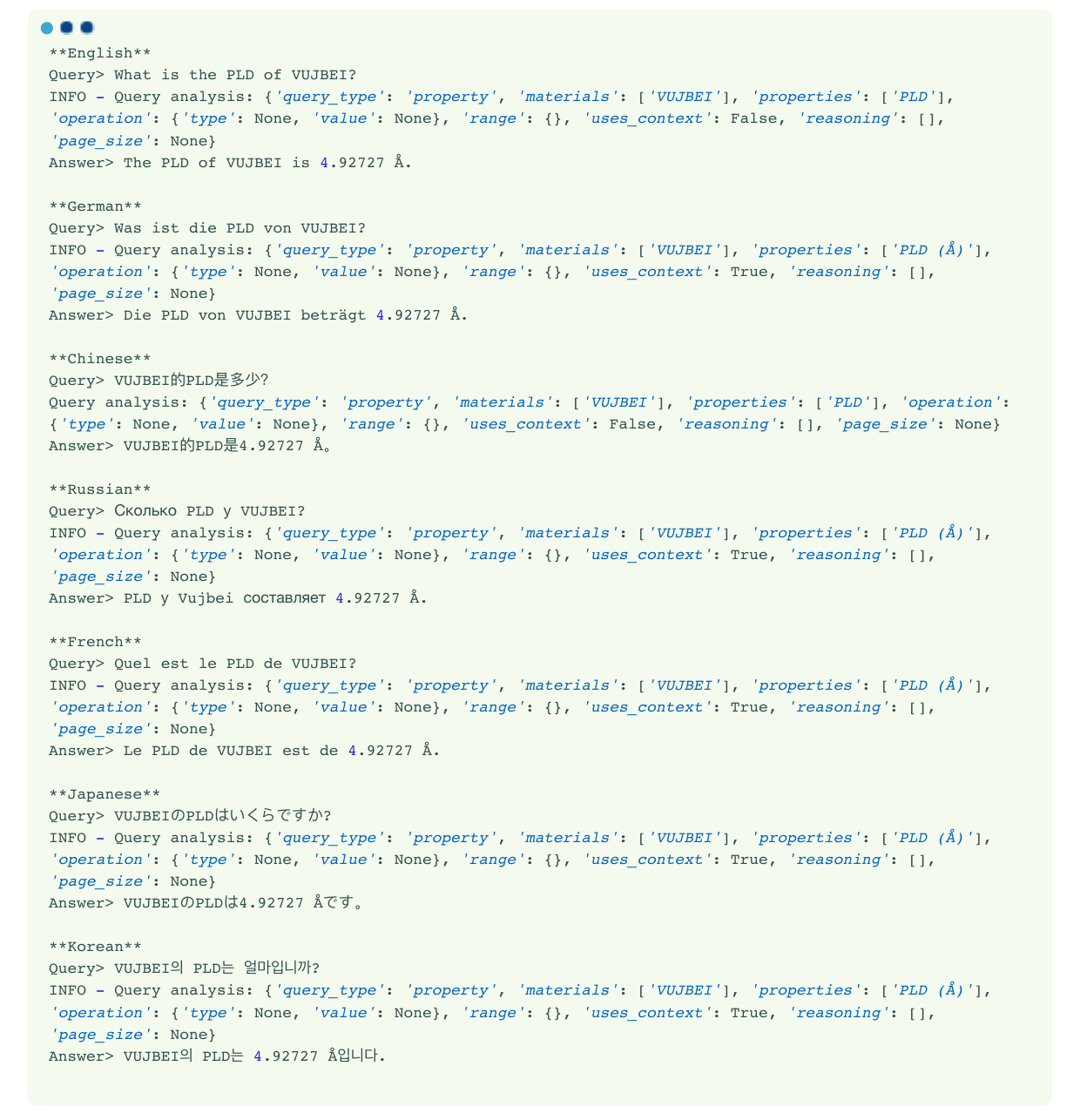}
  \caption{
  \centering
  Multi language support of MOFh6
  }
  \label{er_9}
\end{figure}

\phantomsection
\addcontentsline{toc}{section}{Figure S46: MOFh6 system query parameter fields}

\phantomsection
\addcontentsline{toc}{section}{Figure S47: Data mining statistics, 
  frequency of occurrence of metal elements that make up MOFs (a), 
  crystal system (b), space group symbol (c), 
  publisher publication volume (d)}
\begin{figure}[H]
  \centering
  \includegraphics[width=1\textwidth]{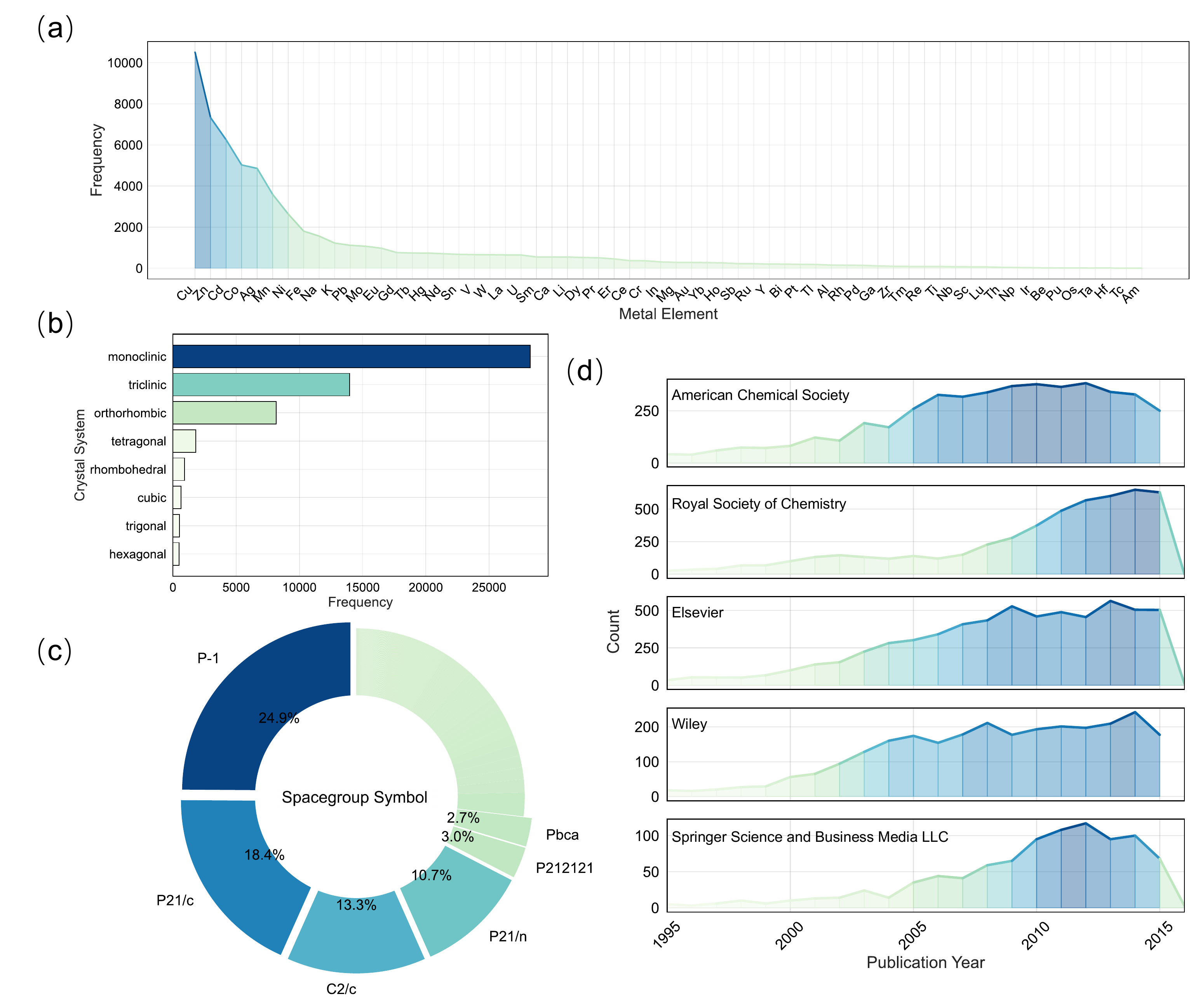}
  \caption{
  \centering
  Data mining statistics, 
  frequency of occurrence of metal elements that make up MOFs (a), 
  crystal system (b), space group symbol (c), 
  publisher publication volume (d)
  }
  \label{LLMM3}
\end{figure}

\phantomsection
\addcontentsline{toc}{section}{Figure S48: Statistics of the top ten journals from different publishers,
ACS (a), RSC (b), Elsevier (c), Wiley (d), Springer (e)}
\begin{figure}[H]
  \centering
  \includegraphics[width=1\textwidth]{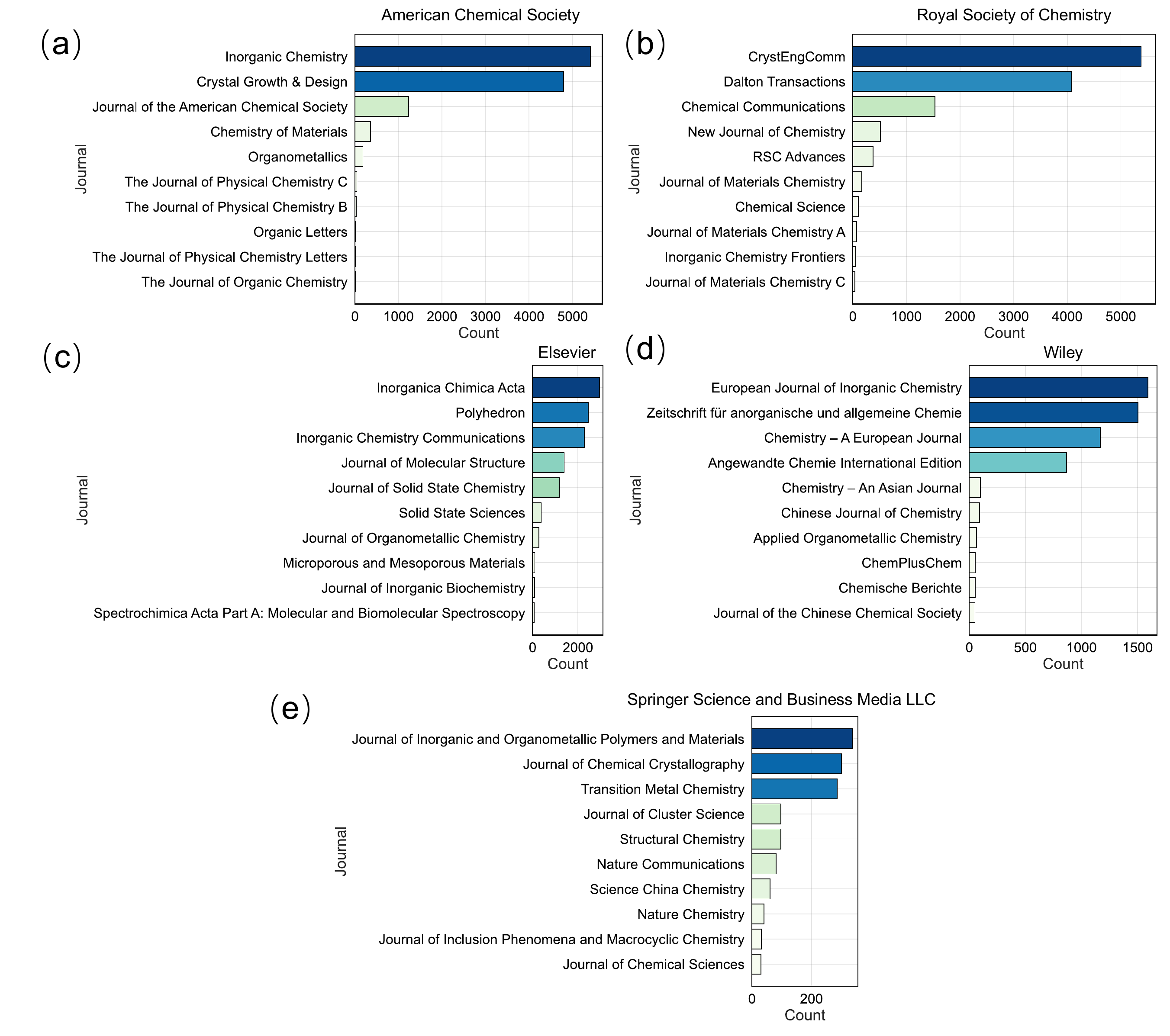}
  \caption{
  \centering
  Statistics of the top ten journals from different publishers,
  ACS (a), RSC (b), Elsevier (c), Wiley (d), Springer (e)
  }
  \label{LLMS31}
\end{figure}

\phantomsection
\addcontentsline{toc}{section}{Figure S49: Structural properties of MOFs calculated by \textit{Zeo++}}
\begin{figure}[H]
  \centering
  \includegraphics[width=1\textwidth]{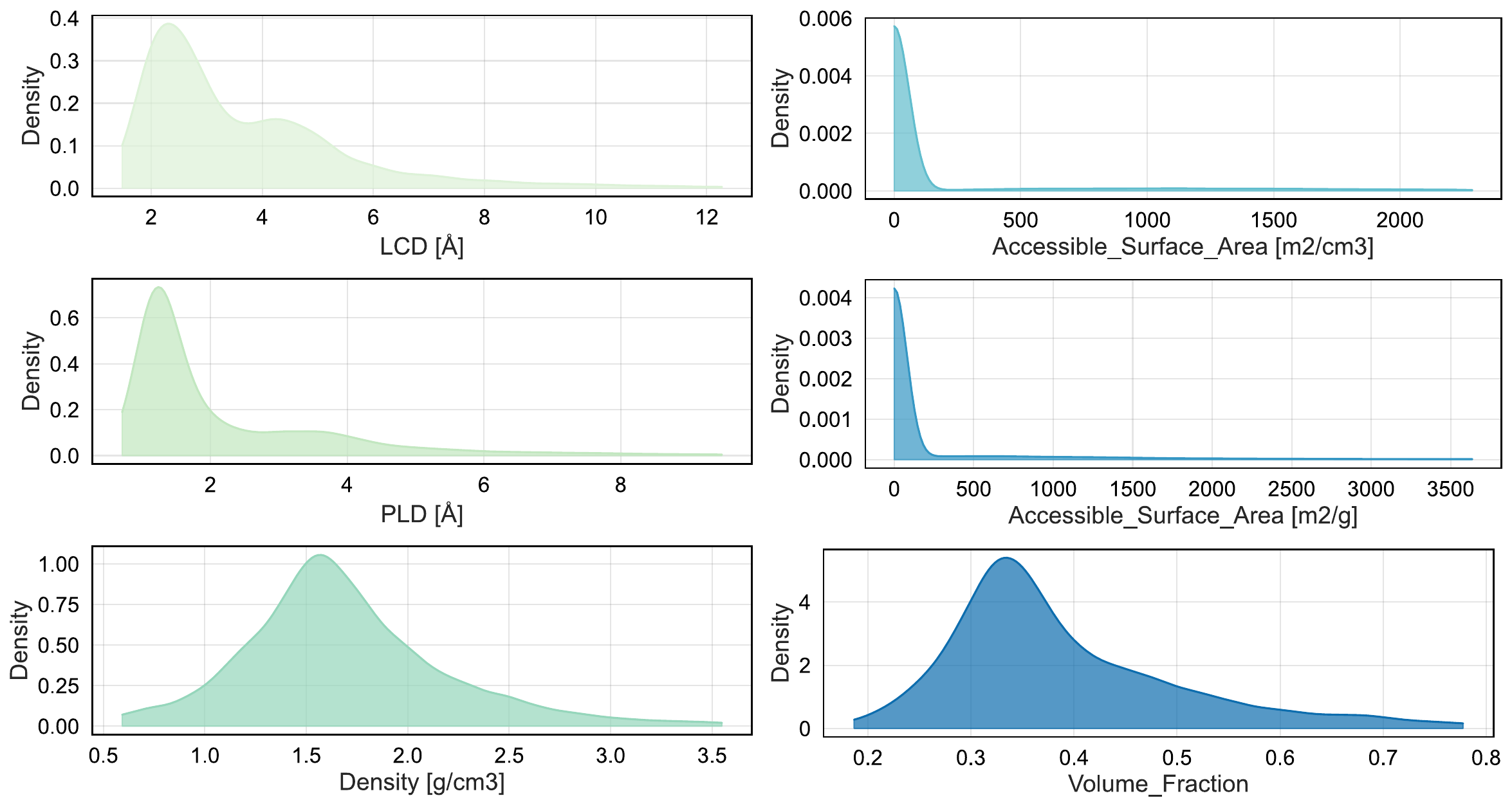}
  \caption{
  \centering
  Structural properties of MOFs calculated by \textit{Zeo++}
  }
  \label{LLMS32}
\end{figure}

\phantomsection
\addcontentsline{toc}{section}{Figure S50: CIF download and visualization of MOFh6}
\begin{figure}[H]
  \centering
  \includegraphics[width=0.8\textwidth]{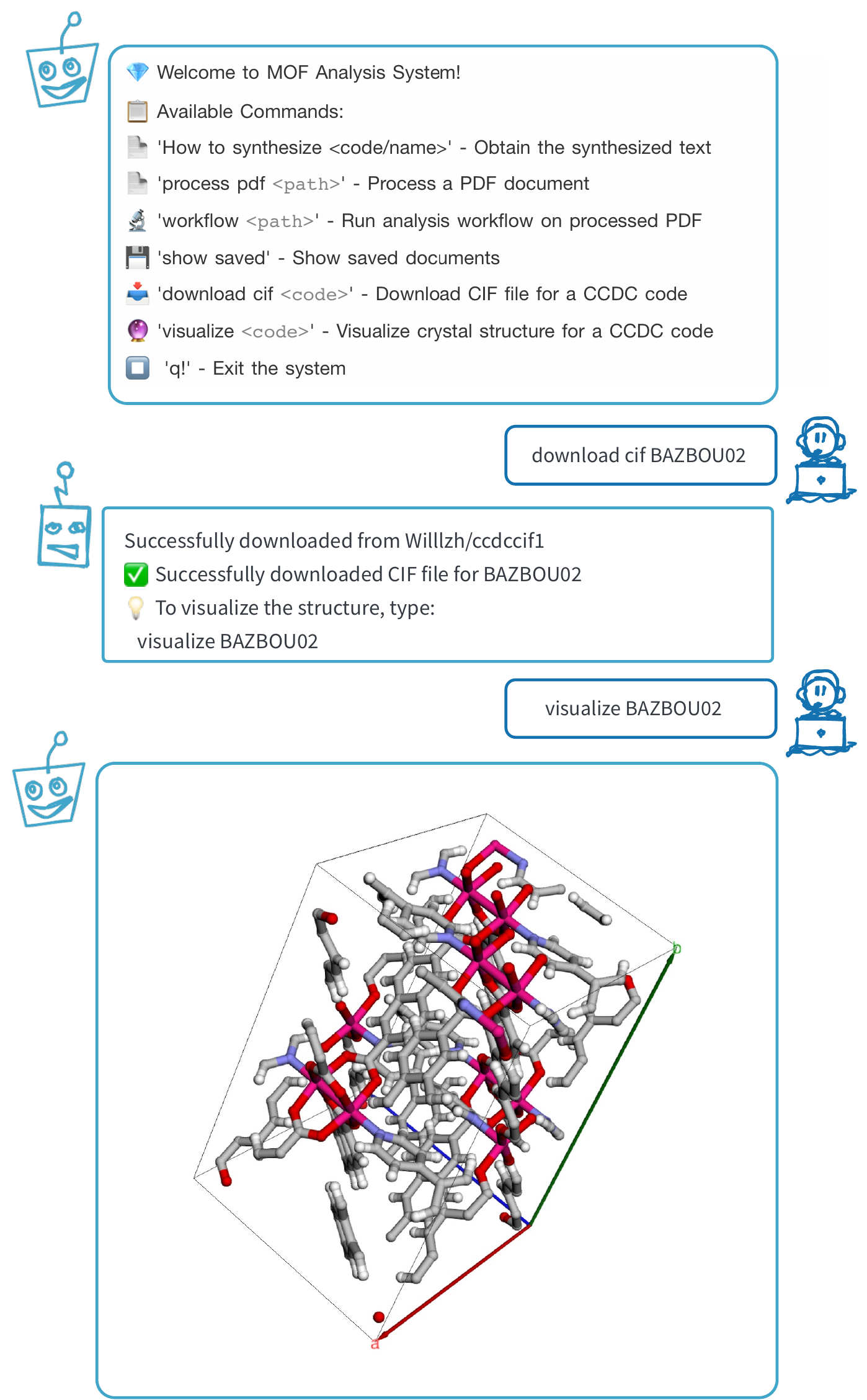}
  \caption{
  \centering
  CIF download and visualization of MOFh6
  }
  \label{viscif}
\end{figure}

\phantomsection
\addcontentsline{toc}{section}{Figure S51: Cost comparison between MOFh6 and similar works}

\phantomsection
\addcontentsline{toc}{section}{Figure S52: MOFh6 data mining processing final localization for specified MOFs synthesis requests}
\begin{figure}[H]
  \centering
  \includegraphics[width=1\textwidth]{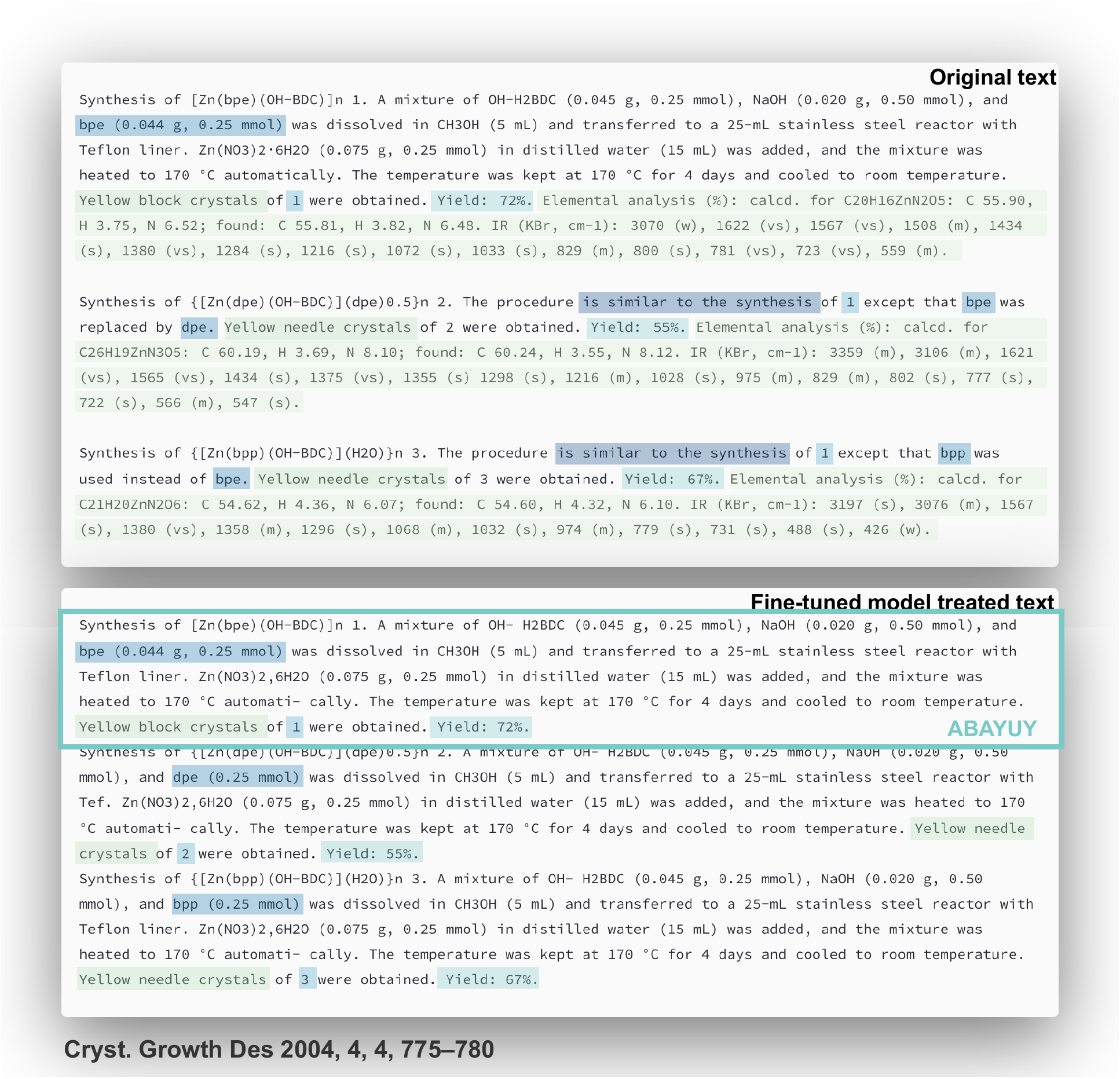}
  \caption{
  \centering
  MOFh6 data mining processing final localization for specified MOFs synthesis requests
  }
  \label{llmcase}
\end{figure}

\newpage
\phantomsection
\addcontentsline{toc}{section}{Table S1: The core agents of MOFh6}
\begin{table}[htb]
  \centering
  \setlength{\abovecaptionskip}{2pt}
  \setlength{\belowcaptionskip}{0pt}
  \caption{
  \centering
  The core agents of MOFh6}
  \label{LLMagent}
  \begin{tabular}{>{\centering\arraybackslash}p{3cm}>{\centering\arraybackslash}p{4cm}>{\centering\arraybackslash}p{3.5cm}>{\centering\arraybackslash}p{1cm}}
  \toprule
  Agent name & Core model algorithm & Output format & System level \\
  \midrule
  Synthetic data parsing agent & GPT-4o-mini fine-tuning & .xlsx & 1 \\
  Table data parsing agent & GPT-4o-mini prompt + Structured parsing & tables\_*.json & \\
  Crystal data comparison agent & GPT-4o prompt + Structural similarity calculation & comparison\_*.json & \\
  Chemical abbreviation resolution agent & GPT-4o prompt + Regular expression pattern matching & acronym\_results\_*.json & 2 \\
  Post processor & GPT-4o mini prompt + Text segmentation + File organization management & identifier\_*.txt & \\
  Result Generator agent & GPT-4o-mini prompt + BM25 algorithm & final\_output\_*.txt & 3 \\
  Structured conversion agent & GPT-4o-mini prompt + Markdown conversion & structure\_*.md & \\
  \bottomrule
  \end{tabular}
\end{table}

\newpage
\phantomsection
\addcontentsline{toc}{section}{Table S2: Models for evaluating the performance of MOFh6}
\begin{table}[htb]
  \centering
  \setlength{\abovecaptionskip}{2pt}
  \setlength{\belowcaptionskip}{0pt}
  \caption{
  \centering
  Models for evaluating the performance of MOFh6}
  \label{LLMIDex}
  \begin{tabular}{ccc}
  \toprule
  Level & Index & Model \\
  \midrule
  \multirow{1}{*}{Sentence}
    & Description of synthetic paragraphs & BiomedBERT \\
  \midrule
  \multirow{13}{*}{Data frame}
    & Metal Source &  \\
    & Organic Linkers Source   &  PubMedBERT \\
    & Modulator Source   &  \\
    & Solvent Source &  \\
    & Quantity of Metal   &  \\
    & Quantity of Organic Linkers   &  \\
    & Quantity of Modulator &  \\
    & Quantity of Solvent   &  \\
    & Synthesis Temperature   &  all-mpnet-base-v2 \\
    & Synthesis Time &  \\
    & Crystal Morphology   &  \\
    & Yield   &  \\
    & Equipment &  \\
  \bottomrule
  \end{tabular}
  \end{table}

\end{document}